\documentclass[10pt,twocolumn,letterpaper]{article}

\usepackage{iccv}

\usepackage{times}
\usepackage{epsfig}
\usepackage{graphicx}
\usepackage{amsmath}
\usepackage{amssymb}

\usepackage{booktabs}       %
\usepackage{amsfonts}       %
\usepackage{microtype}      %
\usepackage{xspace}
\usepackage{wrapfig}
\usepackage{amsthm}
\usepackage[font=footnotesize,skip=3pt,subrefformat=parens]{subcaption}
\usepackage{caption}
\usepackage{multirow}
\usepackage{multicol}
\usepackage[table,dvipsnames]{xcolor}
\usepackage{soul}
\usepackage{pifont}

\usepackage[pagebackref=true,breaklinks=true,letterpaper=true,colorlinks,bookmarks=false,citecolor=teal]{hyperref}

\iccvfinalcopy %

\ificcvfinal\fi

\definecolor{aliceblue}{rgb}{0.94, 0.97, 1.0}
\definecolor{pastelyellow}{RGB}{252, 249, 210}
\definecolor{hotpink}{rgb}{1, 0.078, 0.576}
\definecolor{goldyellow}{rgb}{0.882, 0.756, 0.05}
\definecolor{myred}{RGB}{232, 58, 55}

\newcommand\mpxap{\texttt{mPxAP}\xspace}

\newcommand{\textttsm}[1]{\texttt{\small#1\xspace}}
\newcommand\ours{\textttsm{LUAB}\xspace}
\newcommand\ourin{{ImageNet-AB}\xspace}
\newcommand\ourcoco{{COCO-AB}\xspace}

\renewcommand\footnotemark{}

\begin{document}

\title{Neglected Free Lunch --\\Learning Image Classifiers Using Annotation Byproducts}

\author{
Dongyoon Han$^\star$\\
{\small NAVER AI Lab}
\and
Junsuk Choe$^\star$\\
{\small Sogang University}
\and
Seonghyeok Chun\\
{\small Dante Company}
\and
John Joon Young Chung\\
{\small University of Michigan}
\and\thanks{$^\star$Equal contribution. $^\dagger$ currently at Google. Correspondence to Seong Joon Oh: \href{mailto:coallaoh@gmail.com}{coallaoh@gmail.com}.}
Minsuk Chang$^\dagger$\\
{\small NAVER AI Lab}
\and
Sangdoo Yun\\
{\small NAVER AI Lab}
\and
Jean Y. Song\\
{\small DGIST}
\and
Seong Joon Oh\\
{\small University of T\"ubingen}
}
\maketitle
\ificcvfinal\thispagestyle{empty}\fi

\begin{abstract}
\vspace{-1em}
Supervised learning of image classifiers distills human knowledge into a parametric model $f_\theta$ through pairs of images and corresponding labels $\{(X_i,Y_i)\}_{i=1}^N$. We argue that this simple and widely used representation of human knowledge neglects rich auxiliary information from the annotation procedure, such as the time-series of mouse traces and clicks left after image selection. Our insight is that such \textbf{annotation byproducts} $Z$ provide approximate human attention that weakly guides the model to focus on the foreground cues, reducing spurious correlations and discouraging shortcut learning. To verify this, we create \textbf{ImageNet-AB} and \textbf{COCO-AB}. They are ImageNet and COCO training sets enriched with sample-wise annotation byproducts, collected by replicating the respective original annotation tasks. We refer to the new paradigm of training models with annotation byproducts as \textbf{learning using annotation byproducts (LUAB)}. We show that a simple multitask loss for regressing $Z$ together with $Y$ already improves the generalisability and robustness of the learned models. Compared to the original supervised learning, LUAB does not require extra annotation costs. ImageNet-AB and COCO-AB are at \href{https://github.com/naver-ai/NeglectedFreeLunch}{github.com/naver-ai/NeglectedFreeLunch}.
\end{abstract}

\section{Introduction}

Supervised learning of image classifiers requires the transfer of human intelligence to a parametric model $f_\theta$. The transfer consists of two phases. First, human annotators execute human computation tasks~\cite{vonahn2007} to put labels $Y$ on each image $X$. The resulting labeled dataset $\{(X^i,Y^i)\}_{i=1}^N$ contains the gist of human knowledge about the visual task in a computation-friendly format. In the second phase, the model is trained to predict the labels $Y$ for each input $X$.

\begin{figure}
    \vspace{0.5em}
    \centering
    \includegraphics[width=.981\linewidth]{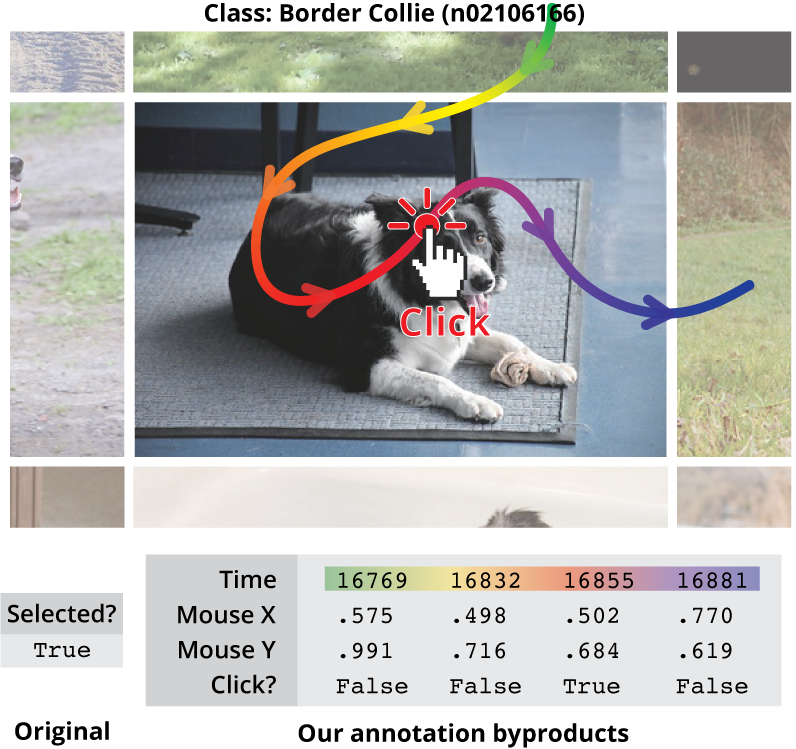}
    \vspace{-.7em}
    \caption{\small\textbf{Annotation byproducts from ImageNet.} Annotators leave traces like click locations as they select images with ``Border Collie''. We argue that such byproducts contain signals that may improve model generalisation and robustness.}
    \label{fig:teaser}
    \vspace{.3em}
    \includegraphics[width=.98\linewidth]{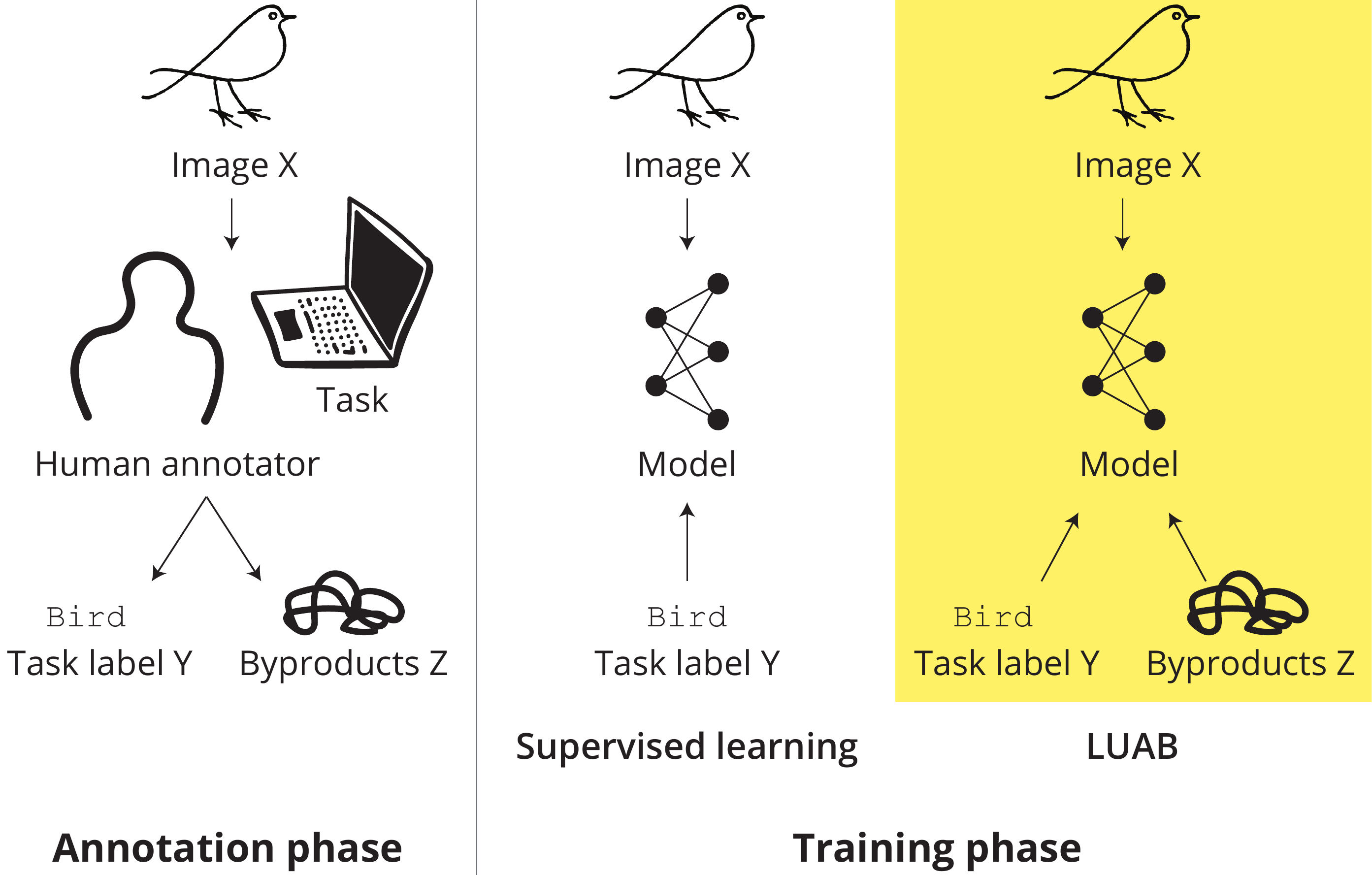}
    \vspace{-.7em}
    \caption{\small\textbf{Learning using Annotation Byproducts (LUAB).} LUAB exploits annotation byproducts $Z$ that are unintentionally generated during the human intelligence tasks for annotation.}
    \label{fig:luab}
    \vspace{-3em}
\end{figure}

In this work, we question the practice of collecting and utilising \textbf{only} the labels $Y$ for each image $X$ for training the models. In fact, common practise simply forgoes a large amount of additional signals from human annotators other than mere labels. When humans interact with computers through the graphical user interface, they leave various forms of unintentional traces. Input devices like the computer mouse produce time-series data in which information about what (\eg, mouse action type) and where (\eg, x-y coordinates in the monitor) are logged with timestamps. We refer to such auxiliary signals as \textbf{annotation byproducts} $Z$. See Figure \ref{fig:teaser} for an ImageNet annotation example~\cite{imagenet,imagenet_v2}. As annotators browse and click on images containing the class of interest, various byproducts are generated, \eg, images over which were hovered during selection, mouse movement speed between images, pixels on which were clicked in an image, images that were deselected due to mistake, and latency between image selections, etc.

We introduce the new learning paradigm, \textbf{learning using annotation byproducts (LUAB)}, as a promising alternative to the usual supervised learning (Figure \ref{fig:luab}). We propose to use the annotation byproducts in the training phase, for further enhancing a model. This is a special case of learning using privileged information (LUPI) \cite{lupi}, where additional information $Z$ other than input $X$ and target $Y$ is available during training but is not given at inference. LUAB is an attractive instance of LUPI, as it does not incur additional annotation costs for privileged information.

We demonstrate the strength of the LUAB framework by contributing datasets \textbf{\ourin} and \textbf{\ourcoco}, where the original ImageNet and COCO classification training sets are enriched with the annotation byproducts. We show that annotation byproducts from image-category labelling interfaces contain weak information about the foreground object locations. We show that performing LUAB with such information improves not only generalisability but also robustness by reducing spurious correlations with background features, a critical issue of model reliability these days~\cite{shetty2019not,lapuschkin2019unmasking,geirhos2020shortcut}.

Our contributions are
(1) acknowledge a neglected information source available without additional costs during image labelling: annotation byproducts (\S\ref{sec:collecting-annotation-byproducts});
(2) LUAB as a new learning paradigm that makes use of annotation byproducts without extra annotation costs compared to the usual supervised learning (\S\ref{sec:exploiting-annotation-byproducts});
(3) empirical findings that LUAB with byproducts weakly encoding object locations improves model generalisability and reduces spurious correlations with the background (\S\ref{sec:experimental-results}); and
(4) release of \ourin and \ourcoco dataset for future research (\href{https://github.com/naver-ai/NeglectedFreeLunch}{github.com/naver-ai/NeglectedFreeLunch}).

\section{Related work}

We collect the annotation byproducts of the annotation process and exploit them for training models. We discuss three related fields of machine learning.

\subsection{Privileged learning}

\textbf{Privileged learning} \cite{vapnik_book_estimation,lupi_vapnik_teacher,lupi} refers to a machine learning scenario where the model is supervised not only with the directly task-relevant information (\eg image label $Y$) but also with auxiliary information called \textbf{privileged information} (PI) that is not available at inference.

Learning using privileged information (LUPI) was first studied in the context of classical machine learning algorithms such as support vector machines (SVM) \cite{lupi,lupi_rank,lupi_facial_feature,lupi_boosting,lupi_metric_learning,lupi_lampert}. 
LUPI has since been successfully applied to deep models with multitask learning framework where the PI is plugged in as auxiliary supervision \cite{lupi_modality_hallucination,lupi_multilabel_fcn,lupi_privileged_tasks}. PI may also be used as a representational bottleneck that regularises the cues for recognition \cite{lupi_group_orthogonal,lupi_dropout,concept_bottleneck}. ``Learning with rationale'' is an instance of LUPI actively being studied in natural language processing (NLP) domain \cite{learn_from_rationales,nlp_human_rationale_prior,classifiers_with_natural_language_explanations,compositional_explanation} with recent applications in computer vision problems \cite{robustness_via_human_annotation,generating_visual_explanations}.

Our learning setup, \textbf{learning using annotation byproducts (LUAB)}, is an instance of privileged learning with the annotation byproducts as the PI.
We hope that LUAB extends the LUPI paradigm by inviting creative methods for utilising the costless annotation byproducts.

\subsection{Collecting auxiliary signals from annotators}

It has been widely observed in the field of human-computer interaction that online annotators leave traces and logs that contain noisy yet important information \cite{nonreactive,instrumenting_the_crowd,workflow_mining}. There have been attempts in crowdsourcing image categories to record human gaze during task execution \cite{gaze_is_great_source,detector_from_gaze,gaze_for_image_description,help_captioning_with_gaze,gaze,human_attention_zeynep,gaze_based_nlp,khurana2023synthesizing}. Since gaze recording devices are costly and intrusive, proxy measurements such as mouse clicks and tracks \cite{whats_the_point,openimages_v6,OpenImagesPoints,using_localized_narrative} and partially visible images \cite{bubbles,bubbles2,bubbles3,linsley2017visual,clickme,fel2022aligning} have also been considered.
Other works measure the annotators' response time as a proxy for the sample difficulty \cite{vitto_image_difficulty,predicting_human_complexity,dulay2022using}. Others have treated the degree of annotator disagreement as the level of difficulty or uncertainty for the sample \cite{classification_with_disagreements,cifarh}. Finally, there exist research topics on estimating the annotators' skills and expertise to reflect them in the training phase \cite{worker_skill_image_difficulty,worker_uncertainty,perceptual_annotation,expertise_estimation,worker_model}. 
In our work, we collect similar signals from annotators, such as mouse signals and interactions with various front-end components. However, our work is the first attempt to collect them at a million scale (\eg ImageNet) that are freely available as byproducts from the original annotation task.

One of the byproducts we collect, namely the click locations during ImageNet annotations, is similar to the ``point supervision'' considered in some previous work in weakly-supervised computer vision tasks\cite{whats_the_point,ren2020ufo,OpenImagesPoints}. While the data format (a single coordinate on an image) is similar, those works are \textit{not directly comparable}. Our click locations are \textit{cost-free byproducts} of the original ImageNet annotation procedure that arises \textit{inevitably} from the annotators' selection of images, while the point supervision requires a dedicated annotation procedure and incurs extra annotation costs.

\subsection{Robustness to spurious correlations}

Many datasets used for training machine learning models are reported to contain spurious correlations that let the model solve the problem in unintended ways \cite{shetty2019not,lapuschkin2019unmasking,cadene2019rubi,bahng2019rebias,geirhos2020shortcut,d2020underspecification,mimetics2021}. 
The presence of such shortcuts is measured through ``stress tests'' \cite{d2020underspecification}: the model is evaluated against a data distribution where the spurious correlations have been altered or eliminated. We take this approach in \S\ref{sec:experimental-results} to measure improvements in robustness due to LUAB.

Prior approaches to enhance the robustness to spurious correlations have utilised \textit{additional human supervision} to further specify the ``correct'' correlations models must exploit. For example, \cite{RRR2017,simpson2019gradmask,chefer2022optimizing,gao2022aligning,petryk2022guiding,pahde2023reveal,rao2023using} regularise the attention maps of image classifiers with respect to various forms of human guidance, such as bounding boxes, segmentation masks, human gaze, and language, to let the classifiers focus on the actual object regions. 
In this work, we use signals that are \emph{unintentionally} generated by humans during widely-used image annotation procedures to enhance the robustness to spurious correlations. Those signals are available \textit{at no extra cost} during the annotation.

\section{Collecting annotation byproducts}
\label{sec:collecting-annotation-byproducts}

To construct a comprehensive package of annotation byproducts, we replicate the annotation procedure for two representative image classification datasets, ImageNet \cite{imagenet}, and COCO \cite{coco}. Resulting datasets with annotation byproducts, \ourin and \ourcoco, will be published.

\subsection{Browsing versus tagging interfaces}

There are two widely-used interfaces for annotating image labels: \textbf{browsing} (\eg, ImageNet) and \textbf{tagging} (\eg, COCO). A browsing interface presents a single concept along with a set of candidate images arranged in a grid and asks the annotator to select the images correctly depicting the concept. A tagging interface presents a single image at a time and asks the annotator to choose one or more objects and concept labels as necessary (survey of interfaces in \cite{annotation_ui_survey}).

The two paradigms have different strengths. Browsing is advantageous for efficient batch processing of images, where the annotation precision matters less. Tagging is helpful for careful labelling and supports the annotation of multiple labels per image. Browsing interfaces have been used for the ImageNet \cite{imagenet,imagenet_v2}, Places \cite{places}, and CUB \cite{cub} datasets. Tagging interfaces have been used for Pascal \cite{pascal}, COCO \cite{coco}, LVIS \cite{lvis}, and iNaturalist \cite{inaturalist}. As representatives of each type, we replicate ImageNet \cite{imagenet,imagenet_v2} and COCO \cite{coco}.

\subsection{ImageNet}
\label{subsec:byproducts-imagenet}

ImageNet \cite{imagenet} is a single-label dataset annotated via browsing. We describe how we replicated the original annotation procedure and present the set of annotation byproducts collected through the browsing annotation. %

\subsubsection{Replicating ImageNet annotations}

We replicate the annotation process for the training split of ImageNet1K (1,281,167 images). 
The original annotation procedure consists of the following four stages \cite{imagenet,imagenet_v2}. (1) Construct the list of classes $\mathcal{C}$ to annotate. (2) Crawl candidate images $I_c^\text{cand}$ for each class $c\in\mathcal{C}$ from the web. (3) Crowdsourced annotators select true images $I_c^\text{select}$ of class $c$. (4) Expert annotators clean up the dataset.

We replicate only the crowdsourcing stages (2) and (3) that are directly related to the generation of annotation byproducts.  Our replication is based on the description in the original ImageNet \cite{imagenet} and ImageNetV2 \cite{imagenet_v2} papers. For stage (1), we use the 1,000-class subset of the original 21,841 WordNet concepts \cite{wordnet}, corresponding to the ILSVRC2012 subset, also known as the ImageNet1K \cite{imagenet}.

\noindent
\textbf{Preparing candidate images $I_c^\text{cand}$ for each class $c\in\mathcal{C}$.}
The candidate images for the original dataset are crawled from Google, MSN, Yahoo, and Flickr \cite{imagenet_v2}. The search keywords are formulated by combining the class names and their ``synsets'' in WordNet \cite{wordnet}. The resulting set of images $I_c^\text{cand}$ becomes the candidate image set for class $c$. The annotators later select a subset $I_c^\text{select}\subset I_c^\text{cand}$ to finalise the set of images that contain the class $c$. 
Our aim is to collect the annotation byproducts for the 1,281,167 original training images of ImageNet1K. We thus let the annotators select the final images from a mixture of the original training images $I_c^\text{imagenet}$ and the set of new candidate images from Flickr $I_c^\text{flickr}$ \cite{flickr}. We set the ratio between the original ImageNet and Flickr-sourced images as 1:3. Our candidate set for each class $c$ is $I_c^\text{cand}=I_c^\text{imagenet}\cup I_c^\text{flickr}$. Then the annotators select the images containing $c$, $I_c^\text{select}\subset I_c^\text{cand}$, where the hope is that $I_c^\text{select}$ contains many original ImageNet samples $I_c^\text{imagenet}$. We report 86.7\% of $I_c^\text{imagenet}$ have been selected as a result. A 100\% recall is conceptually impossible due to boundary cases and label noises in $I_c^\text{imagenet}$ \cite{are_we_done,evaluating_machine_accuracy}.

\begin{figure}
    \centering
    \includegraphics[width=.999\linewidth]{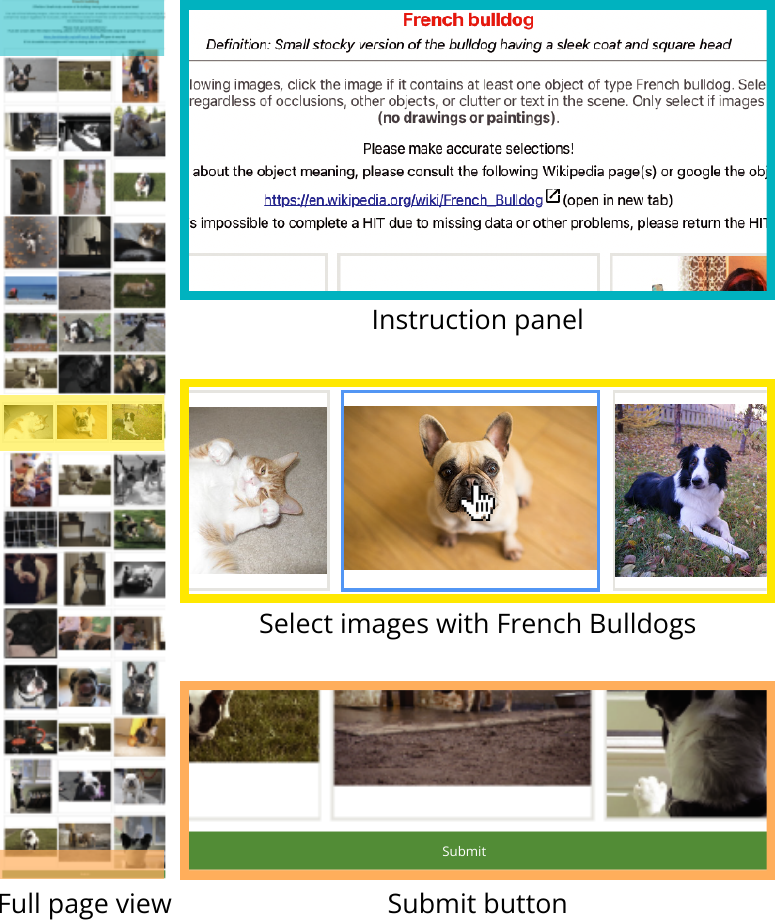}
    \vspace{-2em}
    \caption{\small\textbf{ImageNet annotation interface.} We replicate the interface in \cite{imagenet_v2_arxiv}. Annotators read the category description in the instruction panel, select all the images corresponding to ``French bulldog'', and click on the submit button.
    }
    \label{fig:imagenet_annotation_interface}
    \vspace{0em}
    \includegraphics[width=\linewidth]{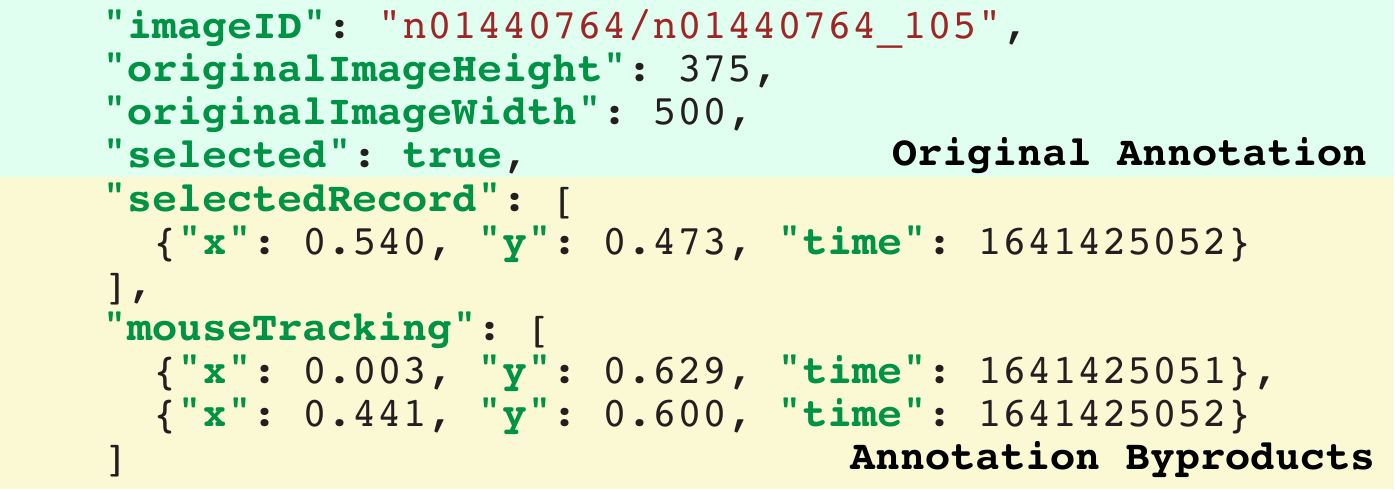}
    \vspace{-2em}
    \caption{\small\textbf{Annotation byproducts from ImageNet.}
    See Appendix Figure %
    \ref{appendixfig:annotation_byproduct_imagenet_sample} 
    for the full list of byproducts.
    }
    \label{fig:annotation_byproduct_imagenet_sample}
    \vspace{-.5em}
\end{figure}

\noindent
\textbf{Crowdsourced annotation via browsing interface.}
Following the original procedure, we let the Amazon Mechanical Turk (MTurk) \cite{mturk} workers complete the selection process $I_c^\text{select}\subset I_c^\text{cand}$ for each class $c$. ImageNet and ImageNetV2 interfaces are shown in Figures 9 and 10 of the ImageNetV2 paper on arXiv \cite{imagenet_v2_arxiv}, respectively. We closely follow the ImageNetV2 interface because the documentation is richer. Our interface is shown in Figure \ref{fig:imagenet_annotation_interface}. Like ImageNetV2, we show 48 candidate images $I_c^\text{cand}$ for a single class $c$ for each task. MTurk annotators click on images containing class $c$ and submit the selections $I_c^\text{select}$. Importantly, we have designed the front-end and back-end to record and save the annotation byproducts in the database.  
The annotation interface and crowdsourcing details are explained in Appendix \ref{appendix:crowdsourcing_details_imagenet}.

\noindent
\textbf{Number of annotators per image.}
The original ImageNet annotation procedure presents each image to 10 annotators for more precise annotations. This would require 240k USD for the annotation. Given the budget constraint, we have collected 1 annotation per image,  spending 24k USD instead. The utility of annotation byproducts demonstrated in \S\ref{sec:experimental-results} is thus \textit{a lower bound on the actual utility}.

\subsubsection{ImageNet byproducts}
\label{subsubsec:byproducts_statistics_analysis}

We show the annotation interface for ImageNet in Figure \ref{fig:imagenet_annotation_interface}. In the ImageNet annotation procedure, annotators click on the images containing the concept of interest. In the process, they leave the time-series of mouse positions (\textttsm{mouseTracking}) and mouse click events (\textttsm{selectedRecord}). The original annotation has not recorded them and only saved whether or not each image is finally selected. During our replicated annotation, we saved them in the database.
We show the list of annotation byproducts in Figure \ref{fig:annotation_byproduct_imagenet_sample}.

Among 1,281,167 ImageNet1K training images, annotators re-selected 1,110,786 (86.7\%) and interacted with 1,272,225 (99.3\%) images, leaving annotation byproducts.

\subsection{COCO}
\label{subsec:byproducts-coco}

COCO \cite{coco} is a multi-label dataset annotated with a tagging interface. We describe the creation of \ourcoco. We present and analyse the annotation byproducts for COCO.

\subsubsection{Replicating COCO annotations}

We replicate annotations for the 82,783 training images of COCO 2014 to collect the annotation byproducts. The original annotation procedure for COCO \cite{coco} consists of four stages. (1) Construct a list of classes to annotate. (2) Crawl and select candidate images from Flickr with more emphasis on images with multiple objects in context. (3) For each image, let crowdsourced annotators put all valid category labels. (4) Expert annotators do a final check-up. 

\begin{figure}
    \centering
    \includegraphics[width=\linewidth]{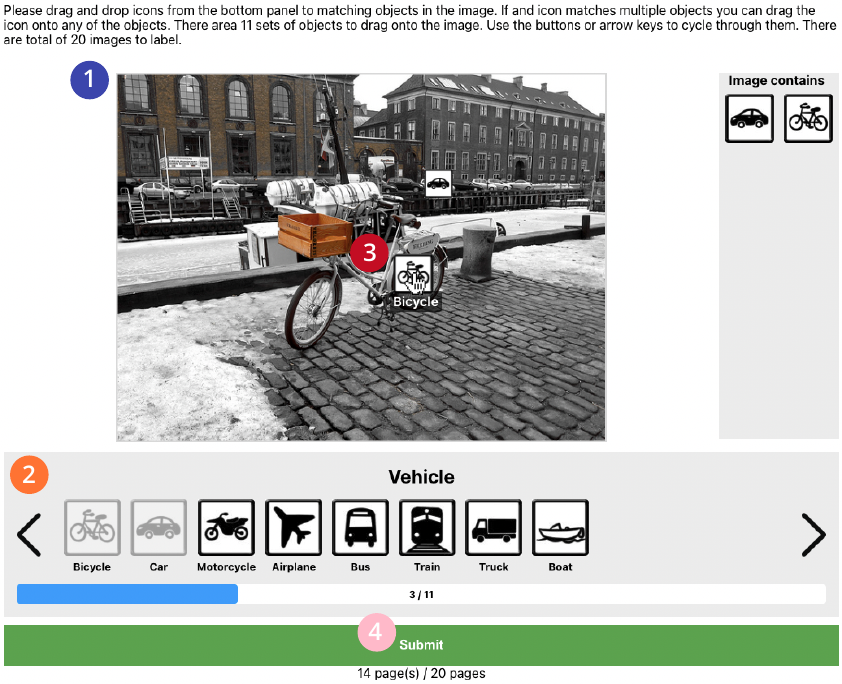}
    \vspace{-2em}
    \caption{\small\textbf{COCO annotation interface.} \includegraphics[width=.9em]{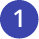} Annotator works on a single image at a time. \includegraphics[width=.9em]{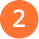} Find the classes present in the image by navigating superclasses. \includegraphics[width=.9em]{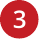} Drag and drop class icons on the objects in the image. \includegraphics[width=.9em]{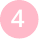} When finished, click on the submit button.}
    \label{fig:coco_annotation_interface}
  \vspace{1em}
    \includegraphics[width=\linewidth]{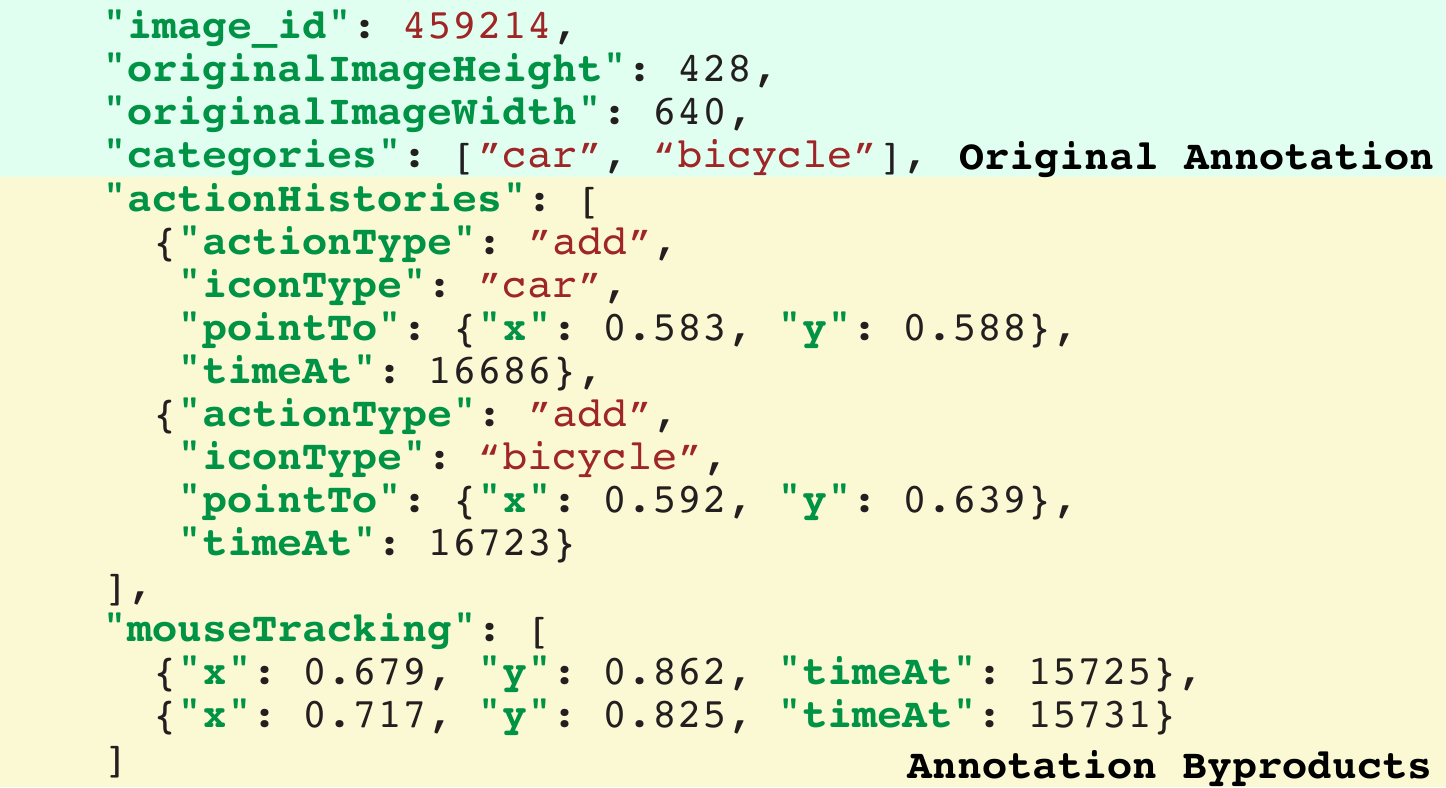}
    \vspace{-1.5em}
    \caption{\small\textbf{Annotation byproducts from COCO.} 
    See Appendix Figure %
    \ref{appendixfig:annotation_byproduct_coco_sample}
    for the full list of byproducts.
    }
    \label{fig:annotation_byproduct_coco_sample}
    \vspace{-1em}
\end{figure}

We only replicate stage (3), which produces direct annotation byproducts, by letting annotators work on the 82,783 training images.
Figure \ref{fig:coco_annotation_interface} shows the COCO annotation interface. We replicate the front-end of the original \cite{coco} (Figure 12a). For every image presented, the annotator must identify as many classes present as possible and place the corresponding class icons on the objects. We have replicated the superclass-browsing interface in \cite{coco} that lets annotators efficiently search through 80 COCO classes via 11 superclasses. The icon can be placed only once on an image per class. That is, even when there are multiple instances of a class, annotators should choose one of them to place the icon on. This is the same in the original COCO interface.
Crowdsourcing details are in Appendix %
\ref{appendix:crowdsourcing_details_coco}.

\subsubsection{COCO Byproducts}

COCO interface (Figure \ref{fig:coco_annotation_interface}) has two main components: (1) the image on which the class icons are placed and (2) the class browsing tool showing the class icons. The annotation byproducts come from these two sources. See Figure \ref{fig:annotation_byproduct_coco_sample} for the full list of annotation byproducts. 

The \textttsm{actionHistories} field describes the actions performed with the mouse cursor on the image. It lists the sequence of actions with possible types \textttsm{add}, \textttsm{move}, \textttsm{remove} and the corresponding location, time, and the category label of the icon. The \textttsm{mouseTracking} field records the movement of the mouse cursor over the image.

Annotators have reannotated 82,765 (99.98\%) of the 82,783 training images. 
We found that only 61.9\% of the class occurrences are retrieved on average. This confirms the findings in Lin~\etal \cite{coco} that the recall rate is low for multi-label annotation tasks and multiple annotators are necessary for every image. While desirable, collecting 10 annotations per image requires 100k USD, beyond our budget. We have instead assigned one annotator per image, spending 10k USD. Our setup presents a lower bound on the actual utility of the original annotation byproducts.

Finally, we emphasise those localisation byproducts are indeed general annotation byproducts for class labelling with a tagging interface. For example, Objects365 classes are obtained by labelling the 365 classes \textit{along with instance bounding boxes} (\S3.2.1 in \cite{shao2019objects365}). Class labels in LVIS are collected \textit{along with corresponding positions}, as in COCO (\S3.1 in \cite{lvis}). Location marking is often inseparable from multi-label annotations. Without any indication of \textit{where}, subsequent quality control stages are highly inefficient. Suppose an annotator labels ``chopsticks'' in a cluttered kitchen photo. It will be challenging to quickly confirm if the label is correct without knowing \textit{where}.
\section{Learning using annotation byproducts}
\label{sec:exploiting-annotation-byproducts}

We introduce the paradigm of \textbf{learning using annotation byproducts (LUAB)}. Compared to conventional supervised learning, we train models with additional annotation byproducts that have previously not been utilised in model training.

\subsection{LUAB with weak localisation signals}
\label{subsec:luab-with-localisation}

Annotation byproducts contain rich information surrounding the input image and the cognitive process of the annotator executing the task. In this work, we focus on the byproducts related to \textbf{object locations}, such as the click locations on images. We expect them to provide the model with a weak signal on the actual foreground pixels of the objects. Albeit weak, we expect them to be helpful information for resolving spurious correlations with background features, a common phenomenon in vision datasets \cite{bg_challenge,shetty2019not}.

\noindent
\textbf{Annotation byproducts encoding object locations.}
We hypothesise that the record of human interaction with the image annotation interfaces provides weak signals for the object locations. For ImageNet (\S\ref{subsec:byproducts-imagenet}), we consider the final click coordinates for every selected image (\textttsm{selectedRecord}). For COCO (\S\ref{subsec:byproducts-coco}), we consider the coordinates of the final \textttsm{add} action of a class icon on the image (\textttsm{actionHistories}). We treat them as proxy, cost-free data for object locations for each image. We note that such points on objects provide rich information about the foreground locations \cite{whats_the_point,OpenImagesPoints}.

\begin{figure}[t]
    \centering
    \small
    \includegraphics[width=1.0\linewidth]{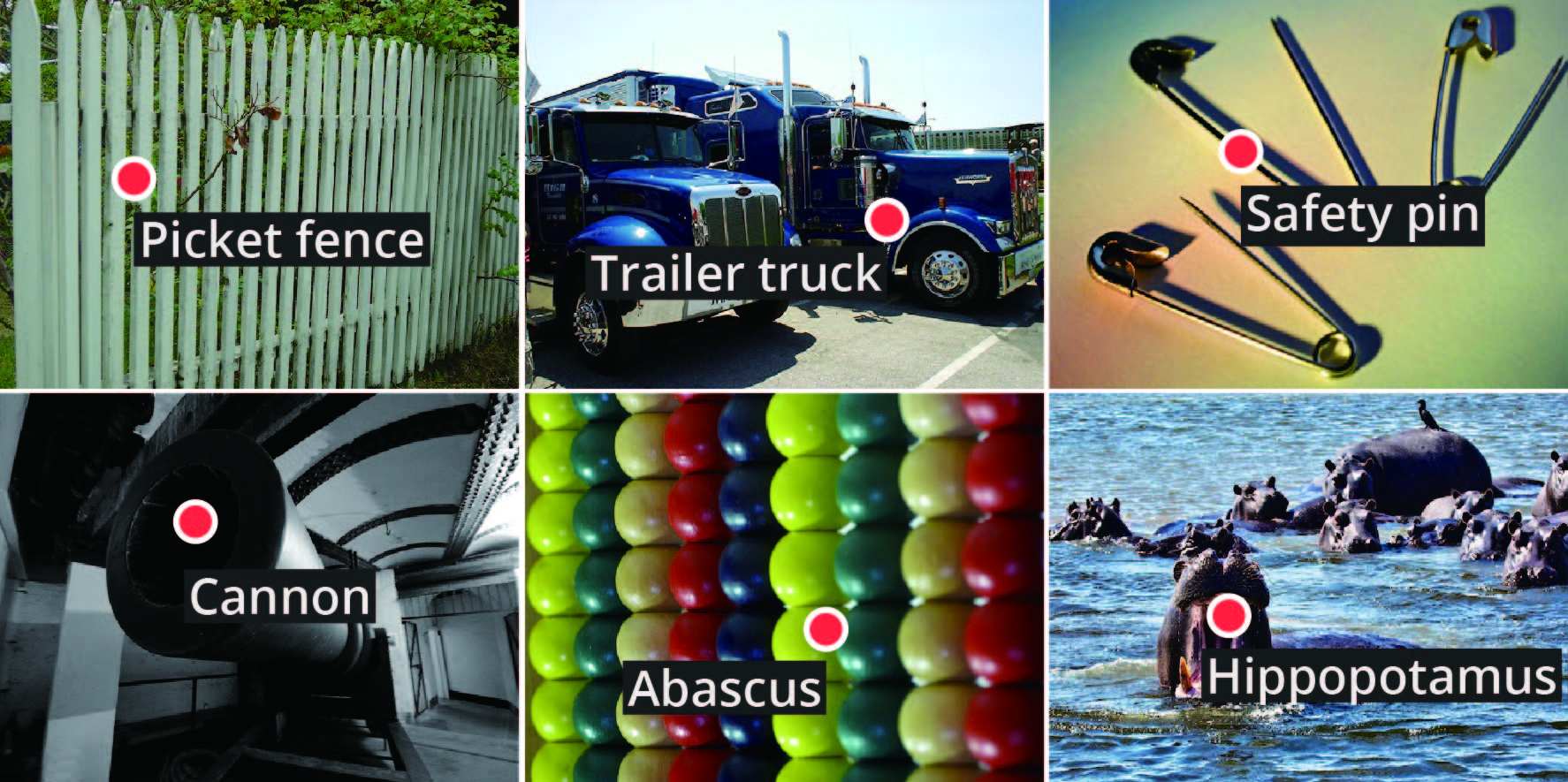}
    \vspace{-2em}
    \caption{\small {\bf ImageNet final clicks}. We visualise random training images; {\color{myred} \textbf{points}} are the final click positions in \textttsm{selectedRecord}.}
    \label{fig:data_vis}
    \vspace{-1em}
\end{figure}

\noindent
\textbf{Precision of object localisation in annotation byproducts.}
We verify the localisation accuracy of the annotation byproducts mentioned above. For ImageNet, we consider the subset of training data with both (1) our annotation byproducts (87\%) and (2) ground-truth boxes provided by the original dataset (42\%). We use the boxes to measure click accuracy. This gives 82.9\% accuracy. Qualitative examples are in Figure \ref{fig:data_vis}. For COCO, we use the ground-truth pixel-wise masks for measuring the precision of icon placements (\#\ignorespaces correct placement/\#\ignorespaces all placements). This gives 92.3\% precision. Therefore, we confirm that the respective annotation byproducts are fairly precise proxies for the actual foreground pixels. 
See Appendix %
\ref{appendix:analysis_annotation_byproducts} 
for more analysis.

\noindent
\textbf{Other annotation byproducts from class labelling.}
We conjecture that one may obtain an estimate for the extent of objects by taking the convex hull of a few mouse trajectory points before and after the click or icon placement. In addition to localisation, annotation byproducts may provide proxy signals on sample-wise difficulty through the completion time \cite{vitto_image_difficulty}. There also exists rich cross-sample association information: where two samples are annotated by the same annotator or on the same front-end page. Such information may help reduce annotator biases \cite{geva2019we}. They are beyond the scope of our paper, but we discuss the possibilities in Appendix %
\S\ref{appendix:byproducts_imagenet}.

\noindent
\textbf{Annotation byproducts beyond class labelling.}
Polygonal instance segmentation \cite{coco} results in byproducts like the order of clicks and the history of
corrections. In the language domain, one may not only record human text answers but the history of corrections in the answer, where we hypothesise that more corrections signify more ambiguity.

\subsection{Multi-task learning baseline for LUAB}
\label{subsec:regression}

The usual ingredients for the supervised learning of image classifiers are image-label pairs $(X,Y)$. Our LUAB framework introduces a third ingredient, weak object location $Z$, for every image $X$. For single-class datasets like ImageNet, the coordinates are given as $Z\in[0,1]\times[0,1]$, a relative position in each image. For multi-class datasets like COCO, this is given as $Z_c \in [0,1]\times[0,1]$ for every class $c$ present in the image.

We propose a simple baseline based on a \textbf{multi-task objective} for the classification of $Y$ and the regression of $Z$. We expect that learning the localisation would condition the network to select features more from foreground object regions \cite{zamir2018taskonomy,lu202012,fifty2021efficiently}.

We write the original network architecture as $g(f(X))$, where $f$ is a feature extractor, and $g$ is a classifier that maps intermediate features to $\mathbb{R}^C$. The regression objective is applied to $h(f(X))$ where $h$ maps the intermediate features to normalised x-y coordinates in $[0,1]\times[0,1]$. For a single-class classification task (\eg ImageNet), the objective is
\begin{align}
\vspace{-1em}
    \min_{f,g,h}\, \mathcal{L}\left(g(f(X)),Y\right)+\lambda||h(f(X))-Z||_{s1},
    \label{eq:main-objective}
    \vspace{-1em}
\end{align}
where $\mathcal{L}$ is the cross-entropy loss and $||\cdot||_{s1}$ is the smooth-$\ell^1$ loss \cite{girshick2015fast}. $\lambda>0$ regulates the weight of the regression term. The objective is identical for the multi-class classification (\eg COCO), except that $\mathcal{L}$ is a binary cross-entropy loss and the regression target is the mean of smooth-$\ell^1$ losses for every class present in the image.
We use the task labels $Y$ from the original datasets for both ImageNet and COCO experiments. The regression term is applied only for samples for which $Z$ is available.

\noindent
\textbf{Discussion.}
We show the minimal utility of the annotation byproducts by considering a simple baseline. We note that one may explore more advanced training schemes like regularising the model's attribution map with $Z$ \cite{RRR2017,simpson2019gradmask,chefer2022optimizing} or forcing the model to pool features with attention $Z$ \cite{lupi_group_orthogonal}. We explore the latter method in Appendix %
\S\ref{appendix:further_experimental_results}.

\section{Experimental results}
\label{sec:experimental-results}

We show the empirical efficacy of \textbf{learning using annotation byproducts (\ours)} that weakly encode object locations. We verify whether the annotation byproducts improve the original image classification performance and robustness by guiding models to focus more on foreground features.

\begin{table*}[!t]
    \small
    \centering
    \tabcolsep=0.06cm
    \begin{tabular}{l|r|ccc|ccc|cc|cc|ccc|cc}
        \toprule
        \footnotesize Model & \footnotesize Params & \footnotesize IN-1k↑ & \footnotesize IN-V2↑ & \footnotesize IN-Real↑ & \footnotesize IN-A↑ & \footnotesize IN-C↑ & \footnotesize IN-O↑ & \footnotesize Sketch↑ & \footnotesize IN-R↑ & \footnotesize Cocc↑ & \footnotesize ObjNet↑ & \footnotesize SI-size↑ & \footnotesize SI-loc↑ & \footnotesize SI-rot↑ & \footnotesize BGC-gap↓ & \footnotesize BGC-acc↑ \\
        \midrule
        R18 & 11.7M & 72.1 & 59.9 & \bf 79.6 & \bf 2.0 & 37.4 & 52.7 & \bf 22.0 & 34.0 & 41.9 & 21.7 & 46.4 & 22.9 & 32.1 & 9.0 & \bf 22.1 \\ 
        +\ours & 11.7M & \bf 72.2 & \bf 59.9 & 79.6 & 1.9 & \bf 37.6 & \bf 53.0 & 21.6 & \bf 34.3 & \bf 44.7 & \bf 21.9 & \bf 47.8 & \bf 23.1 & \bf 32.7 & \bf 8.6 & 20.4 \\
        \midrule
        R50 & 25.6M & 77.4 & 65.2 & 83.5 & \bf 5.5 & 43.8 & 56.7 & 25.4 & 37.8 & 53.7 & 27.8 & 53.9 & 31.9 & 40.1 & 6.3 & 26.7 \\ 
        +\ours & 25.6M & \bf 77.5 & \bf 65.2 & \bf 83.8 & 5.1 & \bf 44.7 & \bf 57.0 & \bf 25.7 & \bf 38.2 & \bf 55.1 & \bf 28.5 & \bf 55.6 & \bf 33.5 & \bf 40.9 & \bf 5.6 & \bf 27.4 \\
        \midrule
        R101 & 44.5M & 78.2 & 66.0 & 84.1 & 7.6 & 47.0 & \bf 60.7 & 26.5 & 38.2 & 55.8 & 29.4 & 53.4 & 33.1 & 38.9 & 5.6 & \bf 30.2 \\ 
        +\ours & 44.5M & \bf 78.6 & \bf 66.4 & \bf 84.3 & \bf 7.8 & \bf 47.9 & 60.5 & \bf 27.0 & \bf 39.0 & \bf 58.5 & \bf 30.0 & \bf 54.4 & \bf 33.3 & \bf 39.8 & \bf 5.5 & 28.2 \\
        \midrule
        R152 & 60.2M & 79.0 & 67.2 & 84.5 & 9.5 & 49.5 & 62.0 & 27.6 & 39.6 & 58.8 & 30.5 & 53.9 & 33.3 & 38.6 & 6.6 & 27.2 \\ 
        +\ours & 60.2M & \bf 79.2 & \bf 67.2 & \bf 84.8 & \bf 9.5 & \bf 49.9 & \bf 62.1 & \bf 27.6 & \bf 39.7 & \bf 59.0 & \bf 31.3 & \bf 55.5 & \bf 34.2 & \bf 40.6 & \bf 5.8 & \bf 31.6 \\
        \midrule
        ViT-Ti & 5.7M & 72.8 & 60.7 & 80.7 & 7.9 & \bf 48.5 & 52.3 & 20.5 & 32.8 & 63.8 & 23.1 & 46.3 & 23.8 & 33.9 & 8.2 & 13.9 \\
        +\ours & 5.7M & \bf 72.9 & \bf 60.8 & \bf 80.9 & \bf 8.4 & 48.4 & \bf 52.9 & \bf 21.1 & \bf 33.8 & \bf 64.2 & \bf 23.7 & \bf 47.4 &\bf  25.4 & \bf 34.7 & \bf 7.8 & \bf 14.4 \\ 
        \midrule
        ViT-S & 22.1M & 80.3 & 69.1 & 86.0 & 20.0 & 60.3 & 53.4 & 29.4 & 42.3 & 73.8 & 31.2 & 54.5 & 32.0 & 39.5 & 6.4 & 17.4 \\ 
        +\ours & 22.1M & \bf 80.6 & \bf 69.7 & \bf 86.4 & \bf 22.8 & \bf 61.2 & \bf 55.1 & \bf 30.0 & \bf 43.0 & \bf 74.1 & \bf 32.3 & \bf 55.1 & \bf 33.7 & \bf 39.6 & \bf 5.9 & \bf 18.7 \\
        \midrule
        ViT-B & 86.6M & 81.6 & 70.3 & 86.6 & 26.1 & 64.1 & 58.0 & 33.0 & 45.7 & 76.0 & 31.7 & 56.6 & 35.1 & 41.3 & 6.4 & 18.1 \\
        +\ours & 86.6M & \bf 82.5 & \bf 71.9 & \bf 87.4 & \bf 31.1 & \bf 66.0 & \bf 58.5 & \bf 35.5 & \bf 48.4 & \bf 77.5 & \bf 35.0 & \bf 57.1 & \bf 36.8 & \bf 41.6 & \bf 5.6 & \bf 23.9 \\
        \bottomrule
    \end{tabular}
    \vspace{-0.5em}
    \caption{\small {\bf Performance of \ours on ImageNet1K.}
    We report in-distribution generalisation metrics ({IN-1k/V2/Real}) and out-of-distribution metrics ({IN-A/C/O/R/Sketch/Cocc/ObjNet}). We also report metrics for detecting spurious correlations with background (SI-Score \cite{si_score} and BG-Challenge \cite{bg_challenge}).
    \ours training with annotation byproducts using a simple point regression target improves the overall performances.
    \ours barely introduces any extra annotation or computational cost.
    }
    \label{table:main-imagenet}
    \vspace{-1.5em}
\end{table*}
\subsection{Results on ImageNet}

\noindent
\textbf{Implementation details.}
We use the \ourin training set with annotation byproducts to train image classifiers. Considered backbones are ResNets~\cite{resnet} (ResNet18, ResNet50, ResNet101, and ResNet152), and Vision Transformers (ViT-Ti~\cite{deit}, ViT-S~\cite{deit}, and ViT-B~\cite{dosovitskiy2020image}). 
To accommodate the multi-task objective, we have attached a separate head for the regression target at the penultimate layer of each backbone.
This head is not used during the inference. We use the standard 100-epochs setup~\cite{resnet} for ResNets; the DeiT training setup\footnote{We train models with the official DeiT codebase~\cite{deit} with default settings for RandAug~\cite{cubuk2019randaugment}, Stochastic Depth~\cite{stochasticdepth}, Random Erasing~\cite{randomerasing, cutout}, Mixup~\cite{mixup}, Cutmix~\cite{cutmix}, and optimization setups -- AdamW~\cite{loshchilov2017decoupled} and cosine learning rate scheduling~\cite{loshchilov2016sgdr}, and gradual warmup~\cite{goyal2017accurate}.}~\cite{deit} is used for ViTs. This is to verify whether the annotation byproducts work together with the popular supervised training regimes.  We select the last-epoch models.
We further include results following the primitive setup~\cite{dosovitskiy2020image} in Appendix Table \ref{table:sub-imagenet}.

\noindent
\textbf{Evaluation.}
Along with the ImageNet1k validation set (IN-1k), we use many variants: ImageNet-{\small V2/Real/A/C/O/R/Sketch/ObjNet} \cite{imagenet_v2,are_we_done,imagenet_a_o,imagenet_c_p,imagenet_sketch,imagenet_r,objectnet}. In particular, we focus on the benchmarks designed to measure spurious correlations with the background cues: SI-Score \cite{si_score} and BG Challenge \cite{bg_challenge}. Both datasets de-correlate the foreground and background features by constructing novel images with foreground and background masks cut and pasted from different images.

\begin{figure}
\small
\centering
\hspace{-1mm}
\begin{subfigure}[t]{0.24\textwidth}
     \centering
     \includegraphics[width=1\linewidth]{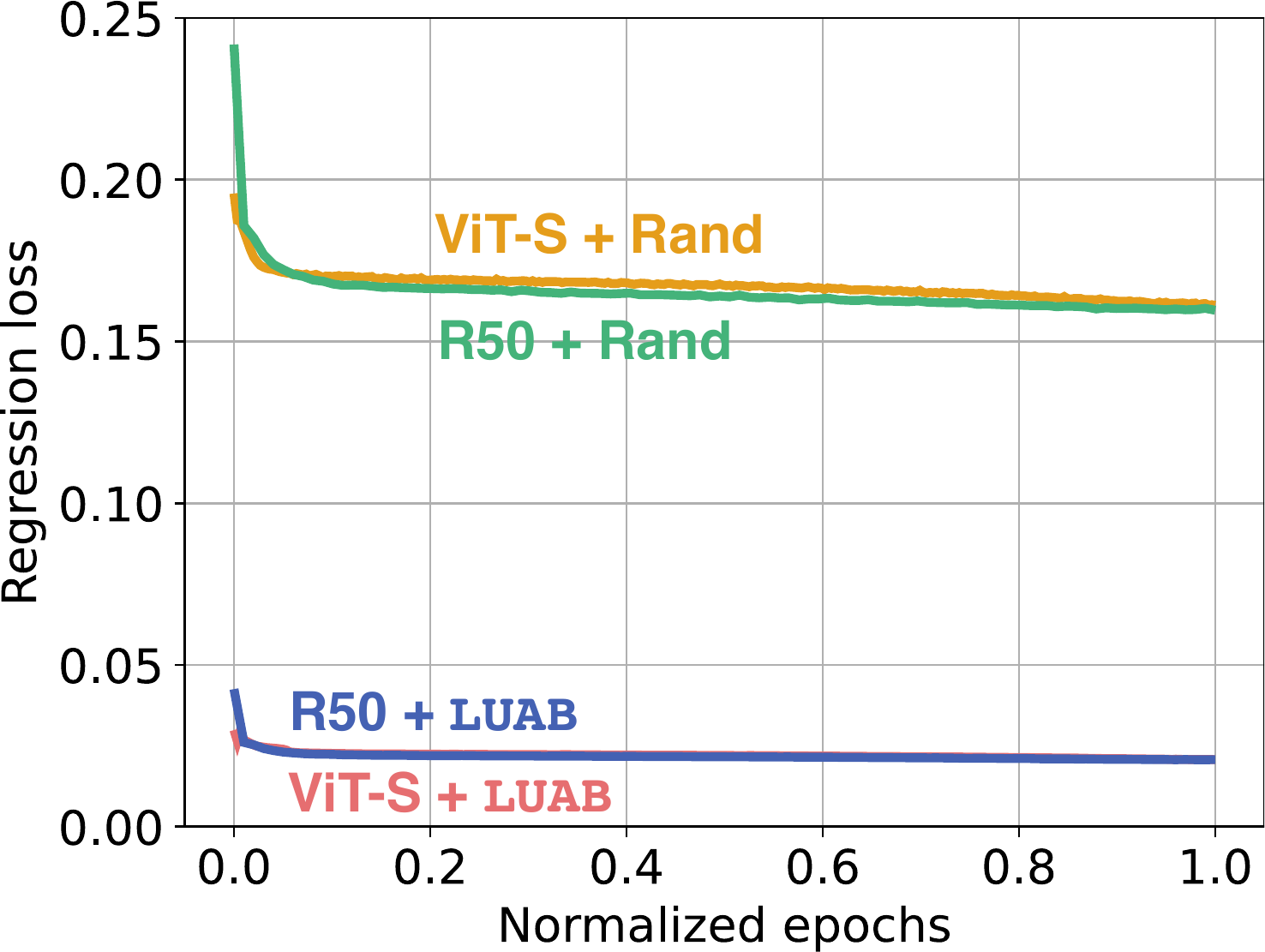}
     \caption{Training reg. loss}
\end{subfigure}
\begin{subfigure}[t]{0.232\textwidth}
     \centering
     \includegraphics[width=1\linewidth]{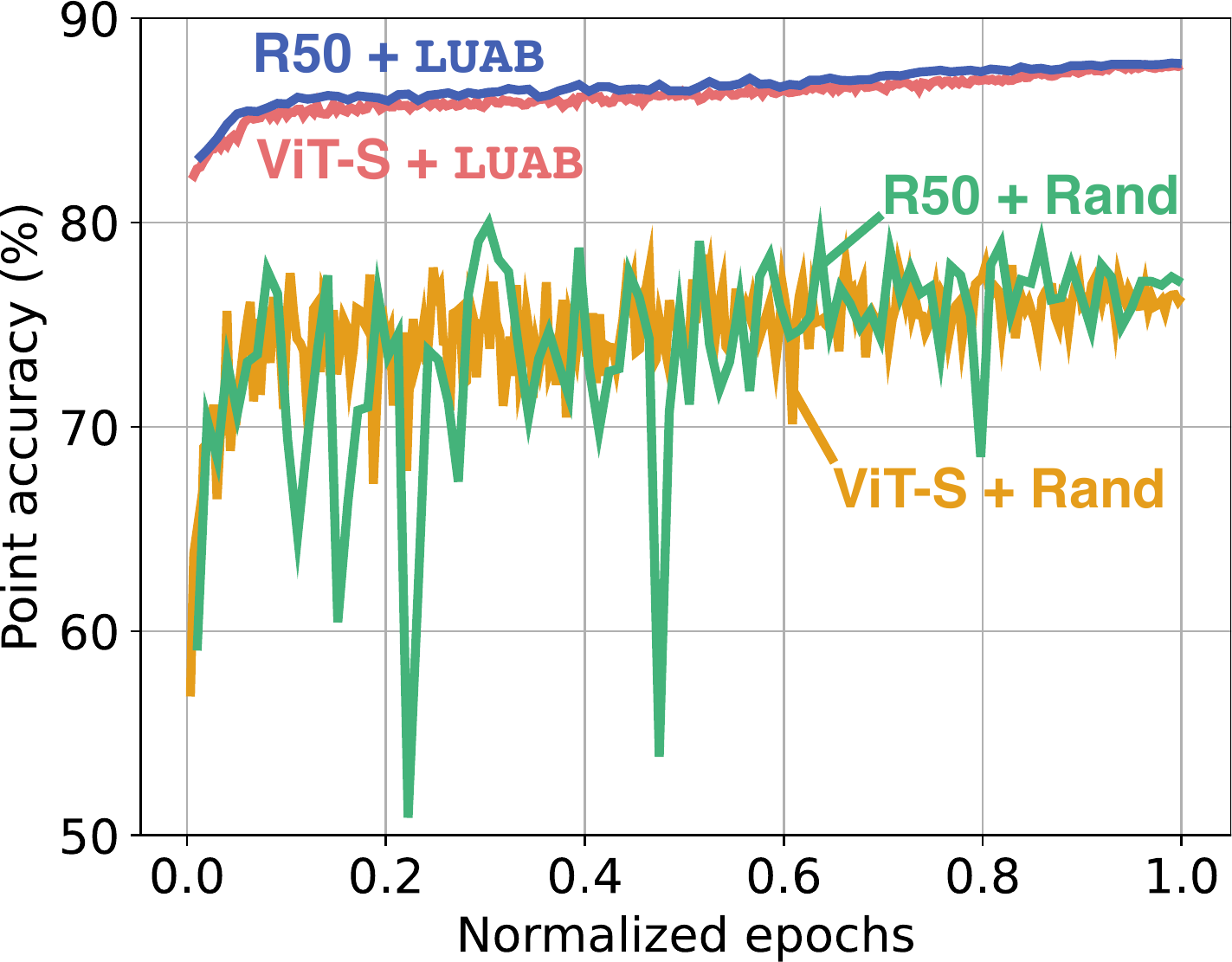}
     \caption{Validation loc. accuracy}
\end{subfigure}
\vspace{-1em}
\caption{\small {\bf Training curves for ImageNet}. 
``Rand'' refers to the regression with respect to a randomly generated location $Z$. 
}
\label{fig:training-curves}
\vspace{0.5em}
\small
\centering
\tabcolsep=0.04cm
\vspace{0em}
\begin{tabular}{l|c|c|c|ccc|cc}
\toprule
\scriptsize Model & \scriptsize Annot. & \scriptsize IN-1k↑ & \scriptsize ObjNet↑ & \scriptsize SI-size↑ & \scriptsize SI-loc↑ & \scriptsize SI-rot↑ & \scriptsize BGC-gap↓ & \scriptsize BGC-acc↑ \\
\midrule                        
R50 & - & 77.4 & 27.8 & 53.9 & 31.9 & 40.1 & 6.3 & 26.7 \\
R50 & Rand & 77.3 & 28.1 & 54.5 & 31.5 & 39.7 & 5.9 & \bf 27.6 \\ 
R50 & \ours & \bf 77.5 & \bf 28.5 & 5\bf 5.6 & \bf 33.5 & \bf 40.9 & \bf 5.6 & 27.4 \\
\midrule 
ViT-Ti & - & 71.8 & 20.1 & 40.6 & 16.5 & 26.2 & 12.1 & 13.6 \\ 
ViT-Ti & Rand & 72.2 & 22.0 & 42.5 & 18.1 & 27.5 & 11.0 & 15.3 \\
ViT-Ti & \ours & \bf 73.0 & \bf 22.1 & \bf 43.4 & \bf 20.0 & \bf 28.7 & \bf 10.9 & \bf 16.1 \\  
\midrule
ViT-S & - & 74.1 & 20.5 & 42.9 & 18.7 & 27.8 & 10.5 & 16.7 \\
ViT-S & Rand & 74.8 & 22.7 & 44.5 & 20.6 & 28.8 & 10.5 & 19.5 \\
ViT-S & \ours & \bf 75.3 & \bf 23.6 & \bf 47.8 & \bf 22.6 & \bf 32.2 & \bf 8.7 & \bf 19.7 \\
\bottomrule
\end{tabular}
\vspace{-1em}
\captionof{table}{\small {\bf Comparison with random point regression on ImageNet.} We compare the accuracies of supervised learning without additional supervision (``-''), with random points as guidance (``Rand''), and with our annotation byproducts (\ours). 
}
\label{table:random_point_exp}
\vspace{-2em}
\end{figure}

\noindent
\textbf{Random point baseline.}
We introduce a baseline trained with the same objective (Equation \ref{eq:main-objective}) but with a uniform-random point $Z$ for each image. This baseline helps us rule out possible regularisation effects due to the multi-task learning itself and focus purely on the information gain from the weak object locations given by the annotation byproducts.

\noindent
\textbf{\ours trains well.}
Figure \ref{fig:training-curves} shows the training curves. The regression loss for $Z$ decreases, and validation localisation accuracy increases for \ours over the epochs, while the baseline random-point supervision yields higher losses and lower localisation accuracies. The baseline performance is fairly high because of the object-centric ImageNet data. We confirm that the annotation byproducts contain localisation information that lets the model predict object locations.

\noindent
\textbf{\ours improves classification performance.}
See Table \ref{table:main-imagenet} for the IN-1k validation accuracies before and after \ours. We observe that \ours introduces gains across the board (\eg 81.6\% to 82.5\% for ViT-B). Similar gains are seen for IN-V2/Real. The \ours help the models generalise better.

\noindent
\textbf{\ours improves out-of-distribution (OOD) generalisation.}
Table \ref{table:main-imagenet} shows that \ours improves the OOD generalisation (columns for {\small IN-A/C/O/R/Sketch}). 30 of the 35 combinations (5 metrics $\times$ 7 models) have seen improvements due to \ours. We hypothesise that the focus on foreground features improves generalisation to novel distributions.

\noindent
\textbf{\ours reduces spurious correlations with the background.}
Table \ref{table:main-imagenet} also shows the results on metrics detecting spurious dependence on background features. For SI-Scores \cite{si_score}, we observe a clear advantage of \ours, beating the baseline performance in \textit{all} considered cases. For BG Challenge \cite{bg_challenge}, \ours surpasses the original models for the majority of cases (12 out of 14). The improvement due to \ours on the benchmarks with de-correlated foreground and background features demonstrates the efficacy of the foreground guidance from the annotation byproducts.

\noindent
\textbf{Improvement is not due to the multi-task objective itself.}
Table \ref{table:random_point_exp} shows greater improvements due to \ours compared to the random point baseline, which merely introduces a multi-task learning objective without additional location information. As such, we attribute the improvements to the weak foreground information in the annotation byproducts.

\begin{table}
\footnotesize
\centering
\vspace{-.5em}
\parbox{.27\linewidth}{
    \centering
    \setlength{\tabcolsep}{.5em}
    \begin{tabular}{l|c} %
    \toprule
    \footnotesize Annot. & \footnotesize Loc↑ \\
    \midrule                     
        R50 & 46.8 \\ 
        +\ours & \textbf{48.4} \\               
    \bottomrule
    \end{tabular}
    \vspace{-1em}
    \caption{\footnotesize {\bf WSOL on ImageNet} \cite{wsoleval}.}
    \label{table:imagenet_wsol}
    \vspace{1.7em}
}
\quad
\parbox{.62\linewidth}{
    \centering
    \setlength{\tabcolsep}{.5em}
    \begin{tabular}{l|c|c|c}
    \toprule
    \footnotesize Annot. & \footnotesize IN-1k↑ & \footnotesize Bbox AP↑ & \footnotesize Mask AP↑ \\ 
    \midrule
    R50  & 77.4 & 37.0 & 34.6\\
    +\ours  & \textbf{77.5} & \textbf{37.4} & \textbf{34.8} \\
    \bottomrule
    \end{tabular}
    \vspace{-1em}
    \caption{\small {\bf Fine-tuning ImageNet models on downstream tasks.} Object detection and instance segmentation.}
\label{table:coco_frcnn}
}
\vspace{-1.5em}
\end{table}

\noindent
\textbf{\ours lets models focus on foreground features.}
Class activation mapping (CAM) \cite{zhou2016learning} identifies the region-wise features that an image classifier uses to make the prediction. By using a weakly-supervised object localisation (WSOL) evaluation against the ground-truth object locations \cite{wsoleval}, one may confirm whether the utilised image features correspond to the object foreground. We show the results in Table~\ref{table:imagenet_wsol}. The 1.6\%p improvement in WSOL accuracy against the original shows that \ours lets the model focus on the foreground.

\noindent
\textbf{\ours improves downstream localisation tasks.} We report the box and mask APs on COCO {val2017} after fine-tuning the baseline ResNet50 and \ours-trained models for Faster-RCNN \cite{fasterrcnn} and Mask-RCNN \cite{maskrcnn}, respectively, in Table \ref{table:coco_frcnn}. \ours improves the downstream performances.

\subsection{Results on COCO}

\noindent
\textbf{Implementation details.}
We use the \ourcoco training set with annotation byproducts. Considered backbones are ResNet18/50/152~\cite{resnet}, and ViT-Ti/S/B~\cite{deit,dosovitskiy2020image}. We attach one regression head per class on the penultimate layer.
We follow the training recipe of the original papers. 
As in ImageNet, we consider the random point baseline: the localisation supervision $Z_c$ is given as a uniform-random point.

\begin{figure}[t]
\small
\centering
\hspace{-1mm}
\begin{subfigure}[t]{0.245\textwidth}
     \centering
     \includegraphics[width=1\linewidth]{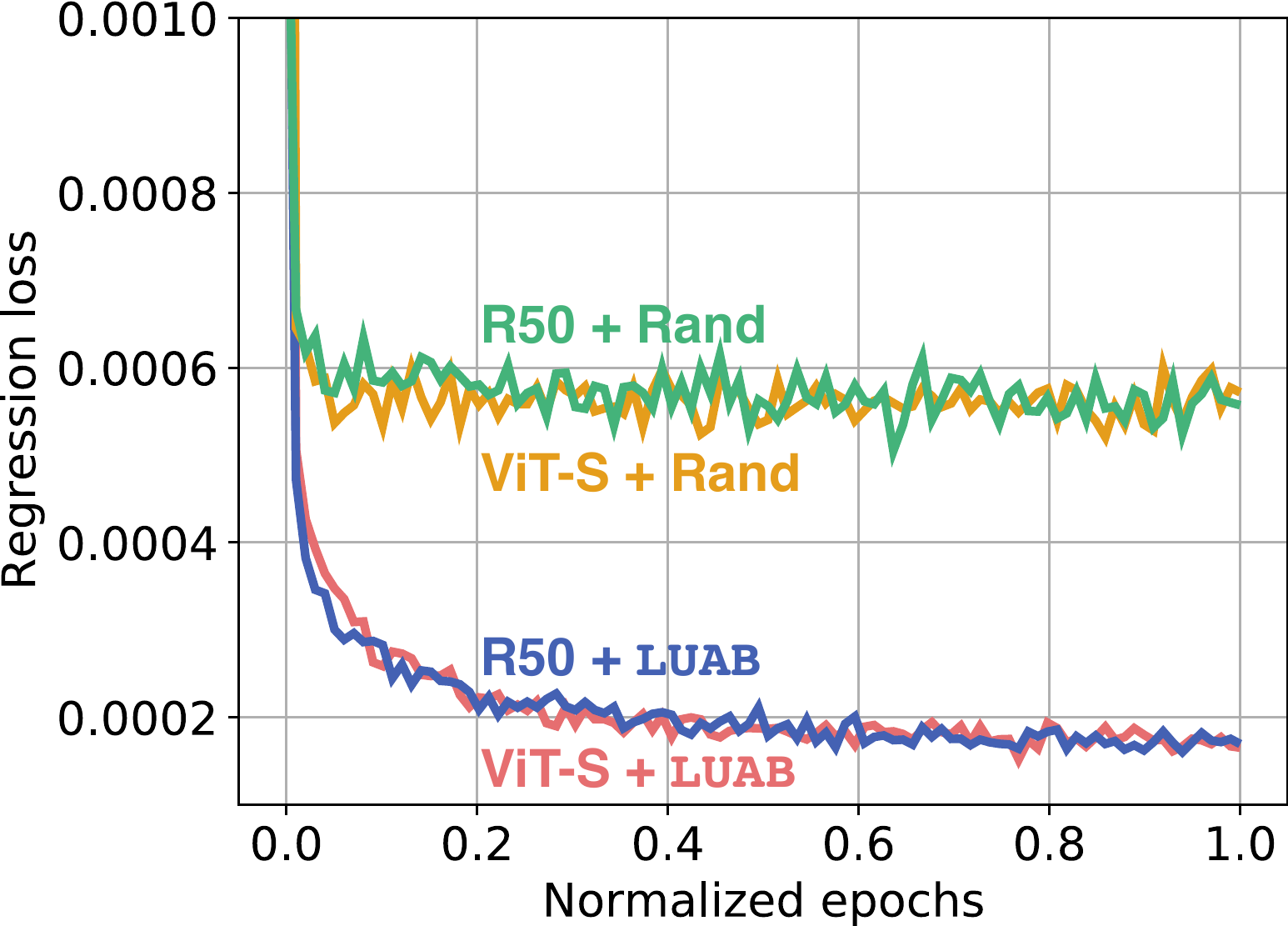}
     \caption{Training reg. loss}
\end{subfigure}
\begin{subfigure}[t]{0.228\textwidth}
     \centering
     \includegraphics[width=1\linewidth]{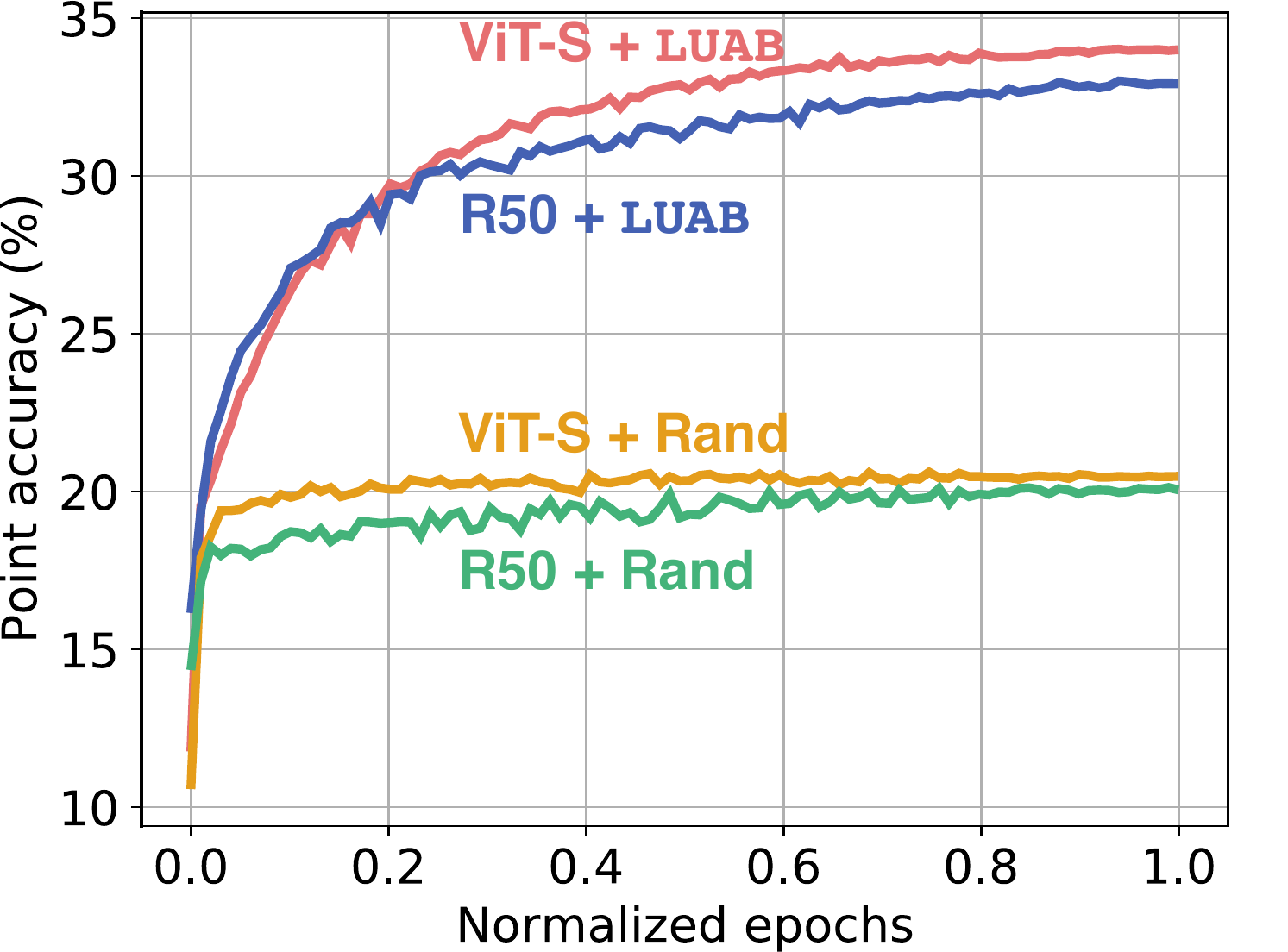}
     \caption{Validation loc. accuracy}
\end{subfigure}
\vspace{-1em}
\caption{\small {\bf Training curves for COCO}. 
``Rand'' refers to the regression with respect to randomly generated locations $Z_c$. 
}
\label{fig:training-curves-coco}
\vspace{-.5em}
\end{figure}

\noindent
\textbf{\ours trains well.}
Figure \ref{fig:training-curves-coco} shows the training curves for COCO with \ours. Compared to the random-point baseline, \ours decreases the regression loss and increases the validation localisation accuracy more quickly. We confirm: \ours confers the model information about where the objects are.

\noindent
\textbf{\ours improves classification performance.}
Table \ref{table:coco} and \ref{table:coco_vit} show that \ours improves the mean average precision (mAP), for example from 73.0\% to 74.2\% for ResNet50.

\noindent
\textbf{\ours reduces spurious correlations with other classes.}
We consider metrics for detecting a spurious dependence on frequently co-occurring objects (\eg monitor and keyboard). $V^\text{avg}$ and $V^\text{min}$~\cite{shetty2019not} compute the difference between the classification scores when class $c$ of interest is removed and when another class than $c$ are removed. $V^\text{avg}$ erases a random class, while $V^\text{min}$ erases the worst-case class for each image. Table \ref{table:coco} and \ref{table:coco_vit} show a consistent decrease in $V^\text{avg}$ and $V^\text{min}$ scores after \ours. This confirms the successful reduction in spurious background correlations via \ours.

\noindent
\textbf{\ours lets models focus on foreground features.}
 \setlength{\intextsep}{0.6em}%
\setlength{\columnsep}{0.7em}%
\begin{wraptable}{r}{.25\linewidth}
\vspace{-.5mm}
\setlength{\tabcolsep}{.2em}
\footnotesize
\centering
\subfloat{\begin{tabular}{l|c} %
\toprule
\footnotesize Annot. & \footnotesize mPxAP↑ \\
\midrule                     
    R50 & 20.8\\ 
    +\ours & \textbf{21.5} \\               
\bottomrule
\end{tabular}}
\vspace{-1em}
\caption{\footnotesize {\bf WSOL on COCO} \cite{wsoleval}.}
\label{table:coco_wsol}
\vspace{-0em}
\end{wraptable}%
As in ImageNet, we measure the CAM performances of the COCO-trained ResNet50 with and without \ours in Table \ref{table:coco_wsol}. 
We compute CAM for every class and report the class-averaged \mpxap~\cite{wsoleval}. We verify that the models attend more to the foreground features after training with \ours.

\begin{table}[t]
\small
\centering
\tabcolsep=.3em
\begin{tabular}{c|ccc|ccc|ccc} %
\toprule
{\footnotesize Model} & {\footnotesize R18} & {\footnotesize Rand} & {\footnotesize \ours} & {\footnotesize R50} & {\footnotesize Rand} & {\footnotesize \ours} & {\footnotesize R152} & {\footnotesize Rand} & {\footnotesize \ours} \\
\midrule
{\footnotesize mAP↑} & 67.9 & 67.8 & \textbf{68.0} & 73.0 & 73.6 & \textbf{74.2} & 73.3 & 74.6 & \textbf{75.4}\\
{\footnotesize $V^\text{min}$↓} & 51.8 & 52.1 & \textbf{51.6} & 47.6 & 47.3 & \textbf{47.0} & 47.4 & 47.8 & \textbf{47.1}\\
{\footnotesize $V^\text{avg}$↓} & 28.7 & 28.7 & \textbf{28.4} & 25.0 & 24.9 & \textbf{24.5} & 24.8 & 25.5 & \textbf{24.7}\\
\bottomrule
\end{tabular}
\vspace{-1em}
\caption{\small {\bf COCO Performance with ResNet.} We compare supervised learning, multi-task learning with random points, and \ours.}
\label{table:coco}
\vspace{-1.0em}
\end{table}

\begin{table}[t!]
\small
\centering
\tabcolsep=.25em
\begin{tabular}{c|ccc|ccc|ccc} %
\toprule
{\footnotesize Model} & {\footnotesize ViT-Ti} & {\footnotesize Rand} & {\footnotesize \ours} & {\footnotesize ViT-S} & {\footnotesize Rand} & {\footnotesize \ours} & {\footnotesize ViT-B} & {\footnotesize Rand} & {\footnotesize \ours} \\
\midrule
{\footnotesize mAP↑} & 72.6 & 72.2 & \textbf{72.7} & 76.2 & 76.9 & \textbf{77.3} & 76.4 & 74.5 & \textbf{77.5}\\
{\footnotesize $V^\text{min}$↓} & 49.1 & 48.9 & \textbf{48.4} & 47.1 & 46.9 & \textbf{45.8} & 46.6 & 47.1 & \textbf{45.6}\\
{\footnotesize $V^\text{avg}$↓} & 27.0 & 26.9 & \textbf{26.8} & 25.7 & 25.6 & \textbf{24.6} & 25.0 & 25.1 & \textbf{24.5}\\
\bottomrule
\end{tabular}
\vspace{-1em}
\caption{\small {\bf COCO Performance with ViT.} We compare supervised learning, multi-task learning with random points, and \ours.}
\label{table:coco_vit}
\vspace{-.5em}
\end{table}

\section{Conclusion}

We propose to log and exploit annotation byproducts that result from human interaction with input devices and various front-end components. We have created \textbf{\ourin} and \textbf{\ourcoco} by replicating the respective annotation procedures and logging \textbf{cost-free} annotation byproducts.
We have introduced a new learning paradigm: \textbf{learning using annotation byproducts (LUAB)}. As an example, we have used the final click and icon placement locations as proxies for the object locations. They let models generalise better and depend less on spurious background features. 

\noindent
\textbf{Limitations.}
We have performed only one annotation pass through ImageNet and COCO, rather than the 10$\times$ repetitions done in the original procedure. We may have seen even stronger results with LUAB if annotation byproducts were collected during the original procedure. 
There are also exciting possibilities for exploiting other types of byproducts; one may also estimate image difficulty and annotator biases from the raw annotation byproducts.
Finally, we have restricted our scope to image classifiers. We believe that the LUAB paradigm will benefit other tasks and domains, such as text, audio, video, and tabular data. 

\noindent
\textbf{Take-home messages for dataset building.}
When building a dataset, one should consider logging and releasing the annotation byproducts, along with the main annotations. They may improve models' generalisation and robustness for free.

\noindent
\textbf{Ethical concerns.}
Our data collection for \ourin and \ourcoco has obtained an IRB approval from an author's institute.
We note that there exist potential risks that annotation byproducts may contain annotators' privacy. Data collectors may even attempt to leverage more private information as byproducts. We urge data collectors not to collect or exploit private information from annotators. Whenever appropriate, one must ask for the annotators' consent.

\noindent
\textbf{Acknowledgements.}
We are grateful to NAVER and DGIST for funding the MTurk annotations. We credit Kay Choi for designing the figures. We thank Elif Akata, Elisa Nguyen, and Alexander Rubinstein for reviewing the manuscript. Experiments are based on the NSML \cite{nsml} platform.

{\small
\bibliographystyle{ieee_fullname}
\bibliography{bib}
}
\clearpage
\appendix
\renewcommand\thefigure{\Alph{figure}}    
\setcounter{figure}{0}  
\renewcommand\thetable{\Alph{table}}    
\setcounter{table}{0}  

\noindent\textbf{\Large Appendix}
\vspace{1em}

We include additional information in the Appendix.
In \S\ref{appendix:links}, you can download ImageNet-AB and COCO-AB datasets and find the directories for front-end code for ImageNet and COCO annotation tools.
In \S\ref{appendix:crowdsourcing_details}, we present details for our crowdsourcing-based ImageNet and COCO re-annotations.
In \S\ref{appendix:byproducts_details}, we present extensive lists of byproducts from \ourin and \ourcoco.
In \S\ref{appendix:analysis_annotation_byproducts}, we present further statistics and interesting features of the annotation byproducts in \ourin and \ourcoco.
In \S\ref{appendix:further_experimental_results}, we include additional experimental details and results that supplement the main-paper results.

\section{Links}
\label{appendix:links}

\noindent
Our main repository is at:
\begin{itemize}
    \setlength{\itemsep}{-0em}
    \item \href{https://github.com/naver-ai/NeglectedFreeLunch}{Neglected Free Lunch (GitHub)}
\end{itemize}

\noindent
Download datasets at:
\begin{itemize}
    \setlength{\itemsep}{-0em}
    \item \href{https://huggingface.co/datasets/coallaoh/ImageNet-AB}{ImageNet-AB (HuggingFace)}
    \item \href{https://huggingface.co/datasets/coallaoh/COCO-AB}{COCO-AB (HuggingFace)}
\end{itemize}

\noindent
Please find the codebase for ImageNet and COCO annotation tools in the root directory:
\begin{itemize}
    \setlength{\itemsep}{-0em}
    \item ImageNet: \href{https://github.com/naver-ai/imagenet-annotation-tool}{\small github.com/naver-ai/imagenet-annotation-tool}
    \item COCO: \href{https://github.com/naver-ai/coco-annotation-tool}{\small github.com/naver-ai/coco-annotation-tool}
\end{itemize}
They are replications of respective original annotation tools: \cite{imagenet,imagenet_v2_arxiv} for ImageNet and \cite{coco} for COCO.

\section{Detailed comparison against previous work}
\label{appendix:detailed_related_work}

\definecolor{baselinecolor}{RGB}{216,216,216}
\definecolor{ourscolor}{RGB}{244,190,190}
\definecolor{groupacolor}{RGB}{215,229,255}
\definecolor{groupbcolor}{RGB}{210,255,215}
\setul{-.2em}{0.4em}
\newcommand{\ulcolor}[2][Red]{\setulcolor{#1}\ul{#2}}
\newcommand{\groupa}{\ulcolor[groupacolor]{Group A}\xspace}
\newcommand{\groupb}{\ulcolor[groupbcolor]{Group B}\xspace}

We cluster the related work into two groups in Table~\ref{appendixtab:conceptual-comparison}. \groupa: Solving image classification with additional annotations (\eg semantic segmentation) \cite{RRR2017, simpson2019gradmask, chefer2022optimizing}. \groupb: Solving various vision tasks with point supervision \cite{whats_the_point, ren2020ufo, OpenImagesPoints}.  

\begin{table*}[t]
    \centering
    \tabcolsep=1.2mm
    \small
    \begin{tabular}{crcclcc}
        \toprule
         &  &  &  &  &  \multicolumn{2}{c}{Cost (sec/im)} \\
         & Category & Approach & Target task (evaluation) & Annotation task $\to$ Annotation & ImageNet & COCO \\
        \toprule
         & \ulcolor[baselinecolor]{Baseline} & Classification & Image classification & cls labelling $\to$ cls labels & 1.13 & 36.3 \\
        \midrule
         & \multirow{2}{*}{\vspace{0em}\ulcolor[ourscolor]{Ours}} & Classification  & Image classification & \multirow{2}{*}{\vspace{0em}cls labelling $\to\begin{cases}\text{cls labels}\\[-.5em]\text{AB} \end{cases}$} & \multirow{2}{*}{\vspace{0em}1.13} & \multirow{2}{*}{\vspace{0em}36.3}  \\
         & & (\ours-Ours) & (ImageNet, COCO) & \\
        \midrule
         & \multirow{2}{*}{\vspace{0em}\groupa} & RRR~\cite{RRR2017}, Gradmask~\cite{simpson2019gradmask}, & Image classification & cls labelling $\to$ cls labels & 1.13 & 36.3 \\
         & & RobustViT~\cite{chefer2022optimizing} & (ImageNet evaluation in \cite{chefer2022optimizing}) & segmentation $\to$ object masks & 80\textsuperscript{*} & 280\textsuperscript{*}  \\
        \midrule
         & \multirow{6}{*}{\vspace{0em}\groupb} & \multirow{2}{*}{\vspace{0em} WTP~\cite{whats_the_point}} & Semantic segmentation & cls labelling $\to$ cls labels & NA\textsuperscript{**} & NA\textsuperscript{**} \\
         \vspace{.5em} & & & (Pascal) & point labelling $\to$ points & NA\textsuperscript{**} & NA\textsuperscript{**} \\
         & & \multirow{2}{*}{\vspace{0em}UFO2~\cite{ren2020ufo}} & Object detection & cls labelling $\to$ cls labels & NA\textsuperscript{**} & 80\textsuperscript{***} \\
         \vspace{.5em}& & & (COCO) & point labelling $\to$ points & NA\textsuperscript{**} & 84.9\textsuperscript{\dag} \\
         & & \multirow{2}{*}{\vspace{0em}OpenImagesV7~\cite{OpenImagesPoints}} & Instance segmentation & \multirow{2}{*}{\vspace{0em}Point verification$\to\begin{cases}\text{cls labels}\\[-.5em]\text{points} \end{cases}$} & \multirow{2}{*}{\vspace{0em}0.8\textsuperscript{\dag\dag}} & \multirow{2}{*}{\vspace{0em}2.8\textsuperscript{\dag\dag}} \\
         & & & (OpenImages) \\
         \bottomrule
    \end{tabular}
    \vspace{-0.5em}
    \caption{\small\textbf{Conceptual comparison against previous work.} \footnotesize \textsuperscript{*}It takes 80 sec/polygon \cite{coco}. ImageNet \& COCO have 1 \& 3.5 polygons per image, respectively. \textsuperscript{**}They report results only on Pascal \& COCO, respectively. \textsuperscript{***}Estimate in \cite{ren2020ufo} is only theoretical and it differs from our actual time measurement of 36.3 sec/im. \textsuperscript{\dag}Adopting \cite{ren2020ufo} to the case where 1 point/cls/im is annotated. \textsuperscript{\dag\dag}\cite{OpenImagesPoints} reports 0.8 sec/click for verifying points.}
    \label{appendixtab:conceptual-comparison}
    \vspace{-1em}
\end{table*}

It is possible to make a quantitative comparison against methods in \groupa. They solve the image classification task with extra mask annotation costs\footnote{Mask: 80 \& 280 sec/im, Cls: 1.13 \& 36.3 sec/im for IN \& COCO.} to improve model robustness. Our innovation is that we achieve this effect {without additional supervision costs}. RRR \cite{RRR2017} and Gradmask~\cite{simpson2019gradmask} were only tested on small-scale datasets but are replicated in the RobustViT paper \cite{chefer2022optimizing} for ImageNet evaluation.
We present a quantitative comparison in Table~\ref{appendixtab:quantitative-comparison} with DeiT-B\footnote{ViT-B trained using the DeiT training setup~\cite{deit}.}.

\definecolor{tableplusgreen}{RGB}{21, 152, 56}
\definecolor{tableminusred}{RGB}{252, 54, 65}
\newcommand\tableminus[1]{\textcolor{tableminusred}{#1}}
\newcommand\tableplus[1]{\textcolor{tableplusgreen}{#1}}
\begin{table*}[t]
    \vspace{.75em}
    \small
    \centering
    \tabcolsep=0.2em
    \begin{tabular}{l|c|ccc|ccc|cc|cc|ccc|cc}
        \toprule
        \scriptsize Model & \scriptsize GT Mask & \scriptsize IN-1k↑ & \scriptsize IN-V2↑ & \scriptsize IN-Real↑ & \scriptsize IN-A↑ & \scriptsize IN-C↑ & \scriptsize IN-O↑ & \scriptsize Sketch↑ & \scriptsize IN-R↑ & \scriptsize Cocc↑ & \scriptsize ObjNet↑ & \scriptsize SI-size↑ & \scriptsize SI-loc↑ & \scriptsize SI-rot↑ & \scriptsize BGC-gap↓ & \scriptsize BGC-acc↑\\
        \midrule
        \ulcolor[baselinecolor]{Our DeiT-B} & \ding{55} & 81.6 & 70.3 & 81.1 & 26.1 & 64.1 & 58.0 & 33.0 & 45.7 & 76.0 & 31.7 & 56.6 & 35.1 & 41.3 & 6.4 & 18.1 \\
        \,\,\,\,\ulcolor[ourscolor]{+LUAB (Ours)}  & \ding{55} & \tableplus{+0.9} & \tableplus{+1.6} & \tableplus{+0.7} & \tableplus{+5.0} & \tableplus{+1.9} & \tableplus{+0.5} & \tableplus{+2.5} & \tableplus{+2.7} & \tableplus{+1.5} & \tableplus{+3.3} & \tableplus{+0.5} & \tableplus{+1.7} & \tableplus{+0.3} & \tableplus{-0.8} & \tableplus{+5.8} \\
        \ulcolor[baselinecolor]{ DeiT-B in }\cite{chefer2022optimizing}& \ding{55} & 80.8 & 69.7 & - & 12.9 & - & - & 31.2 & 30.9 & - & 31.4 & 54.6 & 34.5 & 39.3 & - & - \\
        \,\,\,\,\ulcolor[groupacolor]{+Gradmask}~\cite{simpson2019gradmask} & \ding{51} & \tableplus{+0.3} & \tableplus{+0.0} & - & \tableplus{+2.2} & - & - & \tableplus{+0.0} & \tableplus{+0.1} & - & \tableplus{+2.1} & \tableplus{+0.6} & \tableminus{-0.4} & \tableminus{-0.2} & - & - \\
        \,\,\,\,\ulcolor[groupacolor]{+RRR}~\cite{RRR2017} & \ding{51} & \tableplus{+0.2} & \tableplus{+0.2} & - & \tableplus{+1.9} & - & - & \tableminus{-0.3} & \tableplus{+0.2} & - & \tableplus{+2.2} & \tableplus{+0.7} & \tableminus{-0.1} & \tableplus{+1.1} & - & - \\
        \,\,\,\,\ulcolor[groupacolor]{+RobustViT}~\cite{chefer2022optimizing} & \ding{51} & \tableminus{-0.3} & \tableminus{-0.6} & - & \tableplus{+4.3} & - & - & \tableminus{-0.3} & \tableplus{+1.5} & - & \tableplus{+4.5} & \tableplus{+3.4} & \tableplus{+2.1} & \tableplus{+3.6} & - & - \\
        \bottomrule
    \end{tabular}
    \vspace{-0.5em}
    \caption{\small\textbf{Quantitative comparison against prior work.} We compare ours with the prior arts, including Gradmask~\cite{simpson2019gradmask}, RRR~\cite{RRR2017}, and RobustViT~\cite{chefer2022optimizing} using DeiT-B on ImageNet-1k and variant robustness benchmarks.}
    \label{appendixtab:quantitative-comparison}
    \vspace{-0.5em}
\end{table*}

Our LUAB framework improves the performance on {all} ImageNet benchmarks, whereas \groupa methods show mixed results. Importantly, unlike \groupa methods, {our improvements do not assume the availability of GT masks}. Moreover, LUAB is applicable to general model types, while RobustViT is limited to ViT variants. 

Evaluation of \groupb methods is not compatible, as their target task is not image classification. We report their annotation costs for point supervision in the table. Our contribution to \groupb community is the finding that \textit{weak point supervision may be obtained without additional cost from the class labelling procedure}. OpenImagesV7 \cite{OpenImagesPoints} introduces an efficient labelling scheme, but it relies on a pre-trained segmentation model (IRN \cite{IRN}) to propose points; it is not directly comparable in our setting.

\section{Annotation and crowdsourcing details}
\label{appendix:crowdsourcing_details}

\subsection{ImageNet}
\label{appendix:crowdsourcing_details_imagenet}

We provide further details on the crowdsourced ImageNet annotation. We hired Amazon Mechanical Turk (MTurk) workers from the US region, as the task is described in English. The minimal human intelligence task (HIT) approval rate for the task qualification was set at 90\% to ensure a minimal quality for the task. 

Each HIT contains 10 pages of the annotation task, each with 48 candidate images. Upon completion, the annotators are paid 1.5 USD per HIT. It is difficult to convert this amount to an exact hourly wage due to the high variance and noise in the measured time to complete each HIT. A rough conversion is possible through the median HIT, which took 9.0 minutes to complete. This yields an hourly wage of 10.0 USD, well above the US federal minimum hourly wage of 7.25 USD \cite{minimum_wage}.

When the submitted work shows clear signs of gross negligence and irresponsibility, we reject the HIT. Specifically, we reject a HIT if:
\begin{itemize}
    \setlength{\itemsep}{-0em}
    \item the recall rate, defined as the proportion of selected images $I_c^\text{select}$ among the original ImageNet subset $I_c^\text{in}$, is lower than 0.333; or
    \item the total number of selections $I_c^\text{select}$ among 480 candidates is lower than 30 (there are $480\times 0.75=360$ samples from ImageNet $I_c^\text{in}$ on average); or
    \item the annotator has not completed at least 9 out of the 10 pages of tasks; or
    \item the annotation is not found in our database AND the secret hash code for confirming their completion is incorrect.
\end{itemize}
Among 14,681 HITs completed, 1,145 (7.8\%) have been rejected. Collectively, we have paid $20,304\text{ USD}=13,536\text{ approved HITs}\times 1.5\text{ USD }/\text{ HIT}$ to the MTurk annotators. An additional 20\% fee is paid to Amazon ($4,060.8$ USD). The entire procedure took place between 18 December 2021 and 31 December 2021.

\paragraph{Annotation interface.}
We have tried nudging the annotators to click more frequently on the foreground objects by changing the cursor shape to a red circle and instructing them to ``click on the object of interest'' while selecting the images. According to our pilot study, this increases the chance of annotators clicking on the object of interest from 70.7\% to 91.7\% (p-value $<$0.0005), while not increasing the annotation time meaningfully: 2.02 to 2.09 minutes per page (p-value 0.456).

\subsection{COCO}
\label{appendix:crowdsourcing_details_coco}

For COCO, we follow the ImageNet annotation setup in \S\ref{appendix:crowdsourcing_details_imagenet} for the worker region and worker qualification.

Each annotation page contains a single image to be annotated. We collate 20 pages into a single human intelligence task (HIT). That results in $82,783\text{ images}\times \frac{1\text{ HIT}}{20\text{ images}}=4,140\text{ HITs}$. The compensation for each HIT is 2.0 USD. The median HIT has been completed in 12.1 minutes. This leads to the hourly wage of 9.92 USD, which is above the US Federal minimum wage of 7.25 USD \cite{minimum_wage}.

We reject HITs based on the following criteria
\begin{itemize}
    \setlength{\itemsep}{-0em}
    \item the recall rate, defined as the proportion of retrieved classes among the existing classes, is lower on average than 0.333; or
    \item the accuracy of icon location, defined as the ratio of icons placed on the ground-truth class segmentation mask, is lower than 0.75; or
    \item the annotator has not completed at least 16 out of the 20 pages of tasks; or
    \item the annotation is not found in our database AND the secret hash code for confirming their completion is incorrect.
\end{itemize}

By continuously re-posting rejected HITs, we have acquired the necessary annotation and byproducts on 4140 HITs. Along the way, we have rejected 365 HITs, giving us a rejection rate 8.8\%. Collectively, we have paid $8,280\text{ USD}=4,140\text{ approved HITs}\times 2\text{ USD }/\text{ HIT}$ to 662 MTurk annotators. An additional 20\% fee is paid to Amazon (1656 USD). The annotation took place between 9 January 2022 and 12 January 2022.

\section{Byproducts details}
\label{appendix:byproducts_details}

\subsection{\ourin}
\label{appendix:byproducts_imagenet}

\begin{figure}
    \centering
    \includegraphics[width=\linewidth]{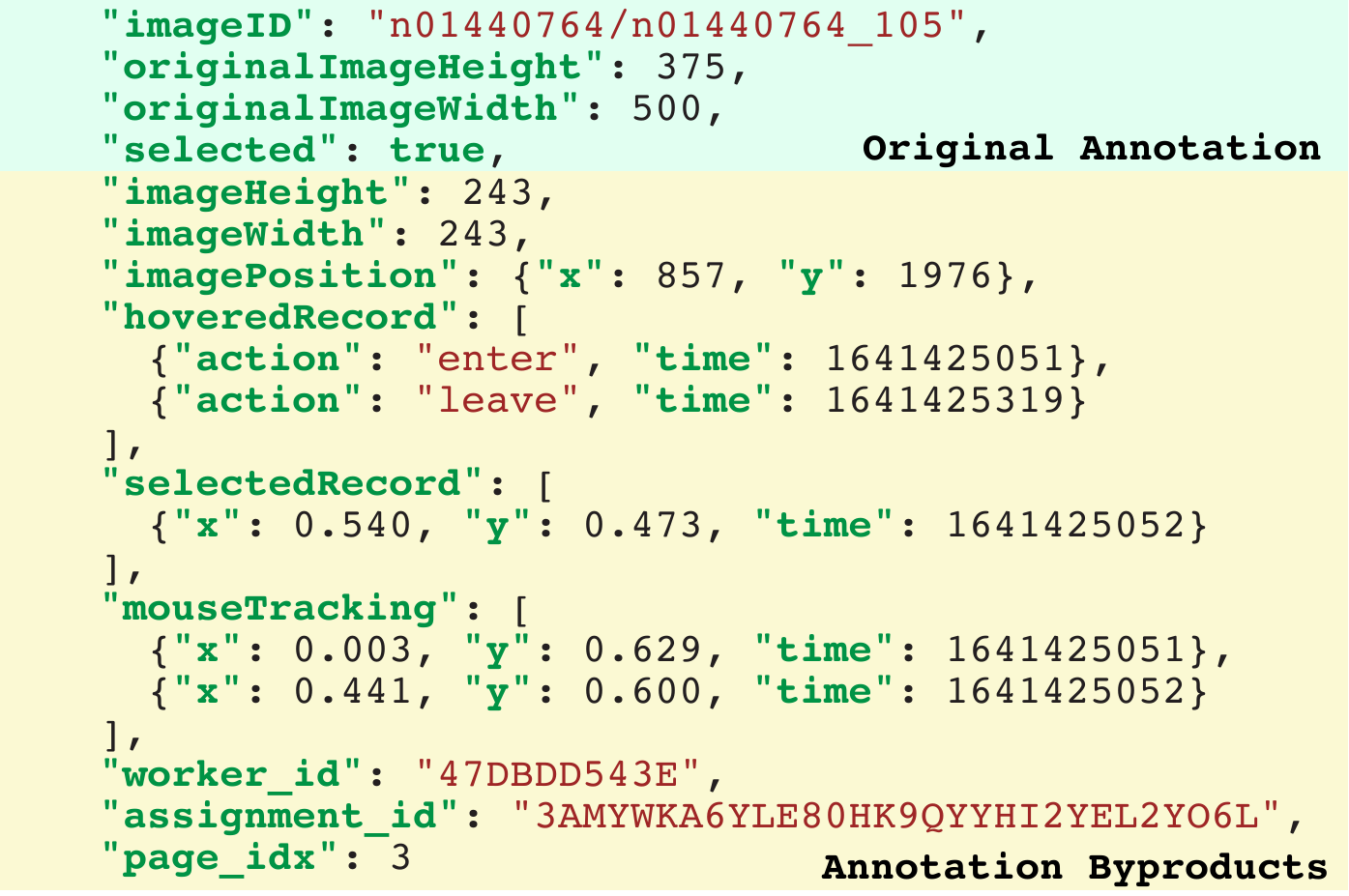}
    \vspace{-1.5em}
    \caption{\small\textbf{Annotation byproducts from ImageNet.} Worker ID has been anonymised via non-reversible hashing. Extended version of Figure
    \ref{fig:annotation_byproduct_imagenet_sample}. 
    }
    \label{appendixfig:annotation_byproduct_imagenet_sample}
    \vspace{-.5em}
\end{figure}

We explain the details of \ourin, the ImageNet1k training set enriched with annotation byproducts.
Annotators use input devices to interact with different components in the annotation interface. This results in a history of interactions per input signal per front-end component. On ImageNet, annotators interact with each image (component) on each page with two types of input signals: mouse movements and mouse clicks 
(Figure~\ref{fig:imagenet_annotation_interface}).
We show the full list of annotation byproducts in Figure \ref{appendixfig:annotation_byproduct_imagenet_sample}. This results in the time series of mouse movements (\texttt{mouseTracking}) and mouse clicks (\texttt{selectedRecord}) for every image. We separately record whether the image is finally selected by the annotator in the \texttt{selected} field. It is \texttt{true} when the length of \texttt{selectedRecord} is an odd number.

In our work, we only demonstrate the usage of additional \texttt{selectedRecord} as a proxy to the object localisation information and show that this alone greatly enhances the models' robustness. However, there exist other byproducts that may further improve the trained models. We introduce them below and hope that future researches find ways to maximally exploit those additional signals.

We record sufficient yet compact information to reproduce the annotation page: x-y coordinates (\texttt{imagePosition}) and the width and height (\texttt{imageWidth} and \texttt{imageHeight}) of each image in the annotation interface. This information can be useful because the mouse movement pattern is highly entangled with the page layout. For example, annotators are likely to minimise mouse movement by following a serpentine sequence. 

We record other annotation metadata for each image, such as the worker identifier (\texttt{worker\_id}), the identifier for the human intelligence task (HIT) that contains this image (\texttt{assignment\_id}), and the page number within the HIT (\texttt{page\_idx}). We have anonymised the worker identifier with a non-reversible hashing function. Those metadata provide information for grouping the annotation instances with increasing specificity: $\{$annotations on the same page$\}\subset\{$annotations from the same HIT$\}\subset\{$annotations by the same worker$\}$. Such information may be helpful for identifying and factoring out group-specific idiosyncrasies. For example, worker \texttt{ABC} may always click near the centre of an image; we may then decide not to use her clicks as a reliable estimate of object locations. Or we may find that the HIT \texttt{DEF} was done in such a rush; we would then reduce the weight for the set of annotations belonging to \texttt{DEF}.

\paragraph{Statistics.}
There are 1,281,167 ImageNet1K training images $I^\text{imagenet}$. There were two annotation rounds. In the first round, human intelligence tasks (HITs) containing all 1,281,167 original images are shown to the annotators. They have re-selected 71.8\% of them. This confirms the observation of \cite{imagenet_v2} that 71\% of the validation set samples were re-selected in their setting. The remaining 28.2\% of $I^\text{imagenet}$ are re-packaged into a second batch of HITs and presented to the annotators. They have additionally re-selected 14.9\% of $I^\text{imagenet}$, resulting in the final 1,110,786 (86.7\%) ImageNet1K training images that are re-selected. Those selected images now come with rich annotation byproducts, such as the time-series of mouse traces and clicks. However, annotation byproducts are available even for images that are not finally selected; they are recorded even for images that annotators cancel the selection or simply hover the cursor over. As a result, 1,272,225 (99.3\%) of the ImageNet1K training set have any form of annotation byproduct available.

\subsection{COCO-AB}
\label{appendix:byproducts_coco}

\begin{figure}[t]
    \centering
    \includegraphics[width=\linewidth]{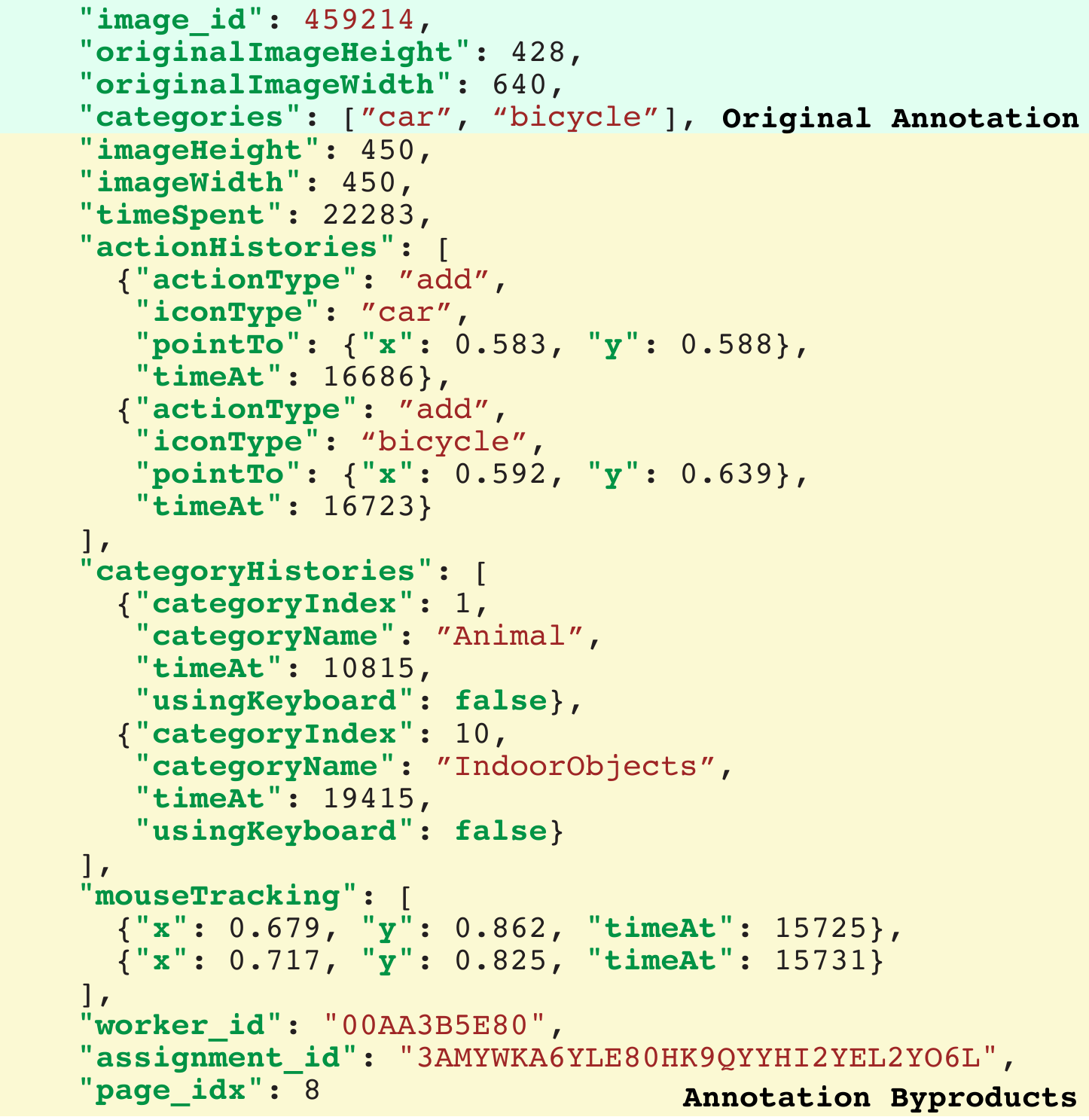}
    \vspace{-1.5em}
    \caption{\small\textbf{Annotation byproducts from COCO.} Worker ID has been anonymised via non-reversible hashing. Extended version of Figure 
    \ref{fig:annotation_byproduct_coco_sample}.
    }
    \label{appendixfig:annotation_byproduct_coco_sample}
    \vspace{-1em}
\end{figure}

We explain the details of \ourcoco, the COCO 2014 training set enriched with annotation byproducts.
COCO interface 
(Figure \ref{fig:coco_annotation_interface}) 
has two main components: (1) the image on which the class icons are placed and (2) the class browsing tool showing the class icons. The annotation byproducts come from those two sources. See Figure \ref{appendixfig:annotation_byproduct_coco_sample} for the full list of annotation byproducts. 

The \texttt{actionHistories} field describes the actions performed with the mouse cursor on the image. \texttt{actionHistories} list the sequence of actions with possible types \texttt{add}, \texttt{move}, and \texttt{remove} and the corresponding location and time. We also record the object class of the icon. The \texttt{mouseTracking} field records the movement of the mouse cursor over the image.

Interactions with the class browsing tool leave a time series of superclasses that the annotator refers to. They are stored in the field \texttt{categoryHistories}. We also allow interactions based on keyboard (left and right arrows); the use of keyboard is indicated in \texttt{usingKeyboard}. 

We record the total time spent for the annotation (\texttt{timeSpent}). To provide the context of the annotation work, we have stored the page number (\texttt{page\_idx}), the identifier for the HIT package (\texttt{assignment\_id}), and the anonymised identifier for the annotator (\texttt{worker\_id}). 

In this work, we only use the last \texttt{add} action in the \texttt{actionHistories} field for each object class to additionally supervise the model to be aware of the actual location of the object in the image. However, the recordings of other interaction histories may be used in future work as additional sources that further improve the trained models.

\paragraph{Statistics.}
Annotators have reannotated 82,765 (99.98\%) of 82,783 training images from the COCO 2014 training set. For those images, we have recorded the annotation byproducts. We found that each HIT recalls 61.9\% of the list of classes per image, with the standard deviation $\pm$0.118\%p. The average localisation accuracy for icon placement is 92.3\% where the standard deviation is $\pm$0.057\%p.

\section{Analysis of annotation byproducts}
\label{appendix:analysis_annotation_byproducts}

\subsection{ImageNet}

We analyse the annotation byproducts in more detail. In particular, we measure the informativeness of mouse clicks and traces for the location of objects in an image. All analyses involving the ``ground-truth (GT) bounding boxes'' is performed on the 42\% of the ImageNet1K training set annotated with instance-wise bounding boxes.

\begin{figure}[t]
    \centering
    \small
    \setlength{\tabcolsep}{.1em}
    \begin{tabular}{cc}
        \includegraphics[width=.49\linewidth]{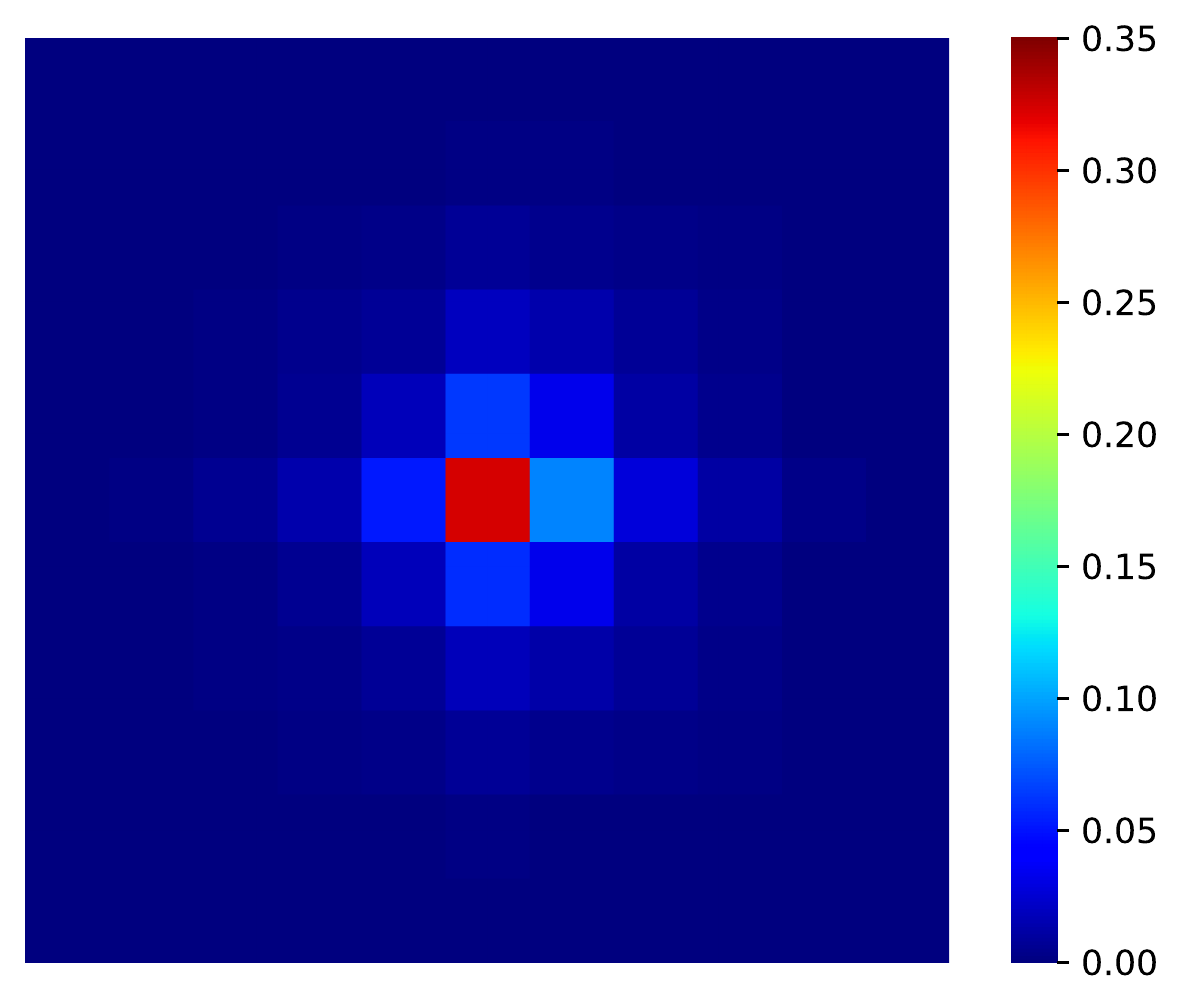} &
        \includegraphics[width=.49\linewidth]{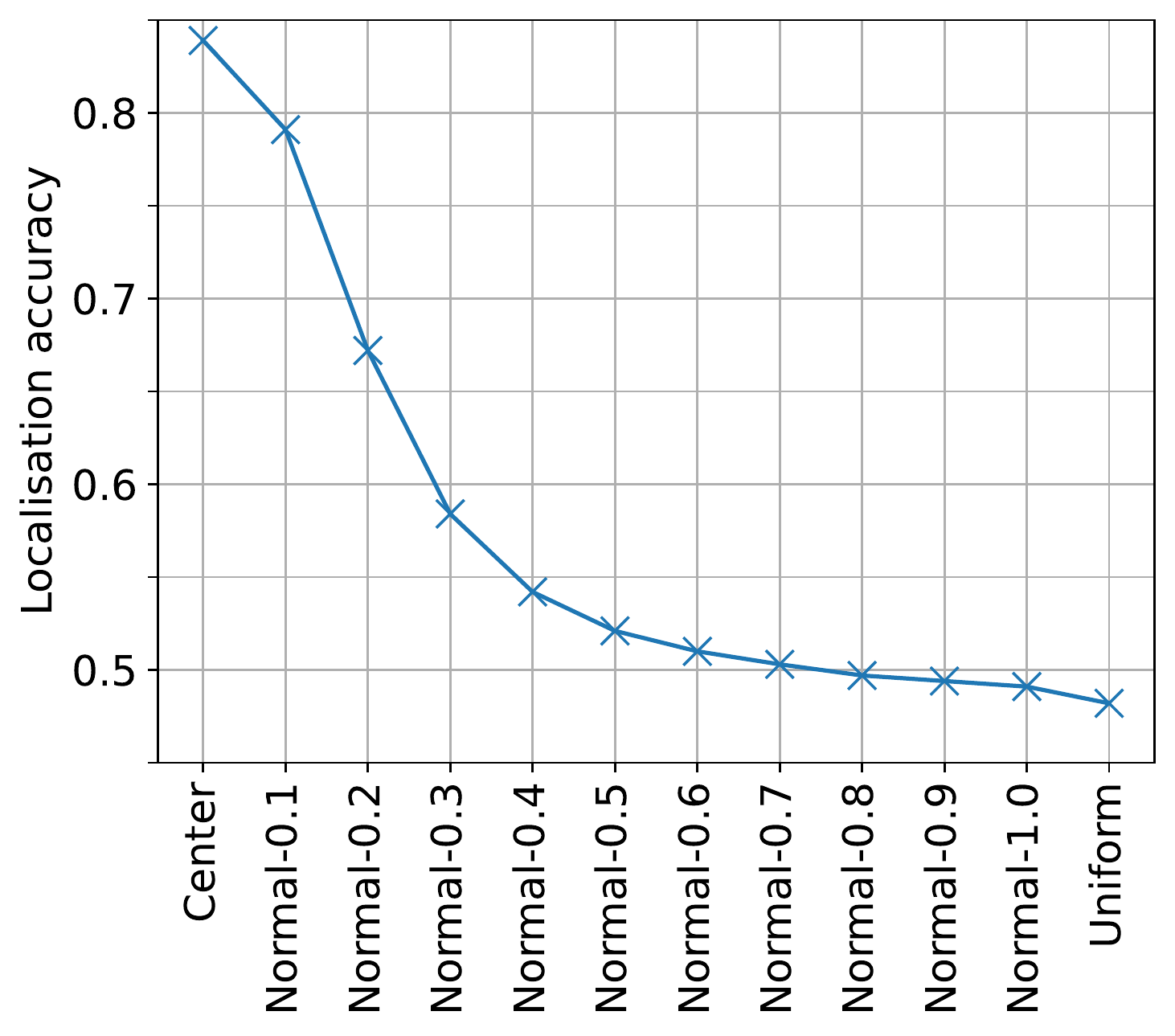}
    \end{tabular}
    \vspace{-1em}
    \caption{\small\textbf{ImageNet GT-box statistics.} \textbf{Left}: distribution of GT box centres on ImageNet1K training set images. \textbf{Right}: localisation accuracy of random clicks $N((\frac{H}{2},\frac{W}{2}),
    \sigma^2)$. We interpolate between centre-always click ($\sigma=0$) and uniform random click ($\sigma=\infty$).}
    \label{appendixfig:imagenet-gt-box}
    \vspace{-1em}
\end{figure}

\paragraph{GT bounding boxes on ImageNet.}
ImageNet is a highly object-centric dataset. This is reconfirmed by the distribution of the centre of the GT boxes in Figure \ref{appendixfig:imagenet-gt-box} (left). More than 30\% of the box centres are located in the 0.82\% area at the centre of the images. 

We measure the localisation accuracy of random image-agnostic clicks in Figure \ref{appendixfig:imagenet-gt-box} (right). We experimented with the random click distribution $N((\frac{H}{2},\frac{W}{2}), \sigma^2)$ where $\sigma\in[0,\infty]$ interpolates between the click-always-at-the-centre strategy ($\sigma=0$) and the uniform random click ($\sigma=\infty$). We observe that clicking at the image centre yields 83.9\% localisation accuracy, actually greater than the localisation accuracy of clicks 82.9\%. Despite a lower overall accuracy, we will see later in the current section that the annotators' clicks contain much richer information about the variation of object locations than simple centre clicks. 

As $\sigma$ increases, the localisation accuracy drops and reaches 48.2\% when clicks are uniformly random $\sigma=\infty$. The 48.2\% value can be interpreted as the average bounding box area in each image. The relatively high average area of the objects again signifies the object-centric nature of the ImageNet dataset.

\begin{figure}[h]
    \centering
    \small
    \setlength{\tabcolsep}{.1em}
    \begin{tabular}{cc}
        \includegraphics[width=.49\linewidth]{figures/click_stats/box_center_distribution.pdf} &
        \includegraphics[width=.49\linewidth]{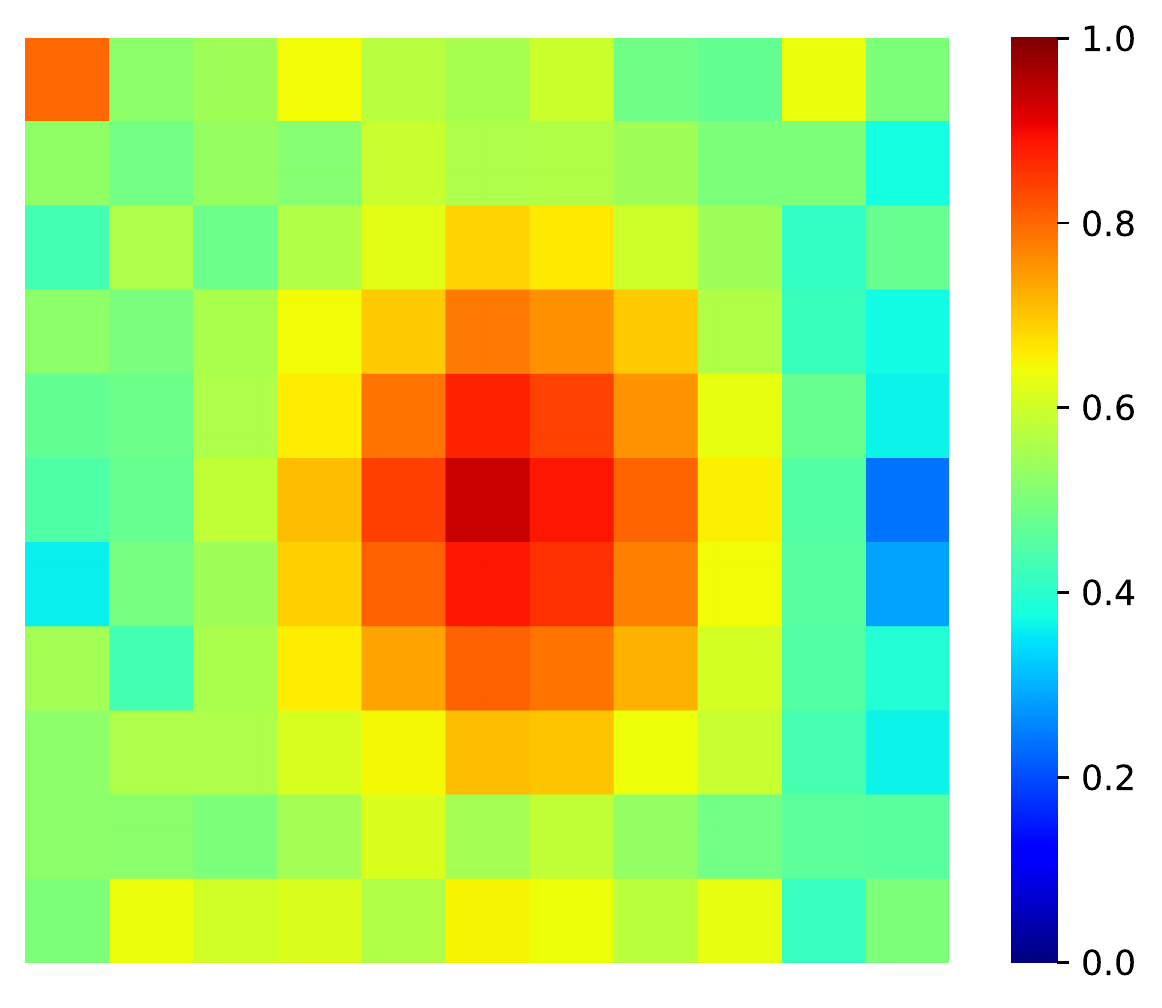} \\
        {\textbf{GT box-centre distribution}} &
        {\textbf{Loc acc per GT location}} \\
        \includegraphics[width=.49\linewidth]{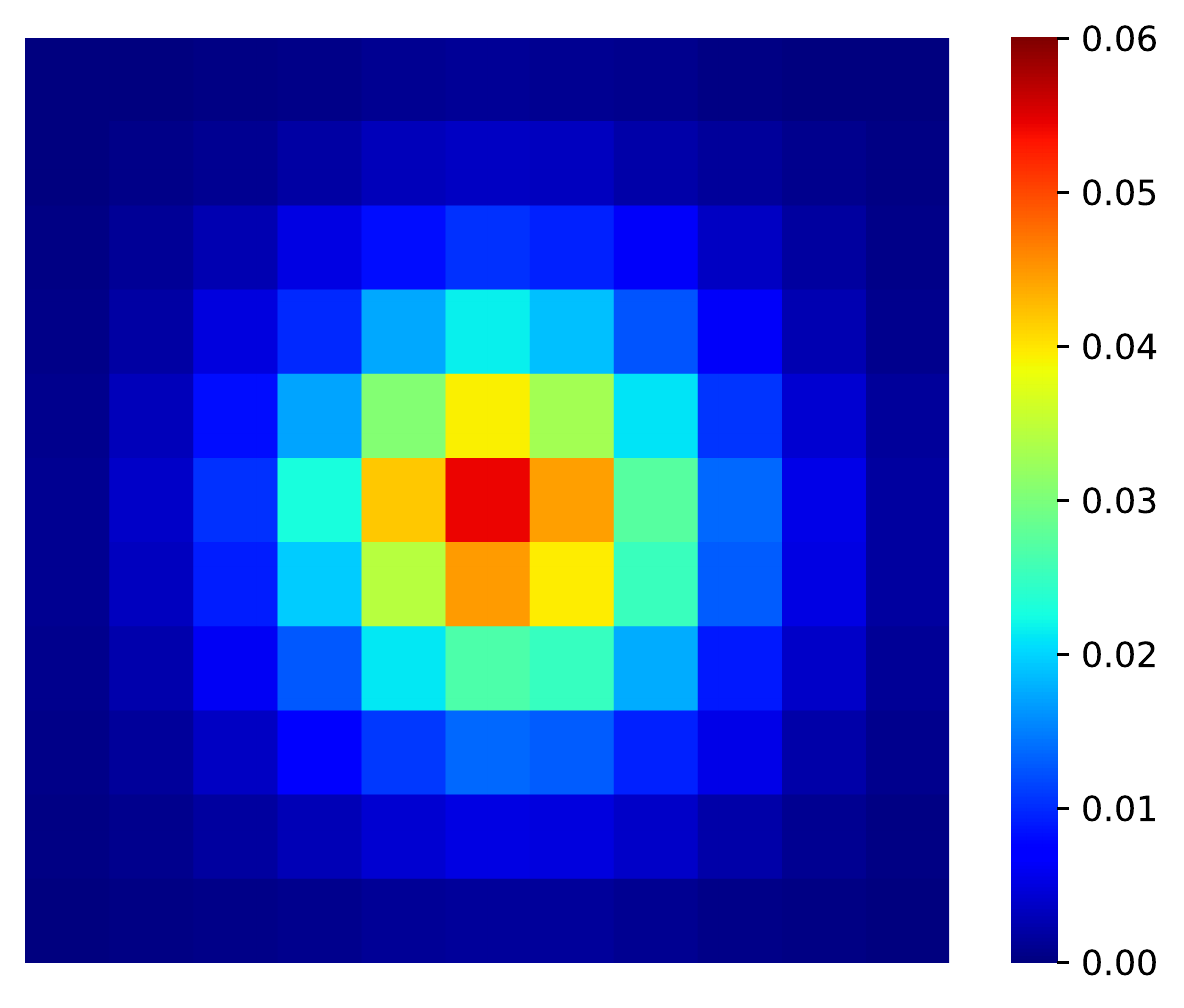} &
        \includegraphics[width=.49\linewidth]{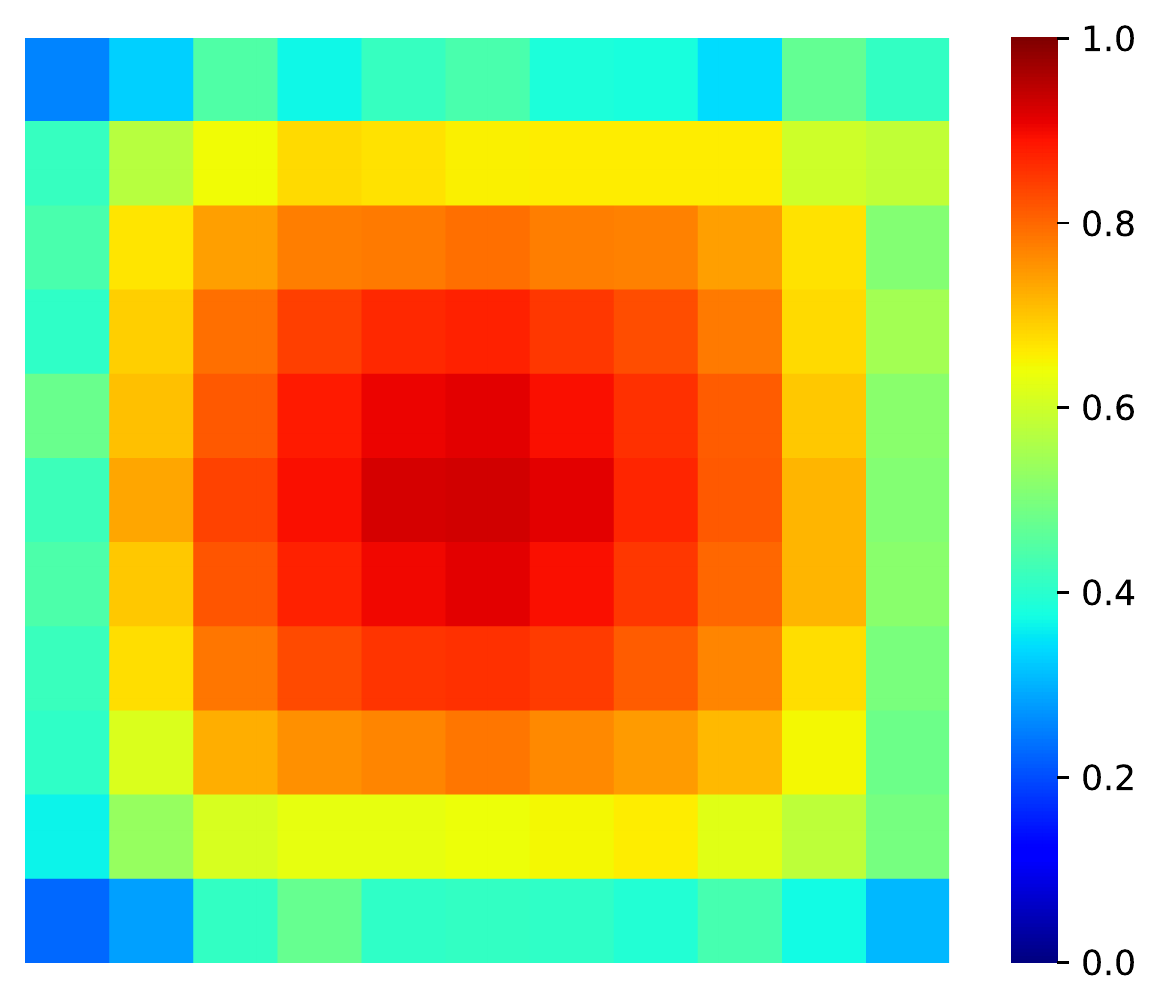} \\
        {\textbf{Click distribution}} &
        {\textbf{Loc acc per click location}}
    \end{tabular}
    \vspace{-1em}
    \caption{\small\textbf{Statistics of clicks.} \textbf{Left column}: distribution of GT box centres and clicks in ImageNet1K images. \textbf{Right column}: localisation accuracy of clicks at each GT box centre location and click location.}
    \label{appendixfig:click_statistics}
    \vspace{-1em}
\end{figure}

\paragraph{Informativeness of clicks.}
We examine whether the clicks contain information about the variation of object locations. The analysis is not as simple as measuring the overall localisation accuracy, since the dataset is highly object-centric: we have seen above that centre clicks already give 83.9\% localisation accuracy, greater than the localisation accuracy of clicks 82.9\%. The majority of information about the object location is contained in 16.1\% of the samples where a simple centre-click strategy cannot guarantee a correct localisation. In this subset of images where objects are not at the centre, the localisation accuracy of clicks is 56.5\%. This implies great information content, as simple centre clicks will give 0\% accuracy on this subset.

To further break down the localisation accuracy based on the location of objects and click locations, we plot the location-wise click accuracy in Figure \ref{appendixfig:click_statistics} (right column). For reference, we also plot the distribution of GT box centres and clicks in the left column. We observe that the localisation accuracy at each GT box location and the click location remain $>40\%$, except at the outermost image borders. This confirms the overall informativeness of clicks for the object locations, despite the severe bias towards the image centre in the dataset.

\begin{figure}[h]
    \vspace{1mm}
    \centering
    \small
    \setlength{\tabcolsep}{.1em}
    \begin{tabular}{c}
        \includegraphics[width=.49\linewidth]{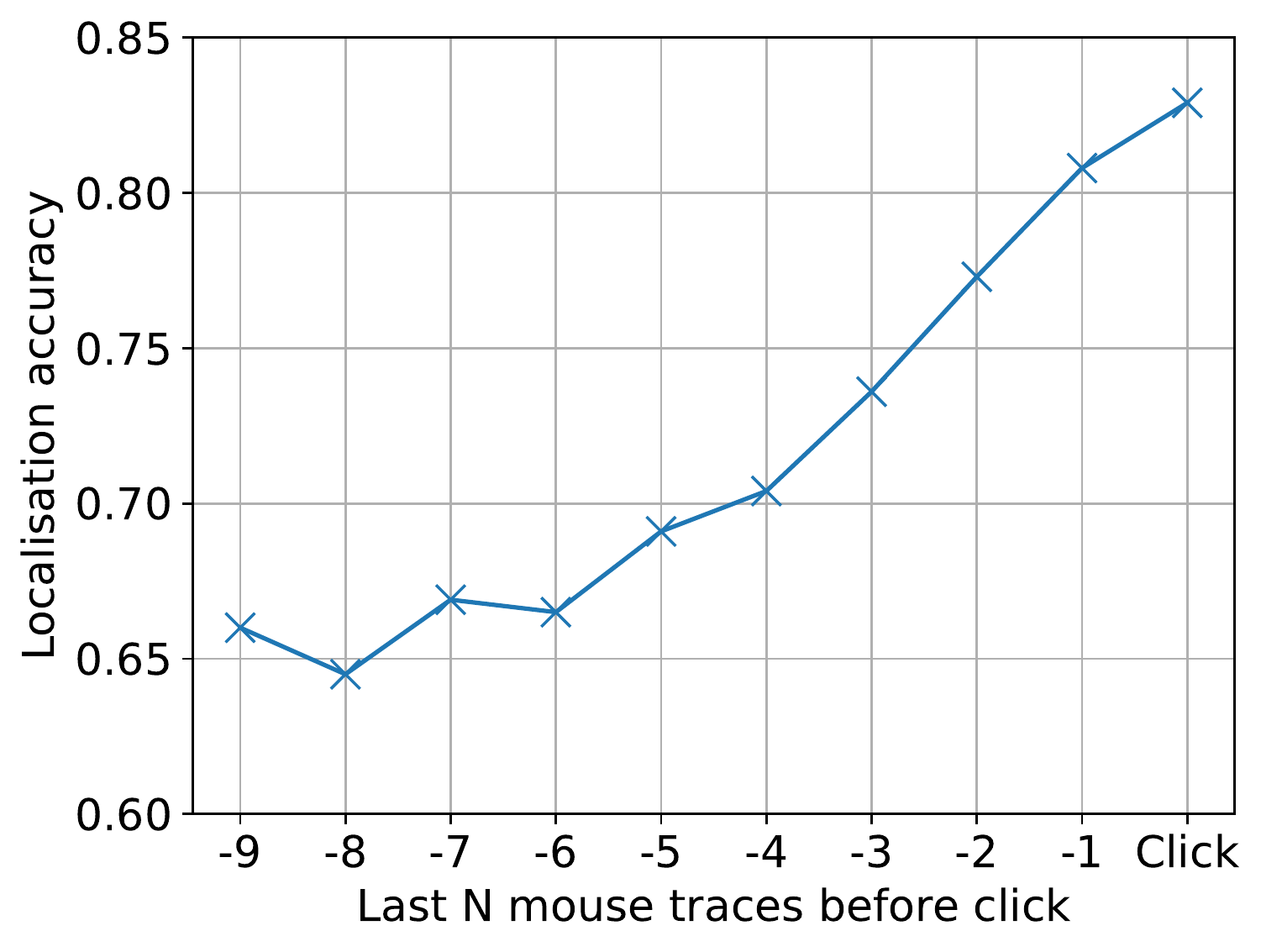} \\
        {\textbf{Last N}}
    \end{tabular}
    \vspace{1em}
    \begin{tabular}{cc}
        \includegraphics[width=.49\linewidth]{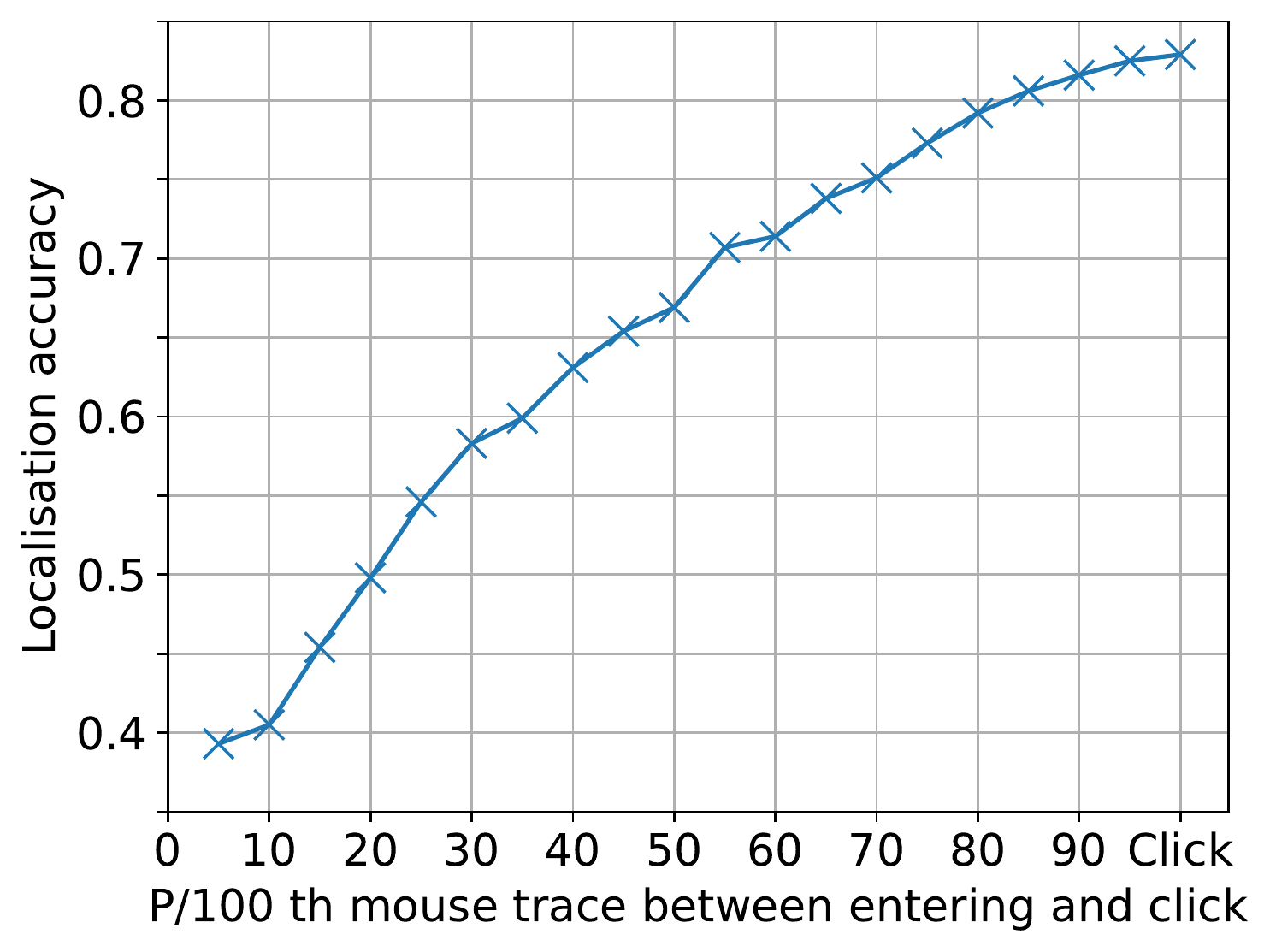} &
        \includegraphics[width=.49\linewidth]{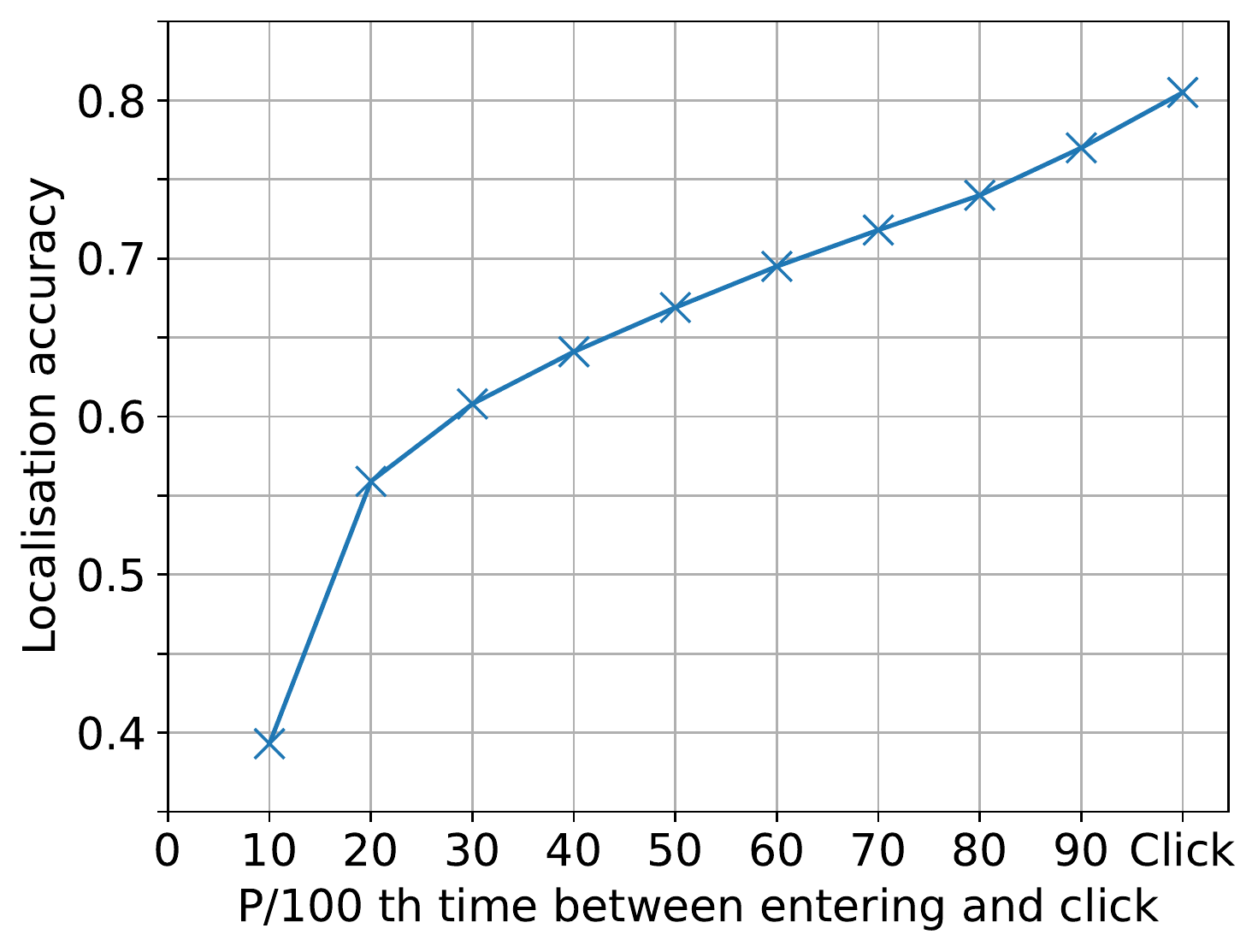}\\
        {\textbf{Trace quantile}} &
        {\textbf{Time quantile}}
    \end{tabular}
    \vspace{-1.5em}    
    \caption{\small\textbf{Statistics for mouse traces before click.} \textbf{Last N}: last N mouse traces before click. \textbf{Trace quantile}: division of each mouse trace from the ``entering image'' event to the ``click'' event in the equal number of mouse track records. \textbf{Time quantile}: same as trace quantile, except that bins are groups by the time. }
    \label{appendixfig:mouse_traces_statistics}
    \vspace{-1em}
\end{figure}

\paragraph{Informativeness of mouse traces.}
Annotation byproducts include not only clicks but the full history of mouse traces over each image. We measure the localisation accuracy of the mouse traces between entering the image and click. The results are reported in Figure~\ref{appendixfig:mouse_traces_statistics}. 
Last few mouse trace records before click (Last N) show a mild drop in accuracy (from 82.9\% to $\sim 65\%$ at 8 traces before click); therefore, the last few points before click may give useful localisation information.
The trace and time quantile results show that the localisation accuracy is very low when the mouse enters an image (39.3\%). The accuracy increases up to the point when the user clicks (82.9\%). We observe that the last 10\% of the mouse traces (both for trace and time quantile) are still fairly precise with accuracy $>80\%$. The above observations imply the possibility that one may also utilise a few mouse trace records before the click event to obtain a weak localisation supervision based on scribbles \cite{whats_the_point}.

\begin{figure}[h]
    \centering
    \includegraphics[width=.9\linewidth]{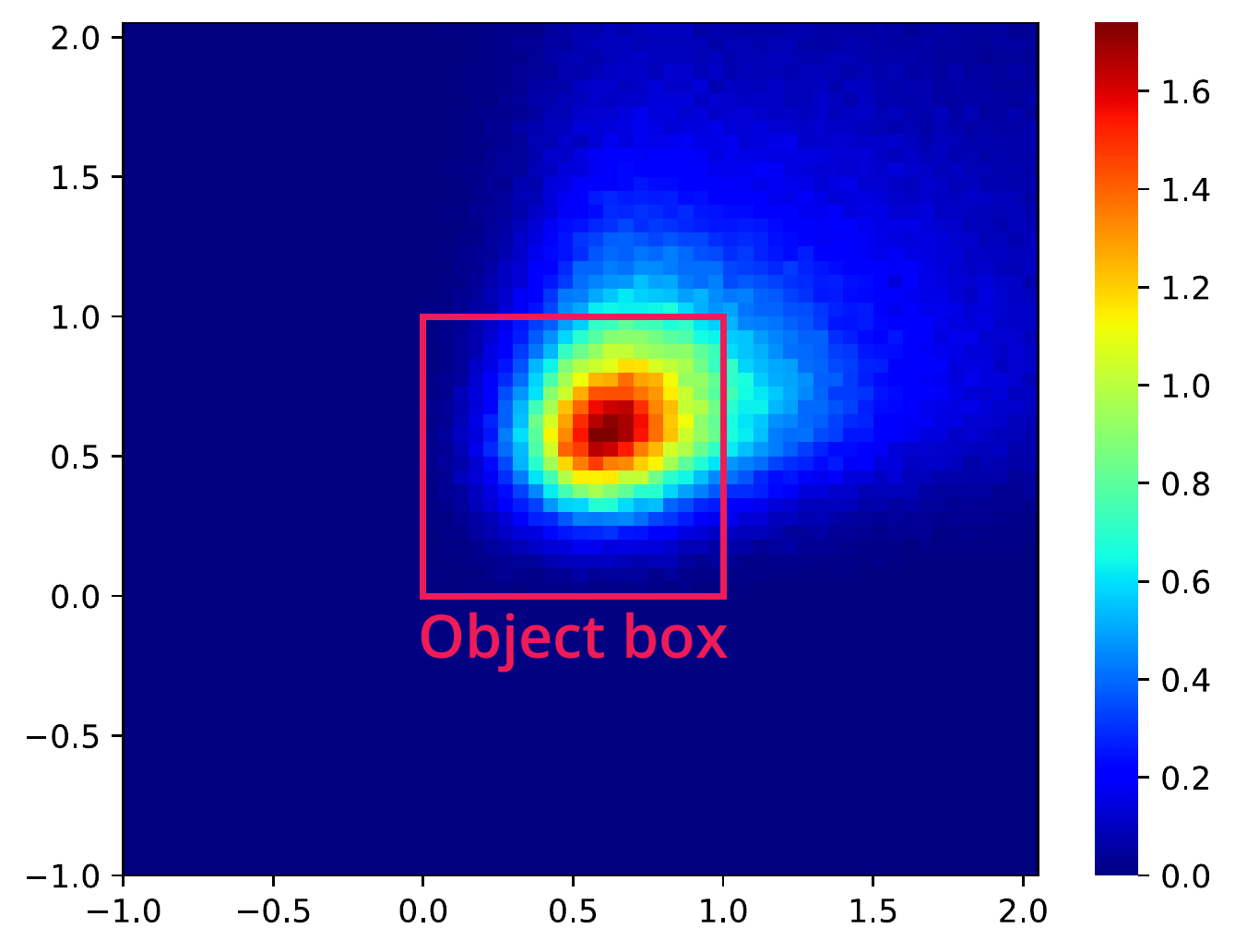}
    \vspace{-1em}
    \caption{\small\textbf{Click histogram relative to GT box on ImageNet.} Distribution of click positions normalised against the GT object box frame at $[0,1]\times [0,1]$.}
    \label{appendixfig:gt_relative_clicks}
    \vspace{-1em}
\end{figure}

\paragraph{Click are systematically biased to the top-right corner.}
Figure \ref{appendixfig:gt_relative_clicks} shows the distribution of clicks relative to the GT object boxes. We observe that the mode of the distribution is close to the centre, but slightly biased to the upper-right corner. The tail of the distribution is more drastically biased towards the top-right corner, almost forming a comet-like shape.
We conjecture that browsing through rows of images makes annotators enter an image through the top side and leave it through the right side. And this leaves such a systematic error around the actual location of the objects.
Given the systematic bias, it would be an interesting future research direction to either post-hoc calibrate click locations or nudge annotators to reduce the top-right-corner bias for better object localisation.

\subsection{COCO}

\begin{figure}[t]
    \centering
    \small
    \setlength{\tabcolsep}{.1em}
    \begin{tabular}{cc}
        \includegraphics[width=.49\linewidth]{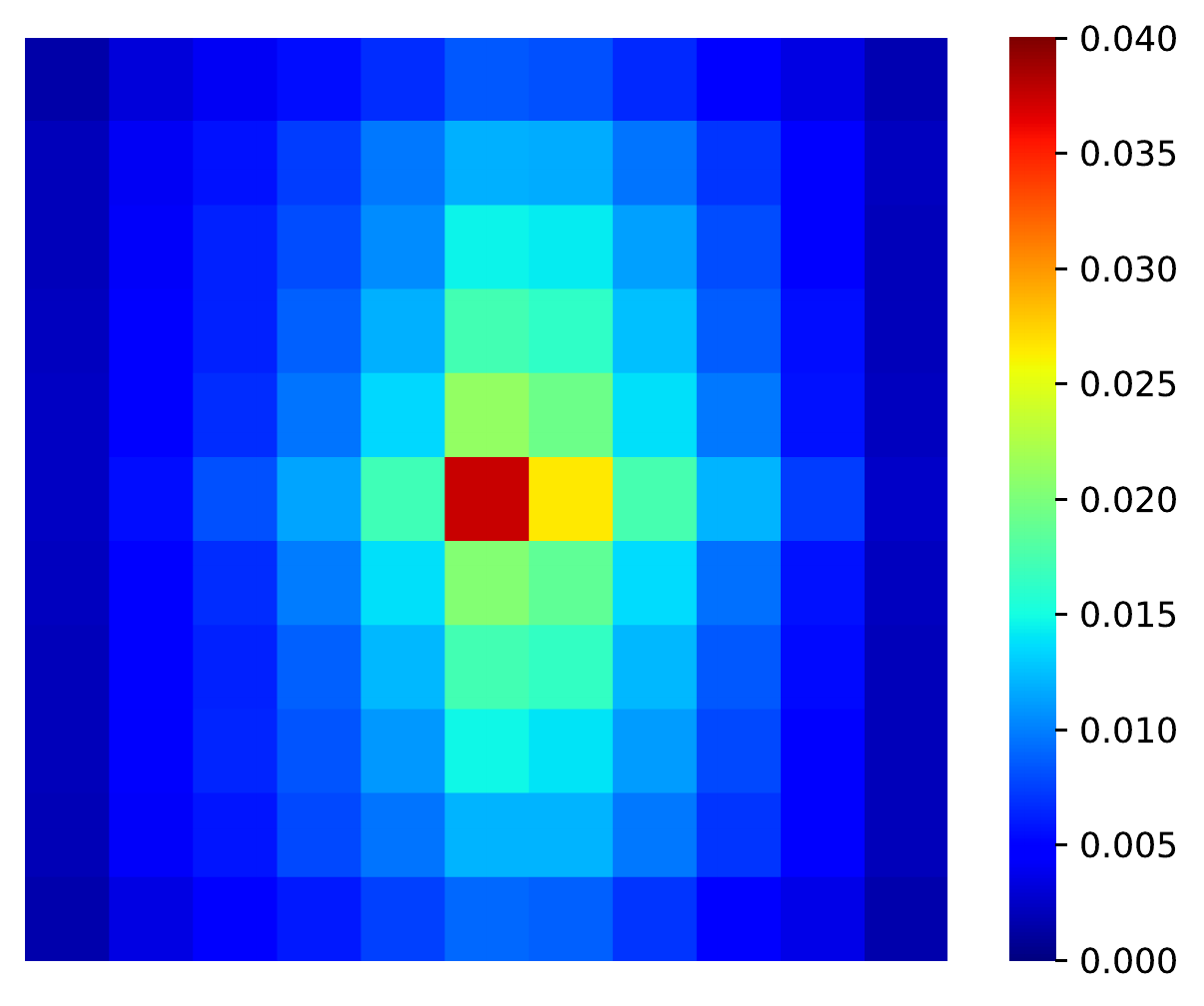} &
        \includegraphics[width=.49\linewidth]{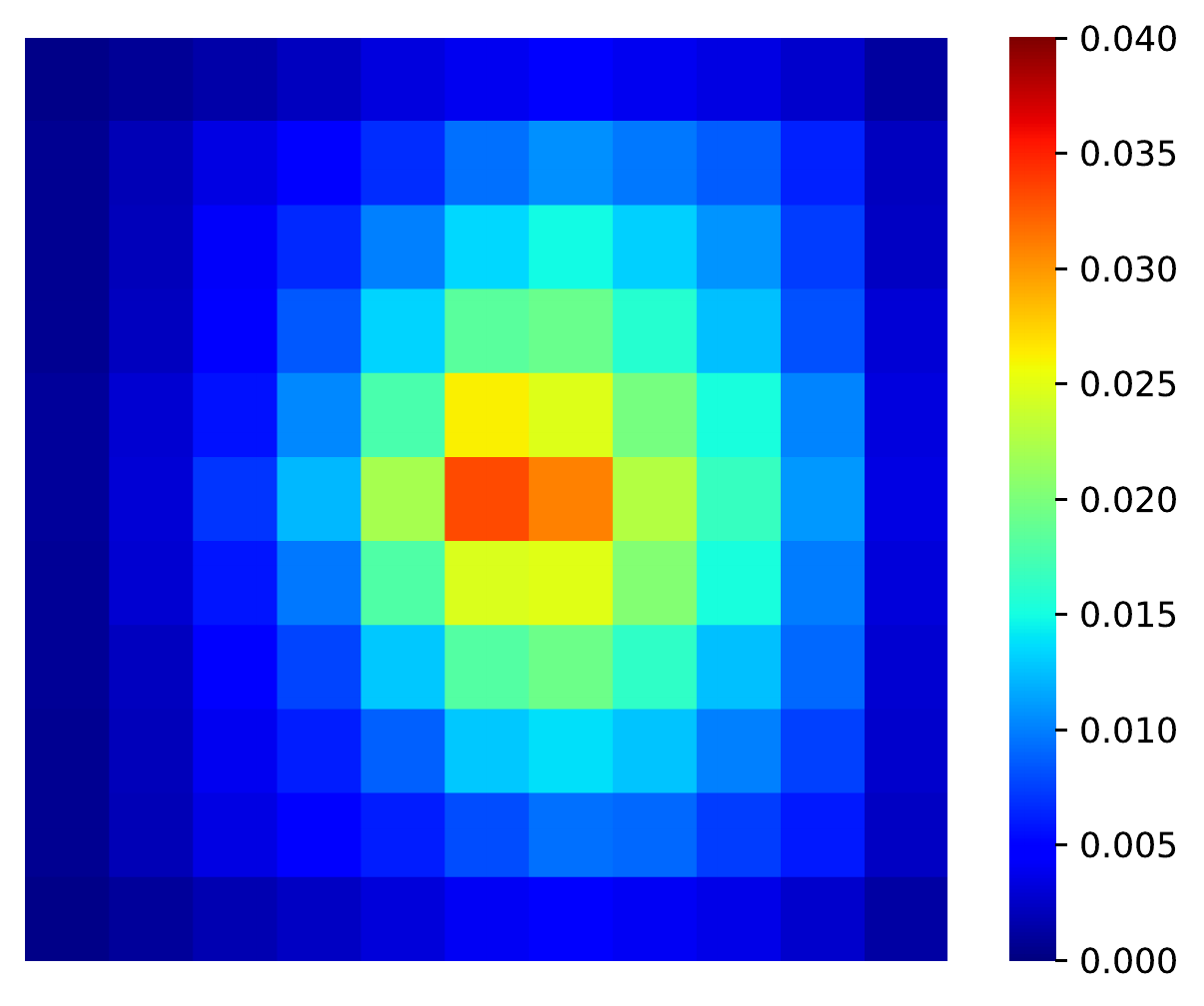} \\
        {\textbf{GT box-centre}} &
        {\textbf{Final ``add'' location}}
    \end{tabular}
    \vspace{-1em}    
    \caption{\small\textbf{Statistics of icon placement.} Statistics for the location of objects and the final icon placements.}
    \label{appendixfig:icon_statistics}
    \vspace{1em}
    \includegraphics[width=\linewidth]{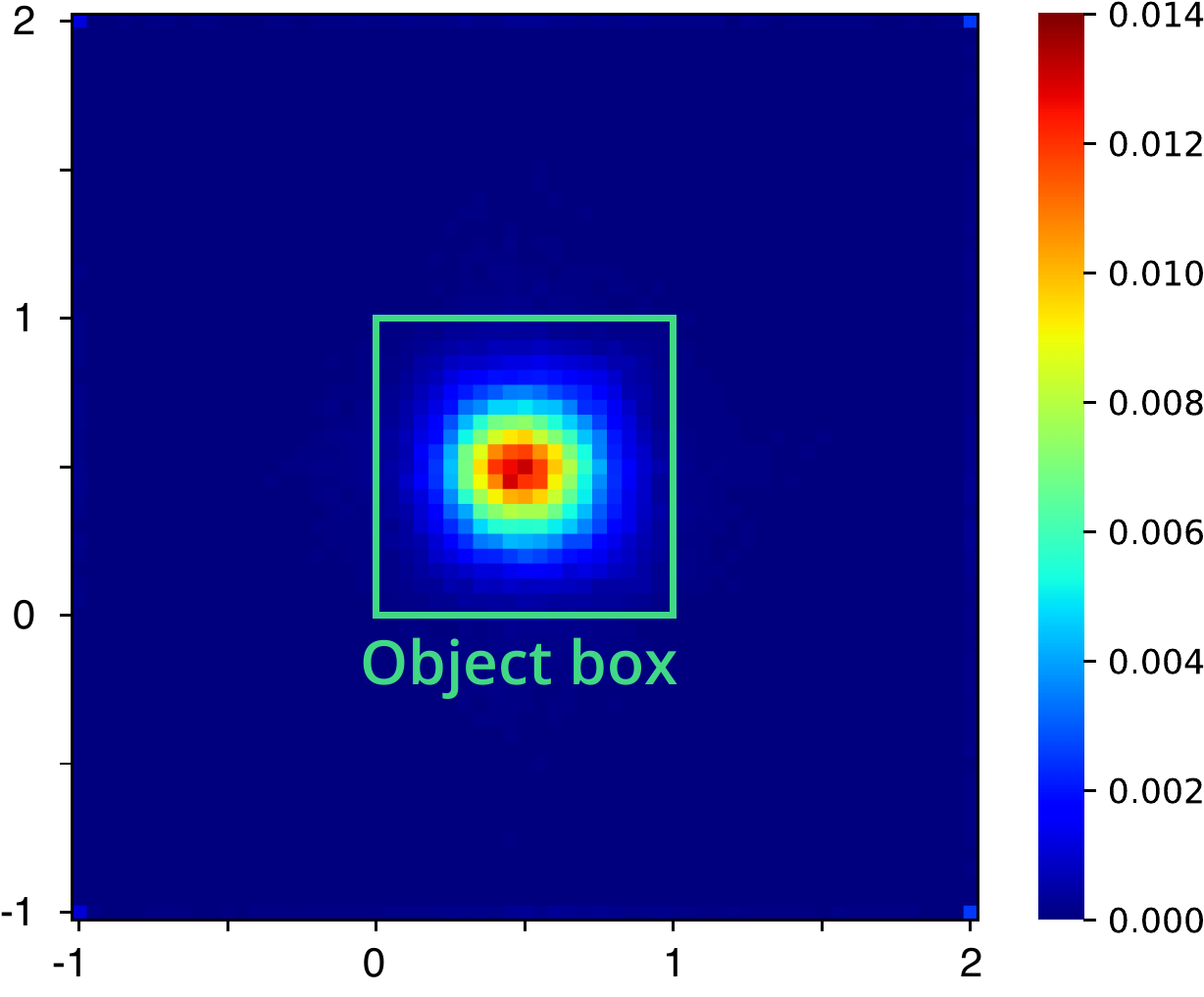}
    \vspace{-2em}
    \caption{\small\textbf{Icon histogram relative to the GT box on COCO.} Distribution of final ``add'' positions normalised against the GT object box frame at $[0,1]\times [0,1]$.}
    \label{appendixfig:gt_relative_icons}
    \vspace{-1em}
\end{figure}

\paragraph{Distribution of objects in COCO.}
COCO is designed to contain multiple objects in the same image. We verify this by computing the histogram of the centres for COCO bounding boxes. Figure \ref{appendixfig:icon_statistics} (left) shows the distribution. Compared to ImageNet (Figure \ref{appendixfig:imagenet-gt-box} left), we observe more diffused box centres in COCO. 
As a result, we observe more diffused object centres for the COCO objects within an image. There are less than 4\% instances in the centre of the image; the ratio was greater than 30\% for ImageNet.

\begin{figure}[t]
    \centering
    \small
    \includegraphics[width=\linewidth]{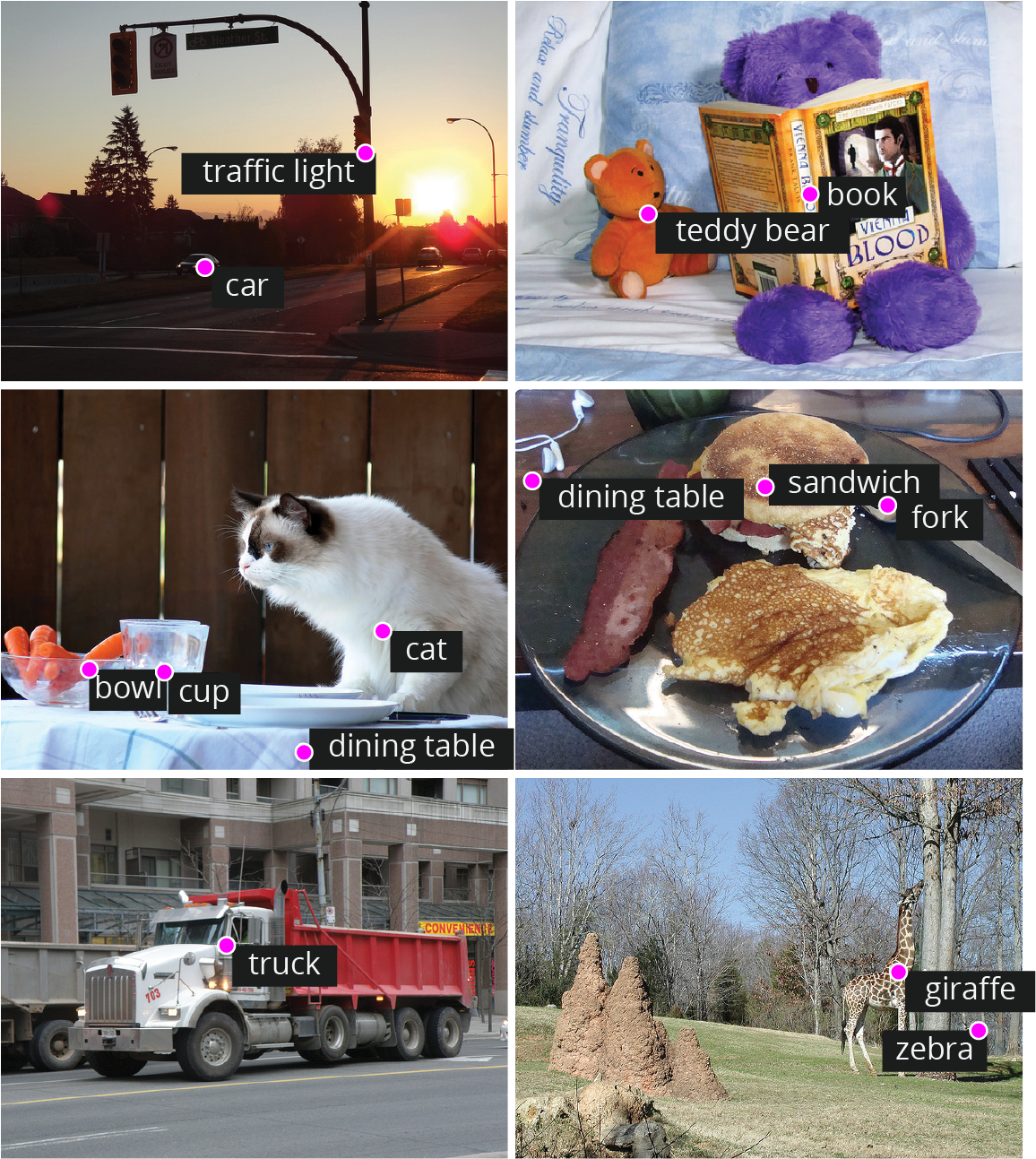}
    \vspace{-1.5em}
    \caption{\small {\bf COCO final icon locations}. We visualise random training images; {\color{hotpink} \textbf{points}} are the final location of the \textttsm{add} action for each category in \textttsm{actionHistories}.}
    \label{fig:data_vis_coco}
\end{figure}

\paragraph{Icon placements.}
Example locations of icon placements are shown in Figure \ref{fig:data_vis_coco}.
The distribution of icon placement locations on COCO images is shown in Figure \ref{appendixfig:icon_statistics} (right). We observe a distribution that is similar to the box-centre distribution, confirming the fairly precise icon placement accuracy of 92.3\% (\S\ref{appendix:byproducts_coco}). We also measure the systematic bias in icon placement with respect to ground-truth bounding boxes in Figure \ref{appendixfig:gt_relative_icons}. We observe no visible bias. This is in stark contrast to the ImageNet click locations in Figure \ref{appendixfig:gt_relative_clicks}. We hypothesise that the tagging interface lets annotators be more focused and be careful with the relative location of the icons with respect to the object regions.

\begin{figure}[t]
    \centering
    \scriptsize
\includegraphics[width=\linewidth]{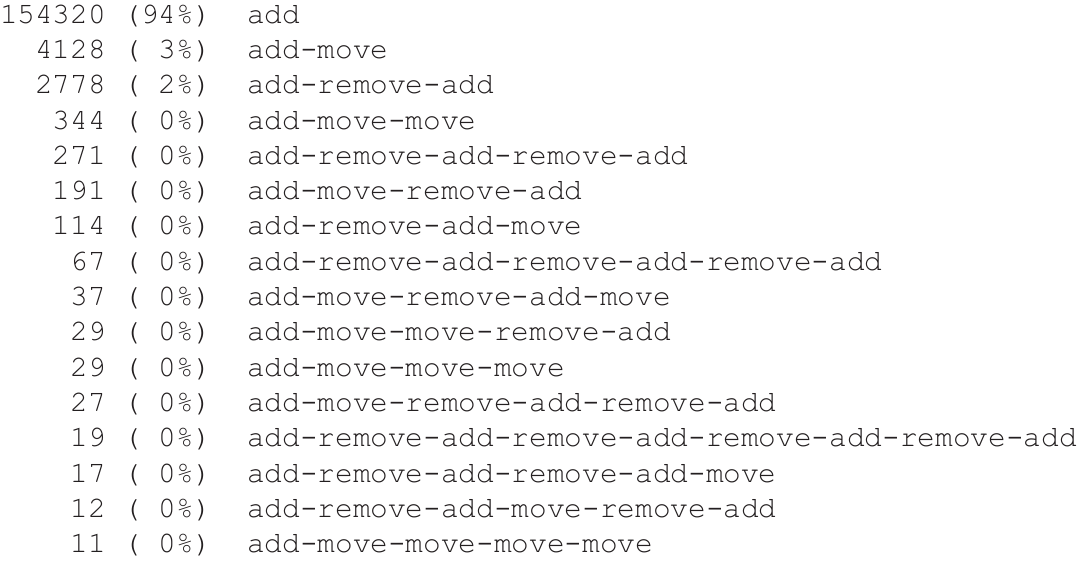}
    \vspace{-2em}
    \caption{\small\textbf{Histogram of action sequences on COCO.} Only showing action sequences with $>10$ occurrences.}
    \label{appendixfig:action_statistics}
    \vspace{-1em}
\end{figure}

\paragraph{Action sequences in COCO annotations.}
Annotators can perform three types of actions with the icons: \textttsm{add}, \textttsm{move}, and \textttsm{remove}. In Figure \ref{appendixfig:action_statistics}, we show the histogram of the action sequences for icons that are eventually placed in the images. The most frequent action sequence is a singleton \textttsm{add} with 94\% frequency. The next common sequence is \textttsm{add-move} with 3\% frequency: the annotator corrects the position once. The third most frequent sequence is \textttsm{add-remove-add} with 2\% frequency: the annotator removes the placed icon and then adds it back. This could indicate the annotator's lack of confidence in either the position of the object or the existence of the object. There are other interesting behaviours. For example, 19 action sequences repeat the addition and removal: \textttsm{(add-remove)*4-add}. We are not sure if this behaviour is due to the annotator's uncertainty or is due to no particular reason (for example, just for fun). In fact, the longest action sequence was \textttsm{add-remove-add-move-(remove-add)*7-move}
\textttsm{-move-(remove-add)*2} (24 actions).

\begin{figure*}[t]
\centering
\includegraphics[width=.9\linewidth]{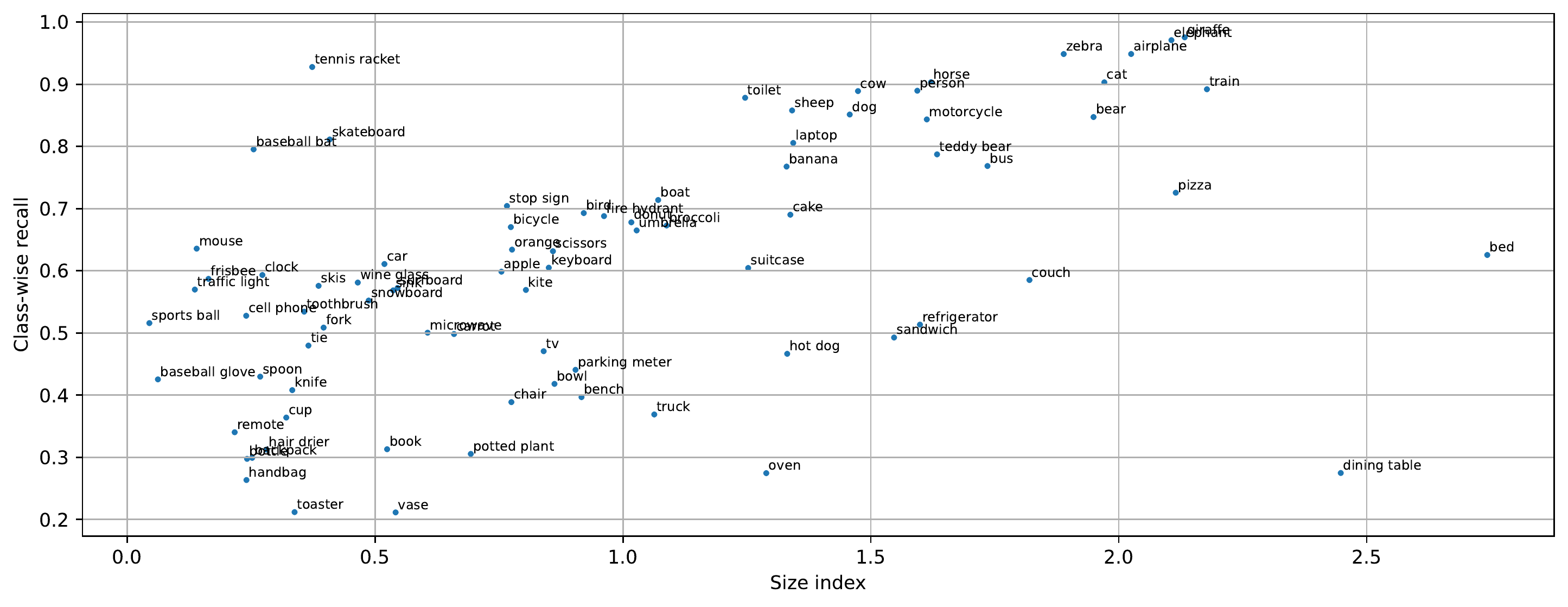}
\vspace{-1em}
\caption{\small\textbf{Recall versus size for each COCO category.}}
\label{appendixfig:recall_size_category}
\end{figure*}

\paragraph{Recall by category and object sizes.}
We study whether the size of objects contributes to the successful annotation of the object. 
Figure \ref{appendixfig:recall_size_category} shows the scatter plot for class-wise recall versus class-wise average size. Class-wise recall measures the chance that an instance of the class in an image is annotated via icon placement. Class-wise sizes are measured by binning the object box by bins $[0,.2^2,.4^2,.6^2,.8^2,1]$. We observe a linear correlation between the object sizes and the recall. This indicates that larger object categories are more likely to be annotated than smaller ones. There are interesting exceptions. For example, sports equipment such as ``tennis racket'', ``skateboard'', ``baseball racket'', ``frisbee'' and ``sports ball'' tends to be annotated successfully compared to their small size. We expect this to be related to the saliency of objects. Sports equipment is likely designed to attract human attention or humans are trained to detect such objects well. In the opposite regime, we find furniture such as ``bed'' and ``dining table'' is less frequently annotated compared to its size. Again, we believe its relative saliency results in low recall. We tend to perceive such furniture more as a background object that is easy to be overlooked in a scene.

\begin{table*}[t]
    \small
    \centering
    \tabcolsep=0.06cm
    \begin{tabular}{l|r|ccc|ccc|cc|cc|ccc|cc}
        \toprule
        \footnotesize Model & \footnotesize Params & \footnotesize IN-1k↑ & \footnotesize IN-V2↑ & \footnotesize IN-Real↑ & \footnotesize IN-A↑ & \footnotesize IN-C↑ & \footnotesize IN-O↑ & \footnotesize Sketch↑ & \footnotesize IN-R↑ & \footnotesize Cocc↑ & \footnotesize ObjNet↑ & \footnotesize SI-size↑ & \footnotesize SI-loc↑ & \footnotesize SI-rot↑ & \footnotesize BGC-gap↓ & \footnotesize BGC-acc↑ \\
        \midrule       
        R18 & 11.7M & 71.8 & 59.7 & 79.4 & \bf 1.9 & 37.1 & 52.6 & \bf 21.9 & 33.8 & 42.7 & 21.8 & 47.5 & 22.2 & 31.9 & 8.6 & \textbf{22.4} \\
        +\ours & 11.7M & \bf 72.0 & \bf 59.9 & \bf 79.5 & 1.8 & \bf  37.8 & \bf 52.6 & 21.7 & \bf 33.8 & \bf 43.6 & \bf 22.0 & \bf 47.6 & \bf 23.5 & \bf 32.2 & \bf 7.4 & 20.1 \\
        \midrule
        R50 & 25.6M & 77.2 & 65.4 & 83.5 & 4.6 & 39.8 & \textbf{57.5} & \bf 25.4 & 37.2 & 53.9 & 27.7 & 54.2 & 31.6 & 39.3 & \textbf{6.0} & 28.8 \\ 
        +\ours & 25.6M & \textbf{77.4} & \textbf{65.8} & \textbf{83.5} & \textbf{5.4} & \textbf{44.1} & 56.2 & 25.1 & \textbf{37.6} & \textbf{54.3} & \textbf{27.7} & \textbf{54.7} & \textbf{31.7} & \textbf{40.2} & 6.4 & \textbf{29.2} \\ 
        \bottomrule
    \end{tabular}
    \vspace{-.5em}
    \caption{\small {\bf An alternative baseline of using annotation byproducts.} We report the performance of the models using annotation byproducts as guidance of feature pooling location at training. The performance improvements here show that this method can also become a potential approach for using annotation byproducts to improve the robustness and localization abilities. A more sophisticated method upon this baseline would improve the numbers more.}
    \label{table:resnet_attpool}
\small
\centering
\vspace{1em}
\tabcolsep=0.06cm
\begin{tabular}{l|r|ccc|ccc|cc|cc|ccc|cc}
\toprule
\footnotesize Model & \footnotesize Params & \footnotesize IN-1k↑ & \footnotesize IN-V2↑ & \footnotesize IN-Real↑ & \footnotesize IN-A↑ & \footnotesize IN-C↑ & \footnotesize IN-O↑ & \footnotesize Sketch↑ & \footnotesize IN-R↑ & \footnotesize Cocc↑ & \footnotesize ObjNet↑ & \footnotesize SI-size↑ & \footnotesize SI-loc↑ & \footnotesize SI-rot↑ & \footnotesize BGC-gap↓ & \footnotesize BGC-acc↑ \\
\midrule
ViT-Ti & 5.7M & 71.8 & 58.8 & 78.6 & 4.8 & 41.4 & 59.1 & 18.6 & 29.6 & 38.7 & 20.1 & 40.6 & 16.5 & 26.2 & 12.1 & 13.6 \\ 
+\ours & 5.7M & \bf 73.0 & \bf 60.2 & \bf 79.8 & \bf 5.7 & \bf 42.5 & \bf 59.9 & \bf 19.4 & \bf 30.8 & \bf 42.6 & \bf 22.1 & \bf 43.4 & \bf 20.0 & \bf 28.7 & \bf 10.9 & \bf 16.1 \\ 
\midrule
ViT-S & 22.1M & 74.1 & 60.8 & 80.4 & 5.1 & 45.0 & 55.0 & 22.9 & 34.7 & \bf 47.0 & 20.5 & 42.9 & 18.7 & 27.8 & 10.5 & 16.7 \\
+\ours & 22.1M & \bf 75.3 & \bf 63.0 & \bf 81.6 & \bf 6.3 & \bf 47.7 & \bf 59.1 & \bf 24.4 & \bf 36.5 & 46.6 & \bf  23.6 & \bf 47.8 & \bf 22.6 & \bf 32.2 & \bf 8.7 & \bf 19.7 \\
\midrule
ViT-B & 86.6M & 75.1 & 61.9 & 81.2 & 6.4 & 48.8 & \bf 56.8 & 24.3 & 36.7 & 48.9 & 21.3 & \bf 47.6 & 22.1 & \bf 31.9 & 8.9 & 18.9 \\
+\ours & 86.6M & \bf 75.9 & \bf 63.0 & \bf 82.1 & \bf 7.6 & \bf 49.9 & 56.5 & \bf 26.4 & \bf 37.2 & \bf 50.3 & \bf 23.2 & 47.4 & \bf 22.5 & 31.7 & \bf 8.0 & \bf 18.9 \\
\bottomrule
\end{tabular}
\vspace{-0.75em}
\caption{\small {\bf Performance of \ourin on ImageNet1K without sophisticated training recipes.} We extend the study in 
Table~\ref{table:main-imagenet}
by training ViTs~\cite{dosovitskiy2020image, deit} with simpler training recipes. We note more significant improvements due to \ourin than shown in
Table~\ref{table:main-imagenet}.
}
\label{table:sub-imagenet}
\vspace{-1mm}
\end{table*}

\section{Additional experimental details}
\label{appendix:further_experimental_results}

\paragraph{Training details.}
For the ImageNet experiments, we use all the default training hyperparameters provided in the DeiT~\cite{deit} codebase\footnote{\url{https://github.com/facebookresearch/deit}} including training epochs 300 with warmup epochs 5, batch size 1024, learning rate 5e-4$\times\frac{\text{batchsize}}{512}$, weight decay 0.05. In addition, we use the default hyperparameters for data augmentations and regularizations -- RandAug~\cite{cubuk2019randaugment} 9/0.5 (\ie rand-m9-mstd0.5-inc1), Label smoothing~\cite{szegedy2016rethinking} 0.1, Stochastic Depth 0.1 with the linear decay of death rate~\cite{stochasticdepth}, and Random Erasing~\cite{randomerasing, cutout} 0.25; Mixup~\cite{mixup} and Cutmix~\cite{cutmix} with the probabilities 0.8 and 1.0, respectively with switching probability 0.5, and the repeated augmentation~\cite{hoffer2020augment} with 3 repetitions. We train the models with the image size of 224$\times$224 and the test crop ratio of 0.875 based on the basic ImageNet training strategy -- RandomResizedCrop, RandomFlip, and ColorJitter following the standard protocol~\cite{resnet,dosovitskiy2020image, deit}. All the models are trained with the multi-task objective using $\lambda{=}10$. 

\begin{figure}[h]
\centering
\includegraphics[width=.8\linewidth]{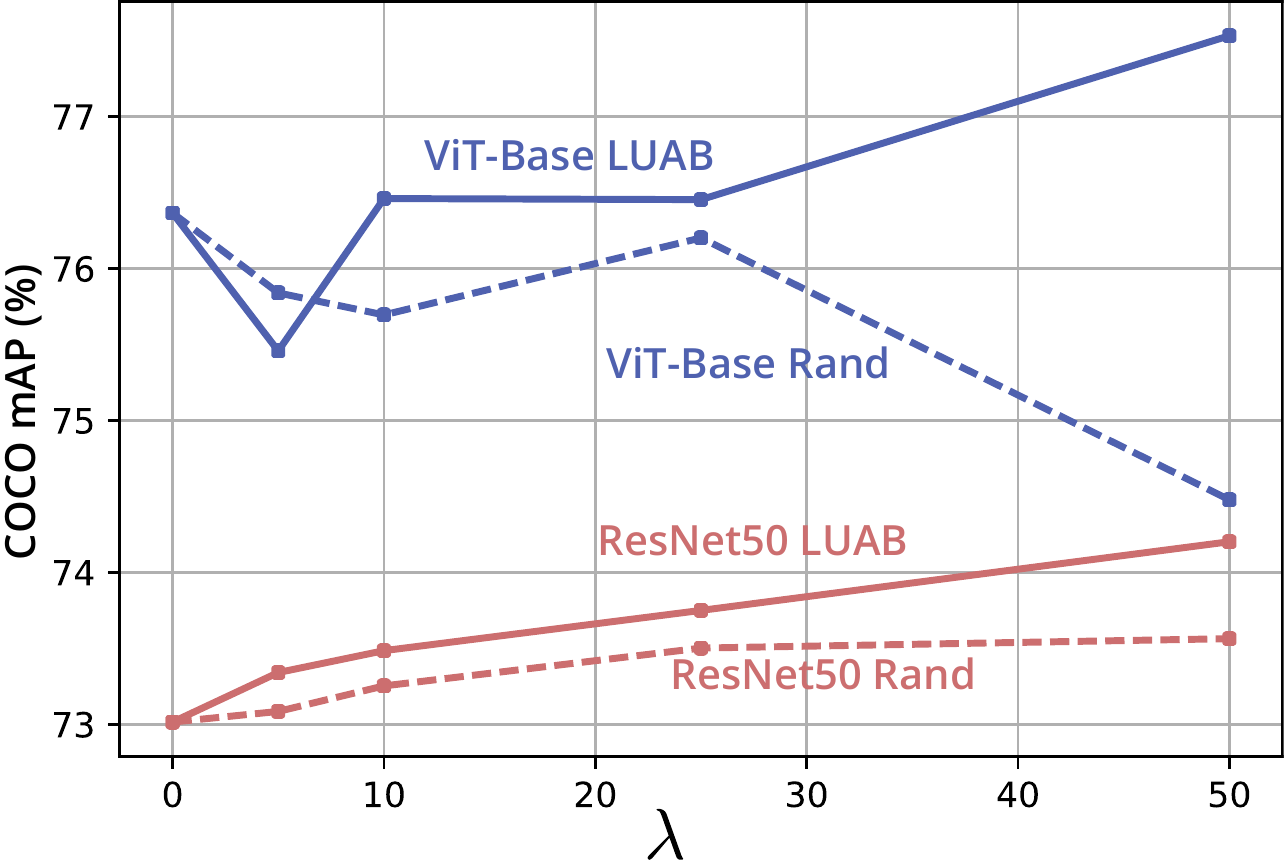}
\vspace{-.5em}
\caption{\small {\bf COCO mAP vs. $\lambda$}.}
\label{table:lambda_coco}
\vspace{0em}
\end{figure}

For the COCO experiments, there is no standard configuration for the image classification task, so we search for hyperparameter sets for convergence of the baseline networks. As a result, we set training epochs to 100 (5 for warmup epochs), batch sizes to 128, image size to 224$\times$224, learning rate to $2\mathrm{e}{-5}$, and weight decay to $0.01$. We use the standard data augmentation of the aforementioned basic ImageNet training strategy for all models. In addition to this, we set the minimum range of RandomResizedCrop to $0.1$, and use Random Erasing~\cite{randomerasing, cutout} with $0.5$. Specifically, we only use We use the AdamP~\cite{heo2020adamp} optimizer for training all backbone networks. For multi-task learning, we observe that small $\lambda$ works well with the small backbone network, and large $\lambda$ is more effective for larger backbone networks. Specifically, we used $\lambda{=}5$ for ResNet18 and ViT-Ti. We used $\lambda{=}50$ for ResNet50, ResNet152, ViT-S, and ViT-B. Figure~\ref{table:lambda_coco} shows that, across all $\lambda$, \ours performs generally better than the models trained with Random points (Rand) or only with task supervision (\ie $\lambda$=0).

\begin{table}[t]
\vspace{0em}
\footnotesize
\centering
\tabcolsep=0.05cm
\begin{tabular}{@{}lccccccc@{}}
\toprule
{Model}             & {IN-1k↑} & {IN-V2↑} & {IN-Real↑} & {ObjNet↑} & {SI-size↑} & {SI-loc↑} & {SI-rot↑} \\ 
\midrule
$\ell^1$ ($\beta{=}1$)  & \underline{77.5}      & \underline{65.2}      & \textbf{78.5}     & \underline{28.5}       & \textbf{55.6}    & \textbf{33.5}   & \textbf{40.9}   \\
$\ell^1$ ($\beta{=}2$)  & 77.4            & 65.2            & 78.2              & 28.0             & 55.2             & 32.0            & 40.5            \\
$\ell^1$ ($\beta{=}0.1$) & 76.5            & 64.0            & 77.7              & 27.1             & 53.2             & 30.0            & 38.6            \\
MSE       & \textbf{77.6}   & \textbf{65.4}   & \underline{78.4}        & \textbf{28.9}    & \underline{55.5}       & \underline{32.6}      & \underline{40.7}      
\\
\bottomrule
\end{tabular}
\vspace{-1em}
\caption{\small\textbf{Exploration of loss functions for regression.}}
\label{appendixtab:loss-functions-regression}
\vspace{1em}
\setlength{\tabcolsep}{.6em}
\small
\centering
\begin{tabular}{lccccc} %
\toprule
Training data & ImageNet & \multicolumn{4}{c}{+Annotation byproducts}\\
\midrule   
\% Data used & 100\% & 100\% & 95\% & 90\% & 80\% \\
ImageNet-1k acc (\%) & 72.8\% & \textbf{72.9}& \textbf{72.9} & 72.4 & 71.7 \\ 
\bottomrule
\end{tabular}
\vspace{-0.75em}
\caption{\small {\bf Data-efficient training with \ours}. The availability of AB let us use slightly less amount of training data (100\%$\rightarrow$95\%).}
\label{table:data-efficiency}
\vspace{0em}
\end{table}

\begin{figure}[t]
    \centering
    \small
    \begin{subfigure}[t]{0.155\textwidth}
         \includegraphics[width=1.0\linewidth, height=1.3\linewidth]{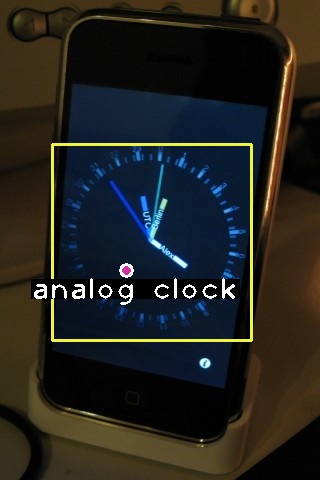}
         \caption{Analog clock}
    \end{subfigure}
    \begin{subfigure}[t]{0.155\textwidth}
         \includegraphics[width=1.0\linewidth, height=1.3\linewidth]{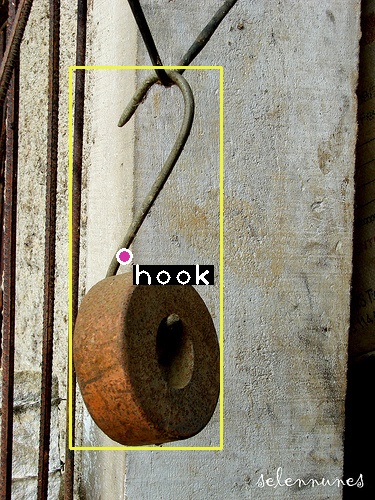}
         \caption{Hook}
    \end{subfigure}
    \begin{subfigure}[t]{0.155\textwidth}
         \includegraphics[width=1.0\linewidth, height=1.3\linewidth]{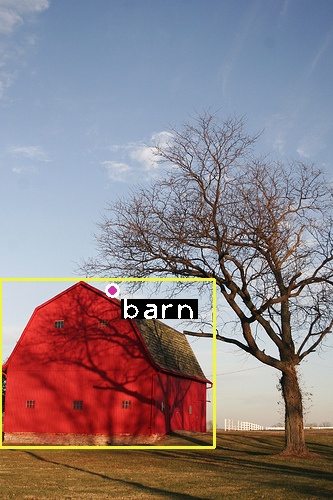}
         \caption{Barn}   
    \end{subfigure}      
    \begin{subfigure}[t]{0.155\textwidth}
         \includegraphics[width=1.0\linewidth, height=1.3\linewidth]{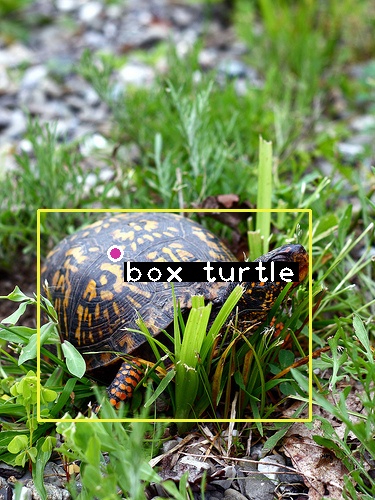}
         \caption{Box turtle}   
    \end{subfigure}    
    \begin{subfigure}[t]{0.155\textwidth}
         \includegraphics[width=1.0\linewidth, height=1.3\linewidth]{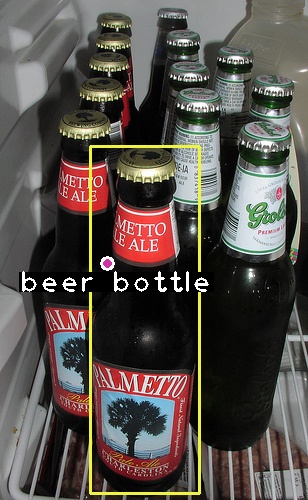}
         \caption{Beer bottle}   
    \end{subfigure}
    \begin{subfigure}[t]{0.155\textwidth}
         \includegraphics[width=1.0\linewidth, height=1.3\linewidth]{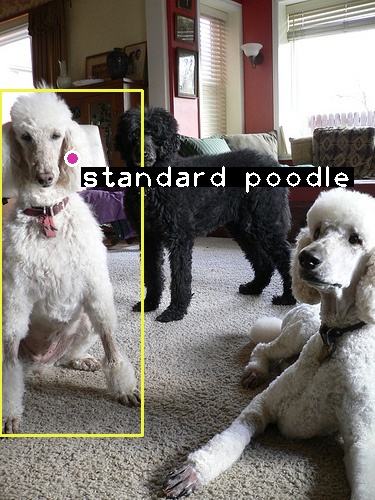}
         \caption{Standard poodle}   
    \end{subfigure}      
    \\
    \begin{subfigure}[t]{0.155\textwidth}
         \includegraphics[width=1.0\linewidth, height=1.3\linewidth]{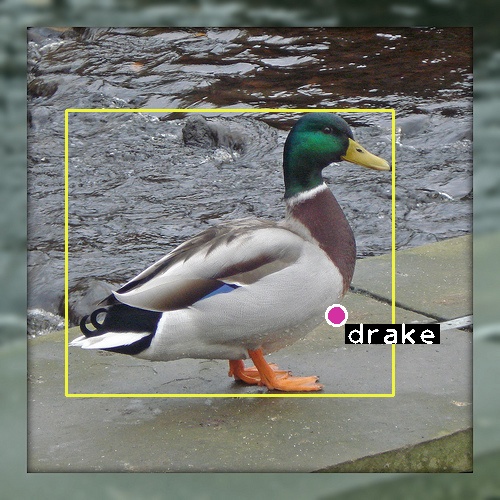}
         \caption{Drake}   
    \end{subfigure}    
    \begin{subfigure}[t]{0.155\textwidth}
         \includegraphics[width=1.0\linewidth, height=1.3\linewidth]{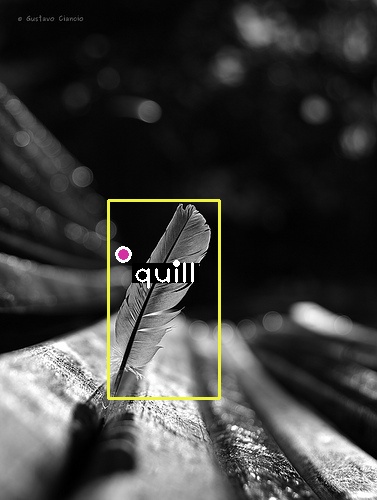}
         \caption{Quill}   
    \end{subfigure}        
    \begin{subfigure}[t]{0.155\textwidth}
         \includegraphics[width=1.0\linewidth, height=1.3\linewidth]{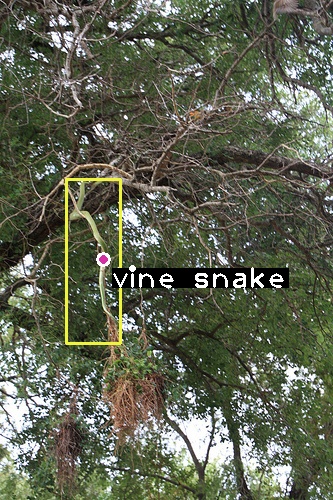}
         \caption{Green mamba}   
    \end{subfigure}     
    \\    
    \begin{subfigure}[t]{0.155\textwidth}
         \includegraphics[width=1.0\linewidth, height=1.3\linewidth]{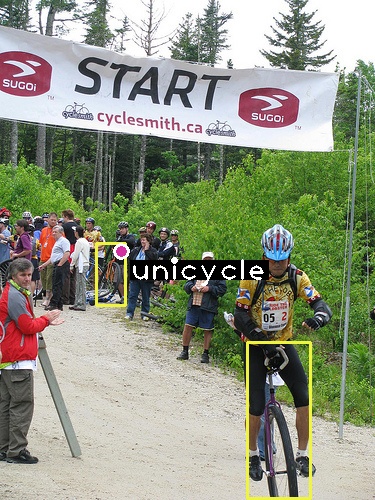}
         \caption{Unicycle}   
    \end{subfigure}        
    \begin{subfigure}[t]{0.155\textwidth}
         \includegraphics[width=1.0\linewidth, height=1.3\linewidth]{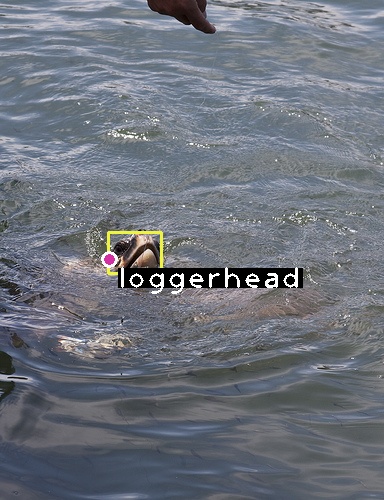}
         \caption{Loggerhead}   
    \end{subfigure}    
    \begin{subfigure}[t]{0.155\textwidth}
         \includegraphics[width=1.0\linewidth, height=1.3\linewidth]{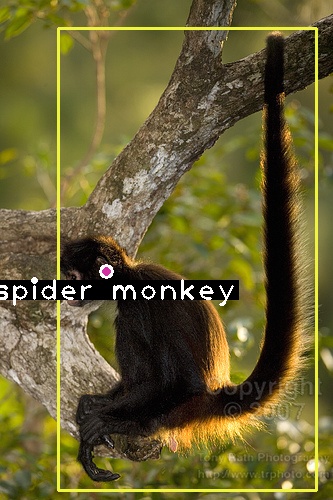}
         \caption{Spider monkey}   
    \end{subfigure}   
    \vspace{-1em}
    \caption{\small {\bf Model prediction visualisation (ImageNet)}. We visualise some validation images in ImageNet with the ground truth \textbf{\color{goldyellow} boxes} and the predicted \textbf{\color{hotpink} points} by our model.}
    \label{fig:vit_b_point_vis}
    \vspace{-1.5em}
\end{figure}

\begin{figure}[t]
    \centering
    \small
    \begin{subfigure}[t]{0.5\textwidth}
        \includegraphics[width=.48\columnwidth]{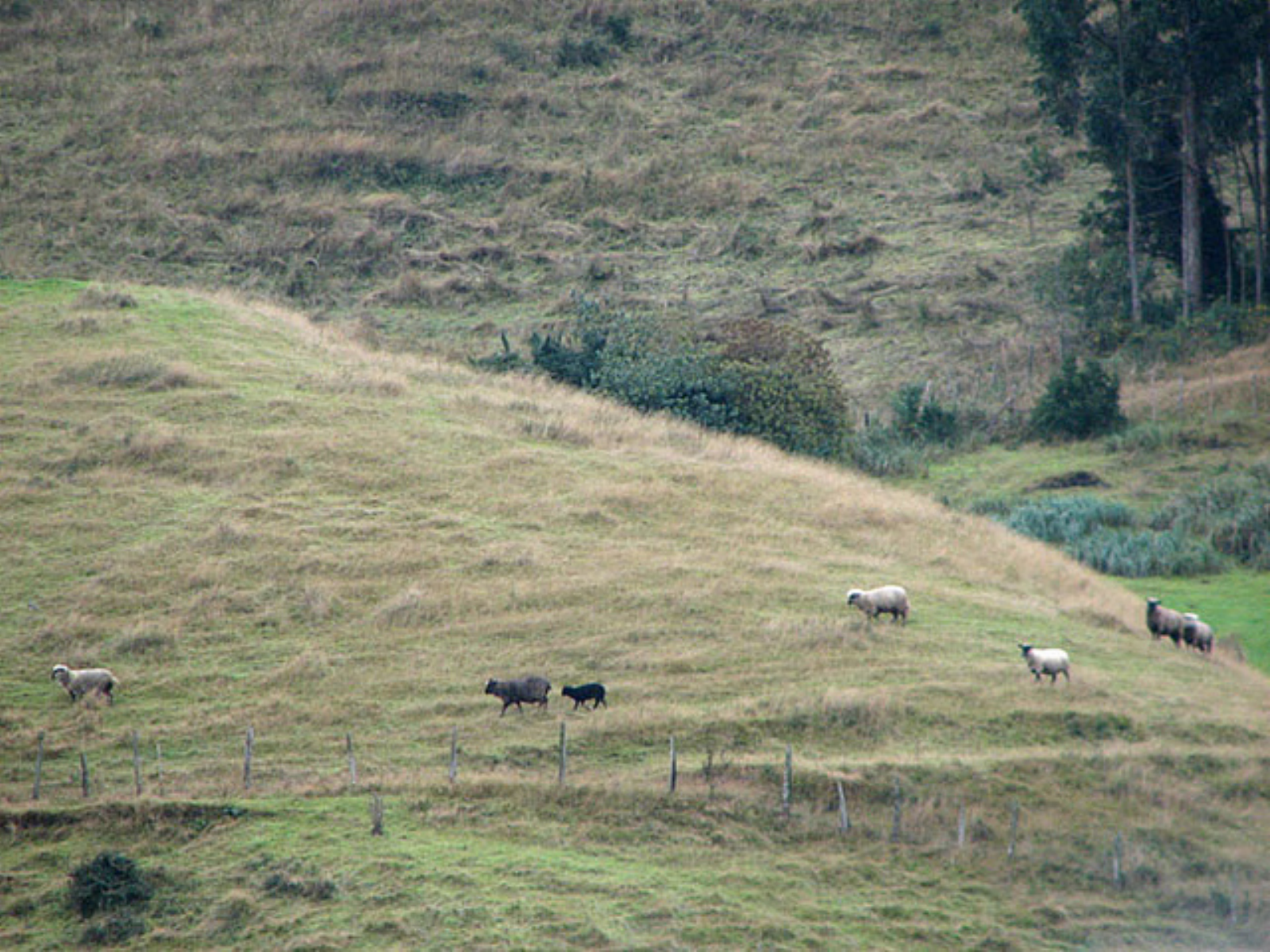}
         \includegraphics[width=.48\columnwidth]{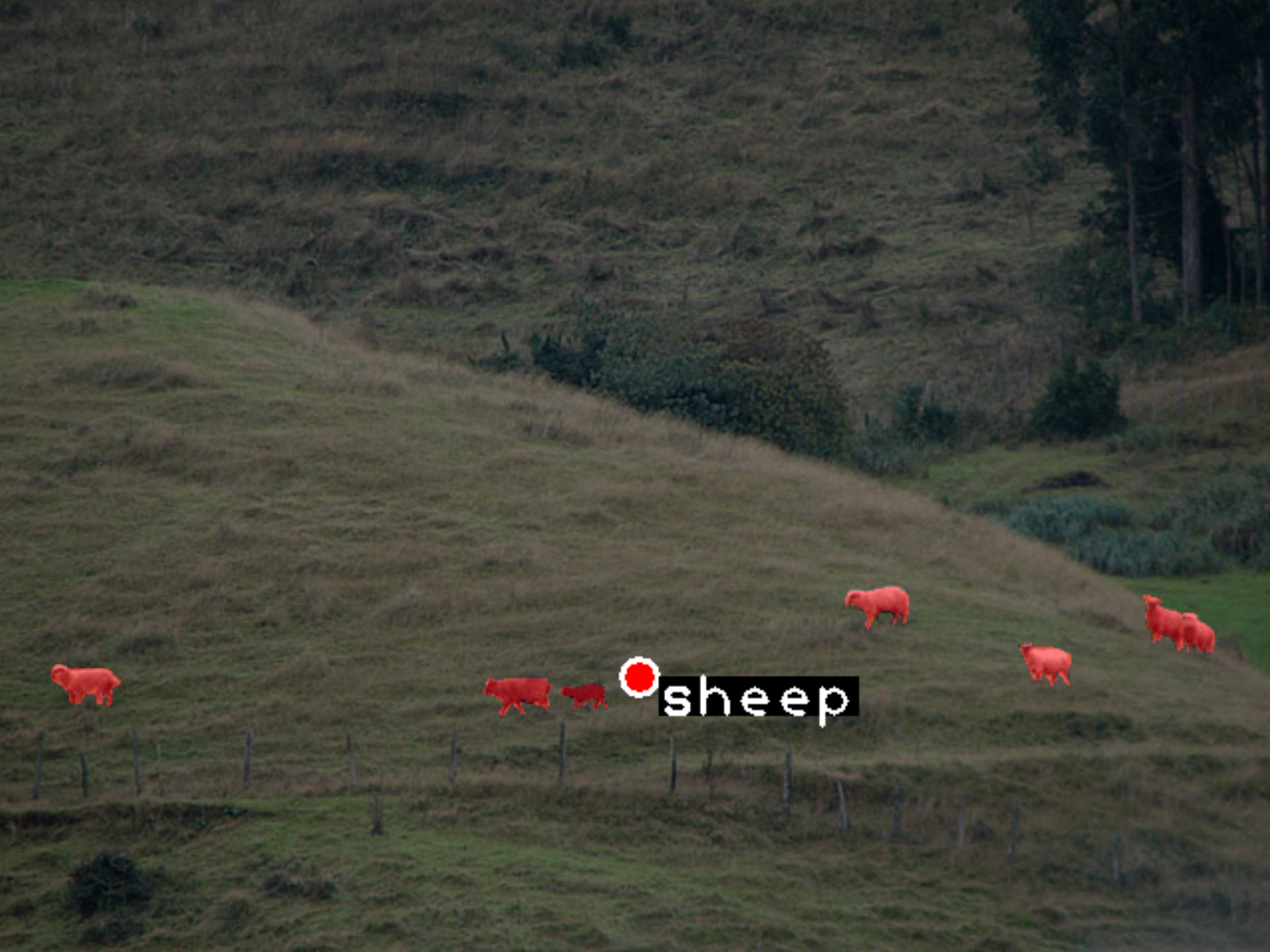}
         
         \caption{Image ID: 116244}
    \end{subfigure}
    \begin{subfigure}[t]{0.5\textwidth}
    \includegraphics[width=.48\columnwidth]{figures/coco_prediction_points/000000416960_input_img.pdf}
         \includegraphics[width=.48\columnwidth]{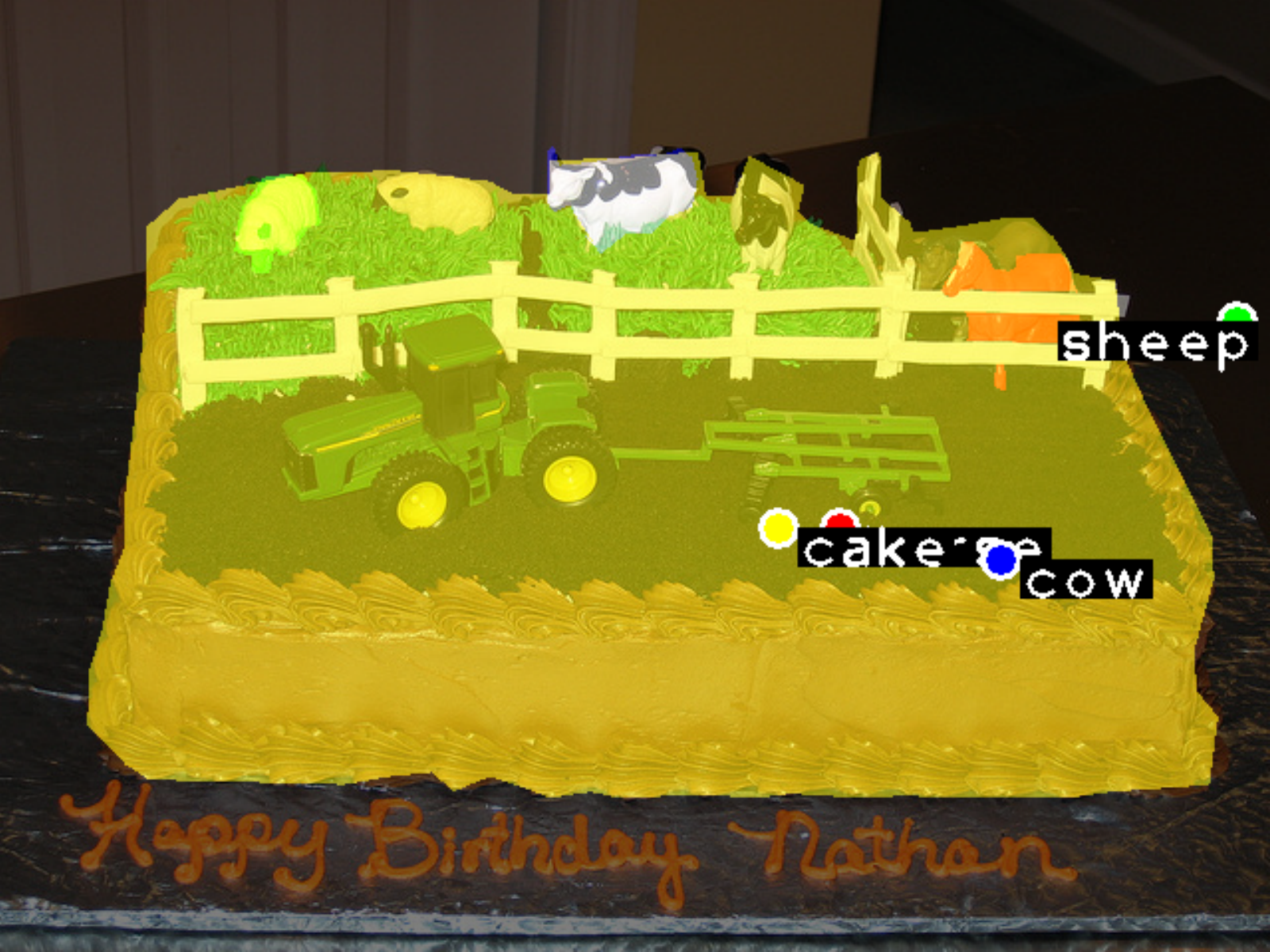}
         
         \caption{Image ID: 416960}
    \end{subfigure}
    \begin{subfigure}[t]{0.5\textwidth}
     \includegraphics[width=.48\columnwidth]{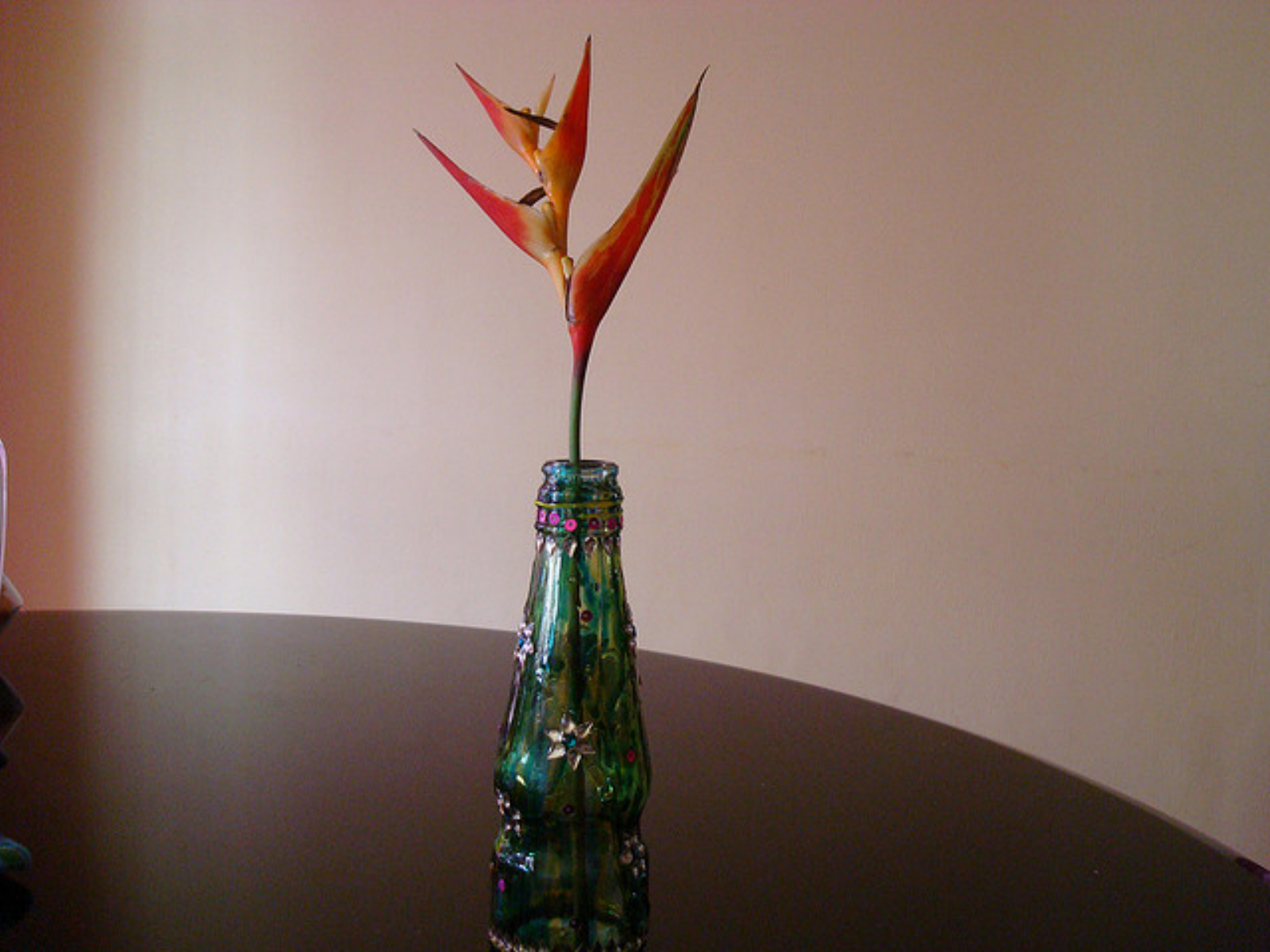}
         \includegraphics[width=.48\columnwidth]{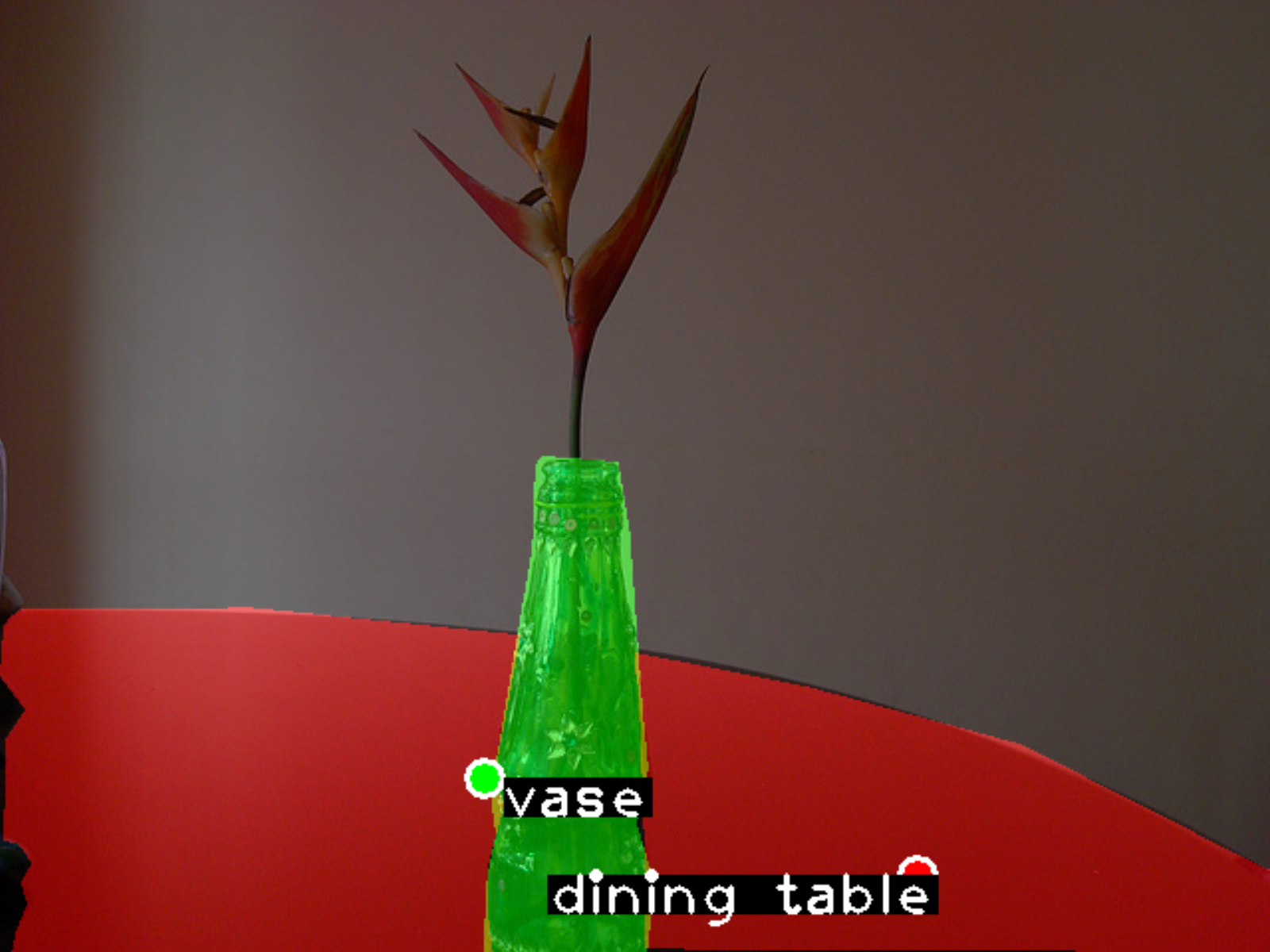}
        
         \caption{Image ID: 430052}
    \end{subfigure}
    \begin{subfigure}[t]{0.5\textwidth}
    \includegraphics[width=.48\columnwidth]{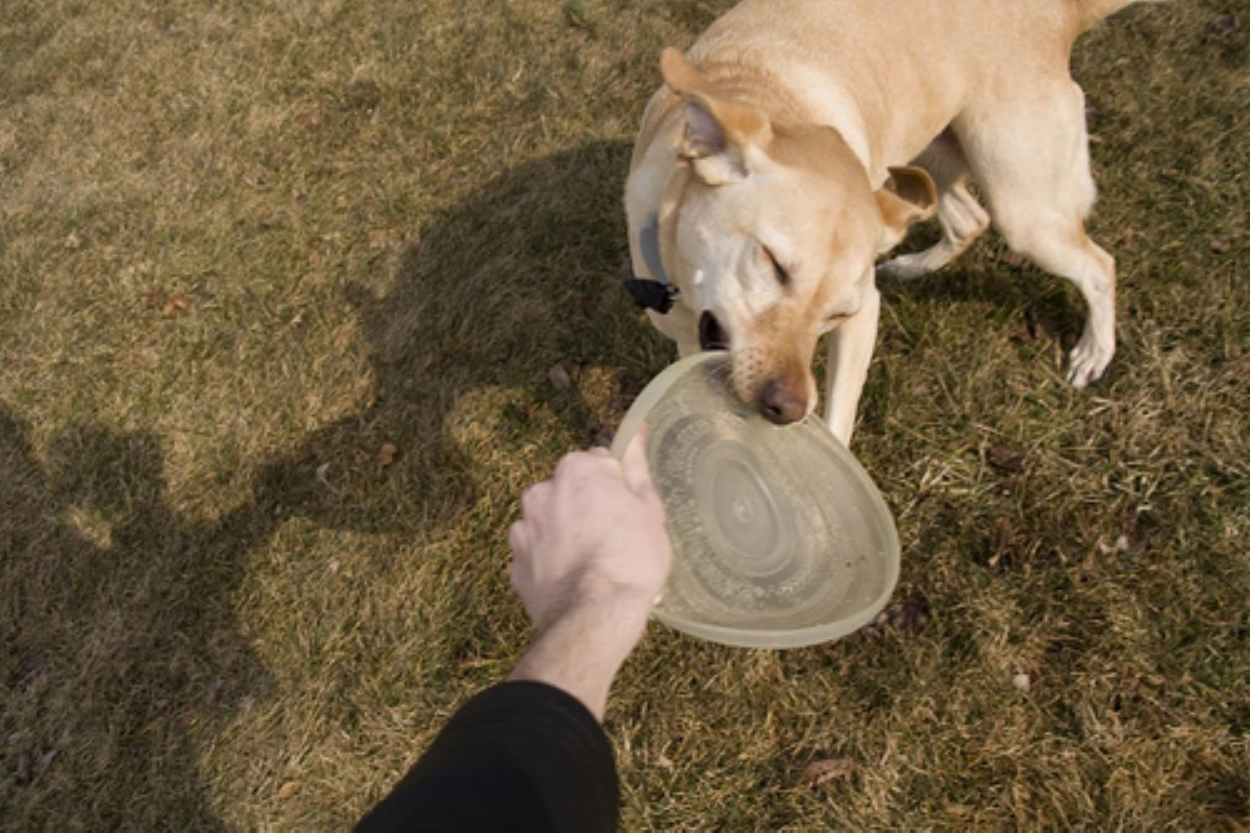}
         \includegraphics[width=.48\columnwidth]{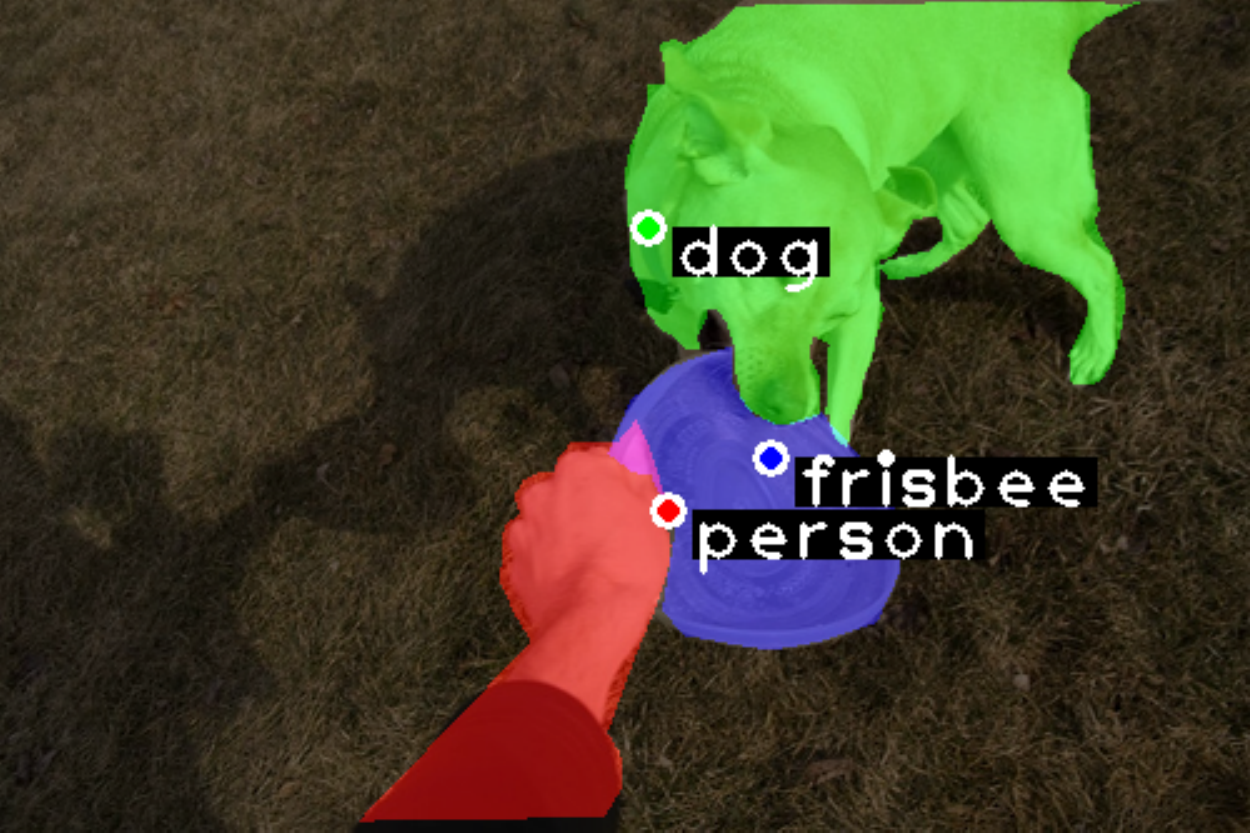}
         
         \caption{Image ID: 442761}
    \end{subfigure}
    \begin{subfigure}[t]{0.5\textwidth}
    \includegraphics[width=.48\columnwidth]{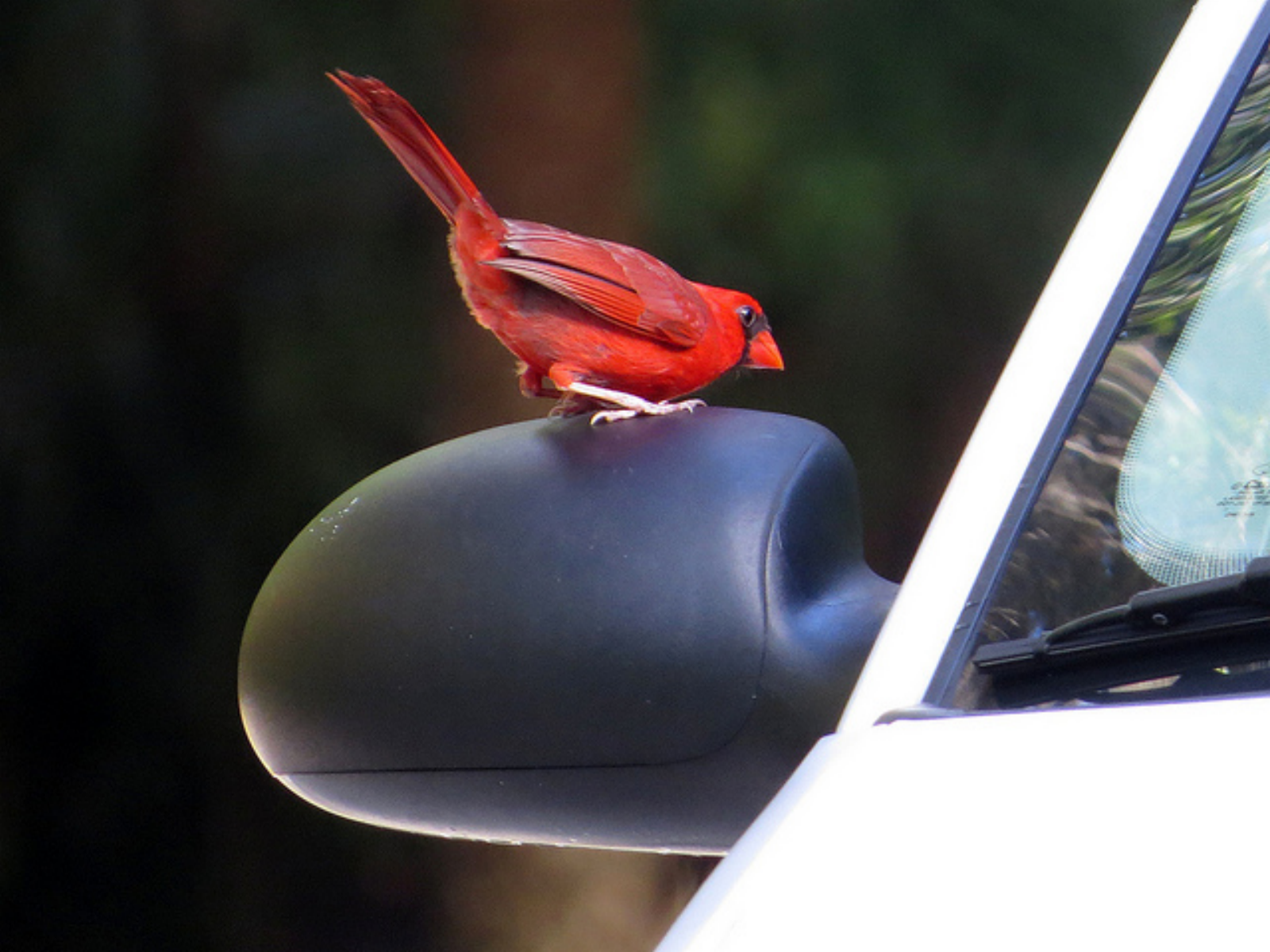}
         \includegraphics[width=.48\columnwidth]{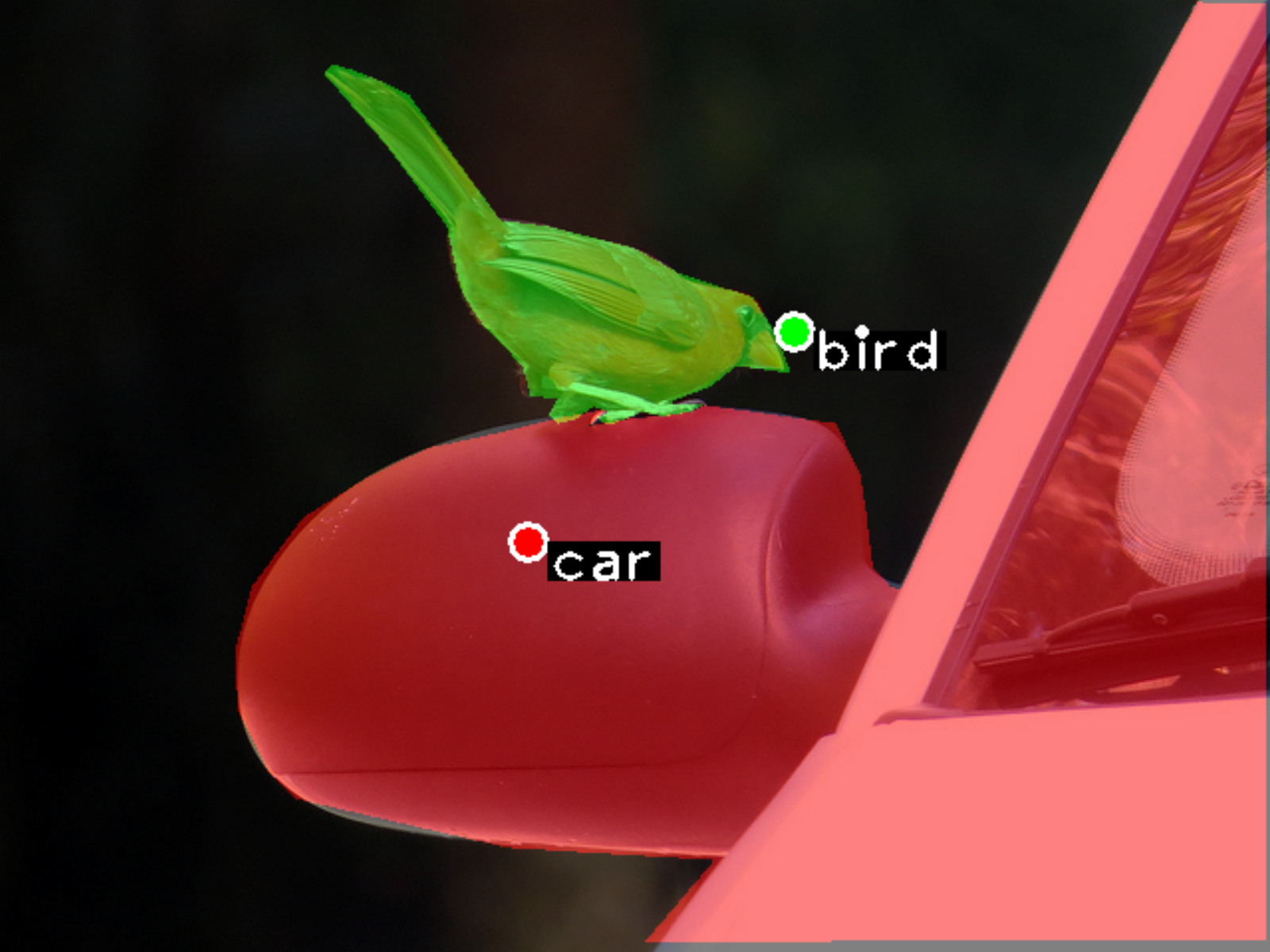}
         
         \caption{Image ID: 217554}
    \end{subfigure}
    \vspace{-1em}
    \caption{\small {\bf Model prediction visualisation (COCO)}. We visualise COCO validation images with the ground truth mask and predicted points by our model.}
    \label{fig:point_vis_coco}
    \vspace{-1.3em}
\end{figure}

\paragraph{Visualisation of the predicted points.} We visualise the points predicted by our \ours-trained models with the annotation byproducts. Figure~\ref{fig:vit_b_point_vis} and \ref{fig:point_vis_coco} show the points predicted by our ViT-B in random ImageNet validation images and by our ResNet50 in random COCO validation images, respectively. We observe the predicted points are aligned with the ground-truth object locations.

\paragraph{Using annotation byproducts for data-efficient learning.} Table~\ref{table:data-efficiency} shows ViT-Ti performances after training with varying amounts of training data. The result shows that we may use 95\% of ImageNet training data without decreasing the performance when annotation byproducts are utilised.

\paragraph{Using annotation byproducts to pool features.} In the main paper, we have introduced a multi-task learning approach with the point-regression objective for the annotation byproducts.
Here, we show another possibility to use the annotation byproducts. We use them as ground-truth attention for a weighted pooling for a convolutional neural network. We design a network architecture with a point-guided (\ie attentive) pooling layer that amplifies the features corresponding to the point coordinates. The experimental result in Table~\ref{table:resnet_attpool} shows that this simple method (without any extensive hyperparameters tuning) improves the overall performance of ResNet18 and ResNet50. As for the multitask learning baseline, this attentive pooling approach improves classification performance, OOD generalisation, and resilience to spurious background correlations. 

\paragraph{Exploration of loss functions.}
Smooth $\ell^1$ (Huber) loss is a natural initial baseline; it has been effective for a similar task of bounding box regression in object detection. We trained ResNet50 with the MSE and Smooth $\ell^1$ loss with $\beta\in\{0.1,1,2\}$. The results in Table~\ref{appendixtab:loss-functions-regression} show that MSE can be an alternative, but the Huber loss is still the best choice.

\paragraph{Impact of \ours without strong augmentations.} 
In the main paper, we have considered the backbones trained with strong augmentations (\eg DeiT) to make the results more relevant to the state-of-the-art models. Here, we examine the impact of \ours without such strong augmentations. We choose ViTs as the baseline models because they usually suffer from data deficiency~\cite{dosovitskiy2020image,deit} and require stronger augmentations. We follow the training setup provided in original ViT~\cite{dosovitskiy2020image}; we limit the strong data augmentation or regularisations previously used. Table~\ref{table:sub-imagenet} shows the performances without strong augmentations such as  RandAug~\cite{cubuk2019randaugment}, Stochastic Depth~\cite{stochasticdepth}, Random Erasing~\cite{randomerasing, cutout}, Mixup~\cite{mixup}, Cutmix~\cite{cutmix} in the DeiT training regime~\cite{deit}. We use a training setup similar to the one in the ViT paper~\cite{dosovitskiy2020image}: learning rate 1e-3 and weight decay 0.3. All the models are trained with the multi-task objective using $\lambda{=}10$ again. We observe that the performance improvements due to \ours are much greater than those in 
Table~\ref{table:main-imagenet}. 
We conclude that the actual impact of annotation byproducts is greater when the performances are not optimised with the use of strong augmentations.

\end{document}